\documentclass[11pt]{article}

\usepackage[preprint]{acl}

\usepackage{times}
\usepackage{latexsym}
\usepackage[T1]{fontenc}
\usepackage[utf8]{inputenc}
\usepackage{microtype}
\usepackage{inconsolata}
\usepackage{graphicx}

\usepackage{booktabs}
\usepackage{tabularx}
\usepackage{adjustbox}
\usepackage{xcolor}
\usepackage[edges]{forest}
\usetikzlibrary{arrows.meta,positioning,calc,fit}
\usepackage{pgfplots}
\pgfplotsset{compat=1.18}

\title{A Two-Way Street: Surveying the Interplay between\\Code-Mixing and Native Languages in Text and Speech}

\author{Debajyoti Mazumder \and Akash Anil \\
  Indian Institute of Science Education and Research, Bhopal \\
  \texttt{debajyoti22@iiserb.ac.in}, \texttt{anila@iiserb.ac.in}}

\begin{document}
\maketitle
\begin{abstract}
Systems for code-mixed (CM) text and speech rarely start from scratch. CM data is scarce, so most work borrows from the languages being mixed, which we call native languages. This transfer can also run in reverse, when CM data is used to improve models for the native languages. Existing surveys focus on processing CM itself, and none brings the two directions together. This paper fills that gap with a systematic survey of transfer in both directions, across text and speech. We organise the surveyed work in a taxonomy of transfer direction, route and modality. Through this taxonomy, we find that most transfer runs from native languages into the CM setting. The reverse direction is younger, and in it code-switching mainly helps by aligning the two languages. Text and speech work share methods but rarely cite each other. On a shared speech benchmark, native data helps more than new architectures. These findings point to open challenges, such as how much native and CM data to mix in training, and how to evaluate CM text written in different scripts.

\end{abstract}

\section{Introduction}\label{sec:intro}
\definecolor{ovN}{RGB}{52,101,164}
\definecolor{ovC}{RGB}{206,92,0}
\definecolor{hiC}{RGB}{0,128,80}
\definecolor{enC}{RGB}{128,0,128}
% Keep column 1 free of top floats so Figure 1 sits atop column 2 of page 1.
\suppressfloats[t]
\begin{figure}[t]
\centering
\begin{tikzpicture}[
  font=\footnotesize,
  lang/.style={draw=black!45, fill=white, rounded corners=2pt,
               inner sep=2pt, minimum width=1.45cm, minimum height=0.42cm},
  outer/.style={draw=black!60, fill=black!4, rounded corners=3pt,
                line width=0.6pt},
  arr/.style={-{Stealth[length=2.6mm,width=2.1mm]}, line width=1.3pt},
  mech/.style={align=center, inner sep=1pt,
               font=\footnotesize\linespread{0.92}\selectfont},
]
\node[font=\footnotesize\bfseries, align=center] (nt) at (0,0.62)
  {Native languages};
\node[lang] (hi) at (0,0.12) {Hindi};
\node[lang] (en) at (0,-0.42) {English};
\node[outer, fit=(nt)(hi)(en), inner sep=3pt] (nat) {};
\node[font=\footnotesize\bfseries, align=center] at (nt) {Native languages};
\node[lang] at (hi) {Hindi};
\node[lang] at (en) {English};
\node[outer, align=center, text width=2.7cm, inner sep=3pt,
      minimum height=1.62cm] (cm) at (4.7,0.1)
  {\textbf{Code-Mix (CM)}\\[3pt]
   \emph{\textcolor{hiC}{Kal} \textcolor{enC}{meeting} \textcolor{hiC}{hai},}\\ \emph{\textcolor{enC}{so I'll be late.}}\\[2pt]
   {\scriptsize\color{black!70}\mbox{`There is a meeting}\\\mbox{tomorrow, so I'll be late.'}}};
\draw[arr, ovN] (nat.north) to[out=58, in=122]
  node[pos=0.5, below=3pt] {\textbf{Native to CM}}
  coordinate[pos=0.5] (ntop) (cm.north);
\node[mech, text=ovN!80!black, above=1pt of ntop] {%
  data $\cdot$ synthesis $\cdot$ models $\cdot$ mapping};
\draw[arr, ovC] (cm.south) to[out=-122, in=-58]
  node[pos=0.5, above=3pt] {\textbf{CM to Native}}
  coordinate[pos=0.5] (cbot) (nat.south);
\node[mech, text=ovC!80!black, below=1pt of cbot] {%
  augmentation $\cdot$ pretraining $\cdot$ LLMs};
\node[draw=black!45, dashed, rounded corners=2pt, fill=white,
      text=black!70, align=center, inner sep=2pt,
      font=\scriptsize\linespread{0.9}\selectfont] (an)
  at ($(nat.east)!0.5!(cm.west)$)
  {Analysis \&\\evaluation};
\draw[black!45, dashed, -{Stealth[length=1.5mm]}] (an.west) -- (an.west -| nat.east);
\draw[black!45, dashed, -{Stealth[length=1.5mm]}] (an.east) -- (an.east -| cm.west);
\end{tikzpicture}
\caption{Two directions of transfer between native languages and the CM setting. Analysis work measures effects on both sides. The example mixes Hindi (green) and English (purple), with its translation below.}
\label{fig:overview}
\end{figure}
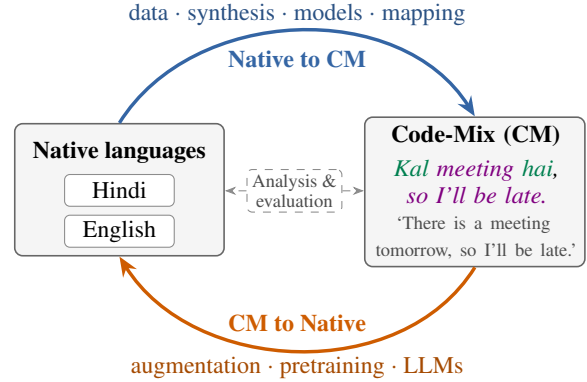

Code-mixing\footnote{As in much of the literature, we treat \emph{code-mixing} and \emph{code-switching} as the same phenomenon.}, the use of two or more languages within one sentence or conversation, is common among multilingual speakers. It is frequent on social media, in messaging and in everyday talk \citep{9074205,dogruoz-etal-2021-survey,srivastava-singh-2020-phinc,hamed-etal-2020-arzen,nguyen-bryant-2020-canvec}, as in the Hindi--English message of Figure~\ref{fig:overview}. Current models handle such input poorly. Large language models lose accuracy on code-mixed input across 18 languages \citep{yang-chai-2025-codemixbench}. Code-switched queries also degrade retrievers, and embedding models drop by up to 27\% across 11 code-switched tasks \citep{zeng-etal-2026-code}. Likewise, speech recognisers' WER rises by 30--50\% relative on code-switched speech \citep{SINGH2025111886}. A core reason is data. Code-mixed data is scarce, while each mixed language has plentiful monolingual data.

Rather than treating code-mixing as an isolated target, we ask what crosses between it and the languages it mixes. A code-mixed (CM) setting never stands alone. It is built from two or more languages, which this survey refers to as \emph{native} languages. Resources cross the boundary between the CM setting and its native languages in both directions (Figure~\ref{fig:overview}). The field has long responded by moving resources from native languages into the CM setting. In text, monolingual or parallel corpora yield synthetic CM sentences \citep{pratapa-etal-2018-language,gupta-etal-2021-training}. In speech, recognisers learn to handle switches from monolingual speech alone \citep{khassanov19_interspeech}. Transfer also runs the other way, though this direction has drawn less attention. Code-switched training data improves zero-shot cross-lingual transfer \citep{ijcai2020p533}. In speech, LLM-generated code-switched text also improves recognition of monolingual English and Mandarin \citep{10389644}.

Yet the literature has never presented the two directions side by side. Our survey shows that their communities rarely meet. Only 10\% of native-to-CM papers cite any CM-to-native paper (Appendix~\ref{app:citesel}). Gains on one side are also rarely checked on the other. Among the papers we review, few test what training on mixed data does to monolingual ability (\S\ref{sec:cm2n_effects}). As a result, findings cannot be compared across directions. In response, we review both directions in one framework, covering text and speech.

The CM setting and many native languages are both low-resource, so knowing which methods and languages help each other is key to better performance. Our taxonomy maps this for researchers facing scarce labelled CM data and costly annotation. With monolingual resources but no CM corpus, one can synthesise CM data (\S\ref{sec:n2cm-synth}) or fine-tune a multilingual model (\S\ref{sec:n2cm-models}). Romanised input may be cheaper to transliterate (\S\ref{sec:n2cm-map}). For a low-resource native language, code-switched training or prompt examples can help (\S\ref{sec:cm2n}). For each option, it summarises what the method needs and what has been shown (Tables~\ref{tab:n2cm} and~\ref{tab:cm2n}). We will release the list of all 1,451 surveyed papers. This survey makes the following contributions:
\begin{itemize}\setlength\itemsep{0pt}
\item \textbf{First systematic survey.} To our knowledge, this is the first systematic survey of knowledge transfer between the CM setting and its native languages, in both directions and across text and speech. It treats transfer from the CM setting back to native languages as a line of work in its own right.
\item \textbf{Taxonomy.} We organise this transfer by direction, route and modality.
\item \textbf{Findings and challenges.} We state four findings and the open challenges they point to.
\end{itemize}

Earlier surveys organise code-switching research by task, era or language. \S\ref{sec:background} compares them and shows that none is organised mainly by transfer direction.

\section{Background and Scope}\label{sec:background}

\paragraph{Transfer.} Transfer occurs when one of five kinds of resource moves between the CM setting and native languages. These are (i)~\emph{data}, including synthetic data, (ii)~\emph{model weights}, through pretraining, fine-tuning or merging, (iii)~\emph{labels or lexicons}, (iv)~\emph{linguistic knowledge}, and (v)~\emph{evaluation}, such as benchmarks reused across the boundary. Each paper is assigned a direction: native-to-CM, CM-to-native, both, or analysis-only. We exclude CM work that states no transfer, and cross-lingual transfer that involves no code-mixing. Inclusion criteria and the search protocol are in Appendix~\ref{app:search}.

\paragraph{Relation to earlier surveys.} The first surveys of code-switching organised the field by task \citep{sitaram2019survey} and by dataset \citep{9074205}. \citet{dogruoz-etal-2021-survey} then added a linguistic and social view, and \citet{winata-etal-2023-decades} traced research trends across decades. In the LLM era, surveys narrowed in scope. Multilingual LLM surveys treat code-switching only as a prompting strategy \citep{qin2024multilingual}. Others focus on speech recognition \citep{11255834,agro-etal-2026-code}, Arabic \citep{hamed-etal-2025-survey} or code-mixed LLMs \citep{gupta2026code}. A recent survey covers 327 LLM-era studies across modalities \citep{sheth-etal-2026-beyond}. Throughout this line of work, the field is organised by task, era or language, never by transfer direction (Table~\ref{tab:surveys} in Appendix~\ref{app:taxonomy}). Some surveys touch on CM-to-native transfer, but only within single sections. None measures how much work runs in reverse, or traces how the two directions evolved. We do both (\S\ref{sec:findings}, Figure~\ref{fig:trend}).

\paragraph{Taxonomy.}\label{sec:taxonomy} Figure~\ref{fig:taxonomy} in Appendix~\ref{app:taxonomy} shows our taxonomy as a tree with three levels. The first level is \emph{direction}, namely native-to-CM, CM-to-native, or analysis. Papers that transfer both ways appear under both directions. The second level is the \emph{route} by which knowledge moves, and these routes structure \S\ref{sec:n2cm-data}--\S\ref{sec:analysis}. The third level is \emph{modality}, text or speech. Each route groups finer \emph{mechanism} labels, such as fine-tuning or pipeline mapping, and Table~\ref{tab:route-mech} gives this mapping. We also label each paper's task family, language-pair group, resource setting and era (Appendix~\ref{app:schema}).

\section{From Native Languages to Code-Mixing: Data, Models and Mapping}
\label{sec:n2cm-data}

Most work that links code-mixing to native languages draws on the native side, where data is plentiful. It reuses monolingual or parallel data directly (\S\ref{sec:n2cm-mono}), turns it into synthetic CM data (\S\ref{sec:n2cm-synth}), adapts native-trained models (\S\ref{sec:n2cm-models}), or maps CM input to native forms (\S\ref{sec:n2cm-map}). Table~\ref{tab:n2cm} compares these approaches, and Appendix~\ref{app:extended} adds examples.

\begin{table*}[t]
\centering\footnotesize
\begin{tabularx}{\textwidth}{@{}>{\raggedright\arraybackslash}p{3.1cm}c>{\raggedright\arraybackslash}p{3.3cm}>{\raggedright\arraybackslash}p{2.4cm}>{\raggedright\arraybackslash}X@{}}
\toprule
Approach & Mod. & What it needs & Evaluated by & What has been shown \\
\midrule
\multicolumn{5}{@{}l}{\textit{Monolingual and parallel data (\S\ref{sec:n2cm-mono})}} \\
Monolingual LM training & T & Monolingual or translated text & CM perplexity & Matches LMs trained on synthetic CM data \citep{9053775} \\
Native samples and resources & T & Labelled native data, treebanks, lexicons & F1, LAS & Helps as a supplement, but trails annotated CM data \citep{10.1145/3726866,iliescu-etal-2021-much} \\
Pooling and LM fusion & S & Monolingual speech, text and LMs & WER, MER & Large gains only when test sentences are in the LM text \citep{8300583} \\
Mixed-lingual TTS & S & Monolingual recordings & MOS & Mixed-lingual voices from monolingual speakers \citep{8554693} \\
\multicolumn{5}{@{}l}{\textit{Synthetic CM data (\S\ref{sec:n2cm-synth})}} \\
Theory-driven generation & T, S & Parallel text, parser or aligner & Perplexity, human rating & About as useful as doubling real CM data \citep{pratapa-etal-2018-language} \\
Neural generators & T, S & Parallel text, some CM data & Perplexity & Learn switching without hand-written rules \citep{winata2018learn,winata-etal-2019-code} \\
Translation pipelines & T & Parallel or monolingual text & BLEU, task F1 & Better translation of heavily mixed input \citep{gupta-etal-2021-training} \\
LLM prompting & T, S & An LLM & Human rating, F1, WER & Quality varies by pair \citep{terblanche-etal-2024-prompting} \\
Audio synthesis & S & Monolingual audio, TTS & WER, MER & Up to 34.4\% lower MER \citep{10446857} \\
\multicolumn{5}{@{}l}{\textit{Adapting native-trained models (\S\ref{sec:n2cm-models})}} \\
Fine-tuning & T, S & Labelled CM data & F1, MER & The default, usually one backbone on one dataset \citep{10849308} \\
Continued pretraining & T & Unlabelled CM corpus & F1 & Stronger where CM corpora exist \citep{nayak-joshi-2022-l3cube} \\
Light adaptation & T, S & Few extra parameters & F1, MER & Merging gives +2--5 F1 \citep{10.1145/3799830.3799852}; adapters add 5.6\% parameters \citep{10446258} \\
Language routing & S & Language labels & MER & Tested mostly on Mandarin--English \citep{wang2023languageroutingmixtureexpertsmultilingual} \\
\multicolumn{5}{@{}l}{\textit{Mapping CM input to native forms (\S\ref{sec:n2cm-map})}} \\
Transliteration and normalisation & T & Transliterator & F1 & Automatic close to manual (76\% vs 78\% F1) \citep{guellil-etal-2018-arabizi} \\
Translation of CM input & T & MT system & Accuracy, F1 & Conflicting evidence against transliteration \citep{gupta-etal-2018-transliteration} \\
Phone-set and script mapping & S & Phone inventories & WER & Small gains, e.g. 1.2\% WER \citep{sivasankaran-etal-2018-phone} \\
Embedding alignment & T, S & Cross-lingual embeddings & F1 & Applied mainly to NER and sentiment \citep{winata-etal-2019-hierarchical} \\
\bottomrule
\end{tabularx}
\caption{Native-to-CM approaches compared. Mod.: modality (T text, S speech). Reported numbers come from different setups and are not comparable across rows.}
\label{tab:n2cm}
\end{table*}

\subsection{Monolingual and Parallel Data}\label{sec:n2cm-mono}

\paragraph{Text.}
The most direct option is to train on native-language data alone. Early CM language models took this route \citep{sreeram2017language}. A later model trains only on monolingual Mandarin and English, but constrains the output embeddings of the two languages to lie close together \citep{9053775}. It matches or beats models trained on synthetic CM data, especially at switch points. Native data also helps beyond language modelling. Monolingual treebanks parse Hinglish tweets without any CM training data \citep{bhat-etal-2017-joining}, and Hindi and English hate samples alone reach 0.60 F1 on Hinglish, against 0.63 with added CM samples \citep{10.1145/3726866}. Native data alone, however, rarely matches annotated CM data. For Spanish--English language identification, a tagger trained only on Wikipedia reaches 92.23\% F1, against 98.43\% for a supervised one \citep{iliescu-etal-2021-much}. Native data thus works best as a supplement. How much annotated CM data it can replace is still open.

\paragraph{Speech.}
Speech faces a sharper version of the problem. A recogniser trained on pooled monolingual data never hears a switch inside an utterance. Early work therefore modelled the switch on top of monolingual data. Dual language models were joined by a switch model \citep{garg18_interspeech}, and constrained output embeddings aligned the two languages without any CM speech \citep{khassanov19_interspeech}. Attention then turned to which monolingual data to use. In hybrid systems, matched monolingual speech helps, while mismatched speech hurts \citep{zhang20r_interspeech}. Later work brings in monolingual resources at decoding, through rescoring, internal language-model estimation and monolingual datastores \citep{9414158,peng2022internal,10890024}. Monolingual utterances can even be concatenated into synthetic CM audio, which trains a recogniser without real CM speech \citep{dhawan2023unified}. Over time, monolingual speech has moved from a fallback to raw material for CM data. Its value still depends on how well it matches the target speech.

Synthesis faces a different problem. In monolingual corpora, each voice is tied to one language, so voice or accent changes at the switch. Early work found mixed-lingual training data preferable where it exists \citep{xue19_interspeech}. Without such data, bilingual phonetic posteriorgrams separate speaker from language \citep{9053094}. Recent work builds CM training clips from monolingual corpora and trains an LLM-based synthesiser on them \citep{11434663}. Synthesis thus follows the same path as recognition, from adapting models to building CM data from native speech.

\subsection{Synthetic Code-Mixed Data from Monolingual Sources}\label{sec:n2cm-synth}

Monolingual data cannot show how two languages combine inside a sentence. Synthetic CM data fills this gap by building mixed sentences from monolingual or parallel text.

\paragraph{Text.}
The first generators drew on linguistic theory. They used the Equivalence Constraint or the Matrix Language Frame to decide where a switch is grammatical \citep{pratapa-etal-2018-language,lee19d_interspeech,rizvi-etal-2021-gcm}. A language model trained first on such synthetic data, then on real CM data, reaches lower perplexity. The gain is about the same as doubling the real CM data \citep{pratapa-etal-2018-language}. Rules, however, may not capture how people actually switch. Neural generators therefore learn switching from data \citep{chang19_interspeech,winata2018learn}, and one pretrained generator adds control over how much to switch \citep{mondal-etal-2022-cocoa}. A third line treats generation as translation, turning parallel or monolingual sentences into CM ones \citep{tarunesh-etal-2021-machine,gupta-etal-2021-training}. The resulting pairs improve translation of heavily mixed input \citep{gupta-etal-2021-training}. Since 2023, prompting LLMs has become the cheapest generator, but its quality depends on the pair \citep{yong-etal-2023-prompting,terblanche-etal-2024-prompting}. ChatGPT writes fluent Singlish but performs poorly on English--Tamil \citep{yong-etal-2023-prompting}. Uneven quality affects every generator, not only LLMs. Filtering out poor sentences helps \citep{sravani-mamidi-2023-enhancing}, but existing metrics are unreliable on code-mixed text \citep{srivastava-singh-2021-hinge}, so there is no trustworthy score to filter by. It is also unclear whether synthetic CM beats generic augmentation. On translation, it is only competitive with backtranslation \citep{jawahar-etal-2021-exploring}.

\paragraph{Speech.}
A recogniser needs CM text for its language model and CM audio for its acoustic model. On the text side, generators follow the same path as in text processing. They moved from the Functional Head Constraint \citep{li-fung-2014-language} to copy-based models \citep{winata-etal-2019-code} and then to lexical replacement or LLMs \citep{10023313,alharbi24_syndata4genai}. For Arabic--English, replacement-based text gave 65.5\% lower perplexity and 7.7\% lower WER \citep{10023313}. A second line synthesises CM audio, starting with TTS and audio splicing. Augmented CM TTS reduced WER by up to 5\% absolute in low-resource settings \citep{sharma20c_interspeech}. Speech Collage splices monolingual segments and smooths the joins, reducing MER by up to 34.4\% in-domain \citep{10446857}. Spliced audio can sound discontinuous, however. Speech editing was proposed in response, and it beats both splicing and TTS \citep{liang2023improving}. Gains are largest over weak baselines. In one study, errors fell by 50\% on code-switched speech but by only 20\% on Mandarin \citep{11434800}. No study in either modality relates generation quality to downstream gain under controlled conditions.

\subsection{Pretrained Models and Their Adaptation}
\label{sec:n2cm-models}

Most CM work does not build new models. It adapts native or multilingual models to CM input.

\paragraph{Text.}
The default route is to fine-tune a multilingual encoder on labelled CM data. Shared tasks made this the norm, especially for Dravidian pairs \citep{sharif-etal-2021-nlp,b-a-2021-ssncse,hossain-etal-2021-nlp-cuet}. Multilingual encoders see little CM text in pretraining. A second line therefore continues pretraining on CM corpora \citep{gupta-etal-2021-task,das-etal-2023-improving}. Where such corpora exist, this is the stronger recipe, as HingBERT shows for Hindi--English \citep{nayak-joshi-2022-l3cube}. Most pairs lack large CM corpora, so lighter routes try to keep the gains at lower cost. Merging a CM-adapted checkpoint with its base model gains 2--5 F1 points over full fine-tuning \citep{10.1145/3799830.3799852}. With LLMs, prompting extends to translation and generation \citep{khatri-etal-2023-translate,gupta2024code,kuwanto-etal-2026-linguistics}. The papers we review mostly report one backbone on one dataset, so it remains unclear when continued pretraining is worth its cost.

\paragraph{Speech.}
Speech adaptation centres on Mandarin--English ASR. Its main problem is that the model must decide which language it hears. Early work added this signal through multitask learning, with auxiliary tasks that predict language or switches \citep{e99-d_10_2554,8300538}. Web-scale multilingual models then changed the recipe. Whisper adapts to Mandarin--English with as little as 1--10 hours of CM data \citep{10849308}, and later work adds language-aware decoding \citep{10889634}. The language decision also shapes architecture. Language-aware encoders separate the two languages \citep{tian2022lae}, and mixture-of-experts models route speech to language experts \citep{wang2023languageroutingmixtureexpertsmultilingual,10890805,10890869}. Adapters offer a cheaper route. One reaches a 14.2\% mixed error rate with only 5.6\% additional parameters over Whisper \citep{10446258}. Most results come from SEAME, so transfer of these designs to other pairs is unclear.

\subsection{Mapping Code-Mixed Input to Native Forms}
\label{sec:n2cm-map}

A fourth route leaves the model unchanged and changes the input instead. It maps CM material to forms that native tools already handle, or aligns both languages in a shared space.

\paragraph{Text.}
Romanised CM text is the main obstacle for native tools, so transliteration is the core operation. Work began with converting Arabizi to Arabic script \citep{darwish-2014-arabizi,al-badrashiny-etal-2014-automatic,shazal-etal-2020-unified}. Transliteration pays off downstream. For Arabizi sentiment analysis, automatic transliteration reaches 76\% F1, close to the 78\% of manual transliteration \citep{guellil-etal-2018-arabizi}. Spelling normalisation also helps tagging \citep{van-der-goot-cetinoglu-2021-lexical} and sentiment \citep{10.1007/978-3-031-23793-5_30}. Whether to transliterate or translate is unsettled. Back-transliterated CM questions outperform translated ones \citep{gupta-etal-2018-transliteration}, while translation helps sentiment \citep{10.1145/3651781.3651819}. A more radical form translates the whole input into one language \citep{may-etal-2014-arabizi,10.1145/3606695}. Embedding alignment avoids explicit mapping, through cross-lingual or meta-embeddings \citep{winata-etal-2019-learning,winata-etal-2019-hierarchical,singh-lefever-2020-sentiment}. The best mapping seems to depend on task and script, which no study tests.

\paragraph{Speech.}
In speech, the mismatch lies in sounds rather than spelling. The central operation is therefore phone-set mapping, which merges the phone inventories of the two languages using acoustic and articulatory cues \citep{4288089,6289009,6763085}. Manual phone merging gives a 1.2\% WER gain \citep{sivasankaran-etal-2018-phone}. A newer line maps the output instead, writing both languages in one script. It uses transliteration-normalised ASR \citep{8639699}, romanisation \citep{10832233} or a language-independent phonetic script \citep{10.3844/jcssp.2025.1176.1186}.

\section{From Code-Mixing to Native Languages: What Code-Mixing Gives Back}
\label{sec:cm2n}

The reverse direction treats code-mixing as a resource rather than a target. Mixed data can improve a model on the native languages, or carry knowledge from a high-resource language to a low-resource one. It takes three forms, namely augmentation (\S\ref{sec:cm2n_aug}), pretraining objectives (\S\ref{sec:cm2n_pretrain}), and curricula and in-context learning for LLMs (\S\ref{sec:cm2n_llm}). Studies of how mixed input affects models are evaluation work, so they appear under analysis (\S\ref{sec:cm2n_effects}). Table~\ref{tab:cm2n} compares the approaches.

\begin{table*}[t]
\centering\footnotesize
\begin{tabularx}{\textwidth}{@{}>{\raggedright\arraybackslash}p{3.1cm}c>{\raggedright\arraybackslash}p{3.3cm}>{\raggedright\arraybackslash}p{2.4cm}>{\raggedright\arraybackslash}X@{}}
\toprule
Approach & Mod. & What it needs & Evaluated by & What has been shown \\
\midrule
Substitution augmentation & T & Bilingual dictionary & Zero-shot accuracy, F1 & Gains on 5 tasks and 19 languages \citep{ijcai2020p533} \\
Generated augmentation & T & Generator or LLM & Accuracy & Low-resource gains without English loss \citep{nagar2025breaking} \\
Contextual switching & T & Model to choose words & POS, NER & Avoids ungrammatical swaps \citep{feng-etal-2022-toward} \\
Mixed data for speech & S & CM text or speech & WER & Also lowers monolingual WER \citep{10389644} \\
Pretraining objectives & T & Parallel or monolingual corpora & BLEU, XNLI & An alignment signal; parallel data may matter more \citep{yang2020alternating,shao-etal-2026-role} \\
Joint monolingual--CM models & S & CM and monolingual speech & WER & Gains on both \citep{madhumani2020learning} \\
Code-switching curricula & T & Synthetic CM corpus & MCQA, MT & Gains in Korean, Japanese and Indonesian \citep{yoo-etal-2025-code-switching} \\
In-context switching & T & An LLM & Accuracy & Up to 14.7pp in low-resource settings \citep{yoo2026gradual} \\
\bottomrule
\end{tabularx}
\caption{CM-to-native approaches compared. Mod.: modality (T text, S speech). Reported numbers come from different setups and are not comparable across rows.}
\label{tab:cm2n}
\end{table*}

\subsection{Code-Switching Augmentation for Cross-Lingual Transfer}
\label{sec:cm2n_aug}

\paragraph{Text.}
Cross-lingual transfer usually trains on English and tests on a language the model has rarely seen. Code-switching narrows this gap by showing target-language words in source-language contexts. The older form is substitution, which swaps words or spans in a monolingual sentence for dictionary translations \citep{3016100.3016300,liu2019attentioninformedmixedlanguagetrainingzeroshot,krishnan-etal-2021-multilingual,betke2022scaling,ji-etal-2023-isotropic}. Applied to fine-tuning data, it improved zero-shot transfer on 5 tasks and 19 languages \citep{ijcai2020p533}. Later variants add consistency regularisation between original and switched inputs \citep{zheng-etal-2021-consistency}, choose the words to switch by saliency \citep{lai-etal-2021-saliency-based}, or switch at test time as well \citep{sheng-etal-2025-test}. A second form generates mixed text rather than editing it \citep{nagar2025breaking}. The recipe now serves spoken language understanding, aspect-based sentiment and MT \citep{liang-etal-2022-label,zhang-etal-2021-cross,song-etal-2019-code}. In cross-lingual retrieval, it gives up to 2x absolute gain for distant languages \citep{litschko-etal-2023-boosting}. Plain switching has a limit, however. Words swapped in without context can make a sentence ungrammatical, which hurts POS tagging and NER. Choosing words that fit the context fixes this across 30+ languages \citep{feng-etal-2022-toward}. How much to switch, and which words, is seldom compared directly.

\paragraph{Speech.}
Few speech papers take this route. One study shows the benefit reaching the native side. LLM-generated code-switched text lowers WER on a code-switching test set. It also lowers WER by 5.3\% and 8.5\% relative on the monolingual English and Mandarin test sets \citep{10389644}. Code-switched speech also enables cross-lingual voice cloning, so a speaker recorded in one language can be synthesised in the other \citep{zhang2021improve}.

\subsection{Code-Switching Pretraining Objectives}
\label{sec:cm2n_pretrain}

\paragraph{Text.}
Augmentation adds switching at fine-tuning time. Pretraining objectives instead build it into the representation from the start. Alternating language modelling pretrains on code-switched sentences built from parallel data. It beats XLM's translation-based pretraining on MT and XNLI \citep{yang2020alternating}. Later work dropped the need for parallel data. Code-switched masked language modelling, which swaps tokens for their model-induced translations before masking, improves cross-lingual NLU and unsupervised MT from monolingual corpora alone \citep{lu-etal-2023-take}. For LLMs, code-switched data is used in pretraining from scratch \citep{wang-etal-2025-investigating-scaling}, and in continued pretraining and instruction tuning \citep{lee-etal-2024-commit}. Curricula move from word-level to sentence-level switching and then to monolingual text. They transfer ability to Korean, Japanese and Indonesian \citep{yoo-etal-2025-code-switching}. Not all evidence is positive. Removing bilingual documents from pretraining cuts translation by 56\% BLEU, but parallel text drives this drop, not code-switched text \citep{shao-etal-2026-role}.

\paragraph{Speech.}
Speech work here is scarce. Code-switching-inspired losses for spoken dialogue improve performance in both monolingual and multilingual settings \citep{colombo-etal-2021-code}. Task-agnostic multitask learning improves both monolingual and code-switched ASR \citep{madhumani2020learning}.

\subsection{Curricula and In-Context Learning for LLM Multilinguality}
\label{sec:cm2n_llm}

\paragraph{Text.}
LLMs add a new lever, since mixing can be applied at inference without retraining. Code-switched demonstrations or instructions steer a model toward a low-resource language \citep{chai2025xcot,shankar-etal-2024-context,11033355}. Gradual code-switching at inference beats cross-lingual in-context learning by 6.0pp in target and 4.8pp in unseen languages. In low-resource target languages, the gain rises to 14.7pp \citep{yoo2026gradual}. A progressive version targets reasoning transfer without expensive distillation \citep{wang2026efficient}. Code-mixing can also serve as an intermediate step. On knowledge queries, native Indian languages trail English by up to about 0.50 in accuracy, while code-mixed versions come within about 0.05. Prompting the model to first rewrite native-language input in code-mixed form yields clear gains over direct prompting \citep{mazumder2026evaluating}. Fine-tuning remains the other route. Mixed-language fine-tuning data is reported to reduce the cross-lingual knowledge barrier \citep{chua2024crosslingual}.

\paragraph{Speech.}
Speech work in this route relies on fine-tuning, and its link to the native side is retention of monolingual ability. A 2020 system recognises code-switched speech without forgetting monolingual recognition, and performs well on both test sets \citep{shah2020learning}. The inference-time lever behind the text gains is untested for speech.

\section{Analysis, Evaluation and Benchmarks}
\label{sec:analysis}

A third body of work neither builds CM systems nor transfers resources. It measures how native-language knowledge relates to CM behaviour, through probing, benchmarks and metrics.

\paragraph{Text.}
Early studies asked whether multilingual encoders represent code-mixed input well. \citet{pires-etal-2019-multilingual} report that mBERT creates multilingual representations that serve code-switching, with caveats for some pairs. \citet{winata-etal-2021-multilingual} find that these representations are not necessarily of high quality. In encoders, code-mixed inputs stay loosely tied to both native languages, and CM adaptation gains English--CM alignment at the cost of English--Hindi alignment \citep{mazumder2026neither}. \citet{santy-etal-2021-bertologicomix} find that natural code-mixed data brings larger improvements than synthetic data. With LLMs, the focus moved to benchmarks, which find that LLMs handle code-mixed input poorly or trail fine-tuned models \citep{zhang-etal-2023-multilingual,yang-chai-2025-codemixbench,mohamed-etal-2026-lost}. Code-mixed prompts also weaken safety alignment \citep{yoo-etal-2025-code}. In one study, attack success rises from 9\% on monolingual inputs to 69\% on code-mixed ones \citep{banerjee2025attributional}. For reasoning, however, mixing helps. Enforcing monolingual decoding lowers math accuracy by 5.6 points \citep{li-etal-2025-impact}.

\paragraph{Speech.}
Speech analysis centres on corpora. Corpora cover Arabic--English, Vietnamese--English, Turkish--German, Mandarin--English and Hindi--English \citep{hamed-etal-2020-arzen,nguyen-bryant-2020-canvec,cetinoglu-2017-code,chua23_interspeech,SINGH2025111886}. Comparisons of monolingual, multilingual and CM-adapted models cut both ways. Monolingual models win on an oral-history archive full of mixed speech \citep{lehecka24_interspeech}, whereas Whisper does better on CM in another study \citep{song2024comparative}. Several papers question WER-based evaluation of CM ASR. Metrics that combine transliteration and normalisation correlate best with human judgement \citep{10023181}.

\paragraph{Effects of mixed input.}\label{sec:cm2n_effects}
A related question is what mixing costs the native side. Most work here measures mixed input rather than mixed training, and finds losses \citep{das2026indi}. Code-mixed prompts lower Pass@1 \citep{sheokand2025codemixbench}, and LLMs often answer in the wrong language \citep{oh-etal-2026-ola}. Safety and sentiment also degrade as mixing increases \citep{jamil-rafi-2026-code,11679710}. Language identification and speech synthesis show the same pattern \citep{ojo2025divers,das2022analysis}. Direct tests of mixed \emph{training} are rare. Where they exist, gains on low-resource languages come without loss on high-resource or monolingual test sets \citep{nagar2025breaking,shah2020learning,madhumani2020learning}.

\section{Findings}
\label{sec:findings}
% Papers per year by transfer direction. Data: tables/papers_per_year.dat
% (bidirectional papers counted in both directions; NONE omitted).
\definecolor{trN}{RGB}{52,101,164}
\definecolor{trC}{RGB}{206,92,0}
\definecolor{trA}{RGB}{140,140,140}
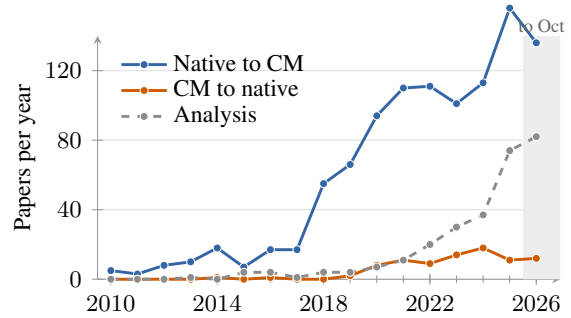
\begin{figure}[t]
\centering
\begin{tikzpicture}
\begin{axis}[
  width=\columnwidth, height=4.8cm,
  font=\footnotesize,
  axis lines=left, axis line style={black!40, line width=0.4pt},
  xmin=2009.5, xmax=2026.9, ymin=0, ymax=140,
  xtick={2010,2014,2018,2022,2026}, minor xtick={2010,...,2026},
  xticklabel style={/pgf/number format/1000 sep=}, 
  ytick={0,40,80,120},
  ymajorgrids, grid style={black!12, line width=0.4pt},
  tick style={black!40, line width=0.4pt},
  ylabel={Papers per year}, ylabel style={yshift=-4pt},
  legend style={draw=none, fill=none, at={(0.03,0.97)}, anchor=north west,
                cells={anchor=west}, font=\footnotesize, row sep=-2pt},
  legend image post style={mark indices={}},
  clip=false,
]
% partial-year band for 2026 (search closed early October 2026)
\addplot[draw=none, fill=black!7, forget plot]
  coordinates {(2025.5,0) (2026.9,0) (2026.9,140) (2025.5,140)} \closedcycle;
\addplot[trN, line width=1.1pt, line join=round, mark=*, mark size=1.5pt,
  mark options={fill=trN, draw=white, line width=0.35pt}]
  table[x=year, y=n2cm] {tables/papers_per_year.dat};
\addlegendentry{Native to CM}
\addplot[trC, line width=1.1pt, line join=round, mark=*, mark size=1.5pt,
  mark options={fill=trC, draw=white, line width=0.35pt}]
  table[x=year, y=cm2n] {tables/papers_per_year.dat};
\addlegendentry{CM to native}
\addplot[trA, line width=1.1pt, line join=round, dashed, mark=*, mark size=1.5pt,
  mark options={solid, fill=trA, draw=white, line width=0.35pt}]
  table[x=year, y=analysis] {tables/papers_per_year.dat};
\addlegendentry{Analysis}
\node[anchor=south, font=\scriptsize, text=black!60, inner sep=1pt] at (axis cs:2026.2,140) {to Oct};
\end{axis}
\end{tikzpicture}
\caption{Included papers per year and direction. Bidirectional papers count in both. Papers before 2010, without a year or without a direction are omitted. 2026 (shaded) runs to early October.}
\label{fig:trend}
\end{figure}

We draw four findings, each measured on the surveyed papers.

\paragraph{Transfer runs almost entirely one way.} Of the 1,451 papers, 1,045 transfer from native languages to the CM setting, 82 transfer back and only 5 do both (Table~\ref{tab:direction-modality}). The imbalance is extreme in speech, with 473 native-to-CM papers against 7 that transfer back. Since 2020, native-to-CM papers have grown from 94 to 156 a year, while CM-to-native papers stayed at 8--18 (Figure~\ref{fig:trend}).

\paragraph{Text and speech share mechanisms, but not literature.} Eight of the eleven transfer mechanisms appear in at least ten papers of each modality (Table~\ref{tab:mechanism-modality}). Yet text and speech papers that use the same mechanism cite each other at only 0.37 of the rate expected under random citation (Appendix~\ref{app:citesel}). The gap is widest for fine-tuning (0.07), adapters and pretraining (both 0). Only synthetic CM data, where speech borrows text generators (\S\ref{sec:n2cm-synth}), shows real exchange (0.57).

\paragraph{CM-to-native transfer mostly mixes languages to align them.} Of the 87 papers that transfer from CM to native languages, 67 insert code-switched text into training, through substitution (33), pretraining objectives (20) or generated data (14). Each places words of two languages in one context, so it works as an alignment signal (\S\ref{sec:cm2n_aug}, \S\ref{sec:cm2n_pretrain}). Where parallel data exists, it carries more of the gain, and code-switching has not been shown to add value beyond it \citep{shao-etal-2026-role}.

\paragraph{On a shared benchmark, native data moves results more than architecture.} SEAME is the only CM benchmark shared by many papers, so we compared 11 papers that report its standard sets, with setups matched within each paper (Appendix~\ref{app:seame}). Once baselines reach 15--20\% MER, changes to architecture or training objective lower dev\_man MER by at most about two points. Adding monolingual Mandarin and English speech to the same model lowers it by 4.6 points \citep{10832233}. Synthetic CM speech still gains 2.0--3.2 points over a fine-tuned Whisper-large-v3 \citep{11249050,yeo2026improving}.

\section{Open Challenges and Future Directions}
\label{sec:challenges}

\paragraph{Balancing monolingual and code-switched data.}
Pretraining language is reported to matter little once a model sees code-mixed data during fine-tuning \citep{tatariya-etal-2023-transfer}. We found no paper that isolates when monolingual transfer turns harmful, for instance through language confusion \citep{lee-etal-2025-controlling}. In the other direction, how much code-switched data the native side needs before gains appear is equally unsettled (\S\ref{sec:cm2n_aug}). Controlled sweeps of the data mix would settle both.

\paragraph{Evaluation across scripts.}
The choice between romanised and native script recurs, from language identification for 22 Indic languages \citep{madhani-etal-2023-bhasa} to safety detection that degrades under transliteration \citep{jamil-rafi-2026-code}. Reference-based metrics break down without a clean monolingual target, and popular CM metrics correlate poorly with human acceptability \citep{srivastava-singh-2021-hinge,10.1145/3748312}. Orthographically informed and semantic metrics are early responses \citep{11464888,jiang2026towards}. A transliteration-normalised score, validated per pair against human judgements, is a next step.

\paragraph{Low-resource constituents.}
Indic--English and Chinese--English dominate, while African pairs are rare, with Yoruba--English and Bambara speech among the few cases \citep{leventhal-etal-2026-kunnafonidilaw,okocha2026beyond}. These constituents lack the strong native models that native-to-CM transfer assumes, so transfer from related high-resource languages is worth testing.

\paragraph{Reporting standards.}
Transfer papers should report monolingual results, natural-CM baselines and several backbones, to let mechanisms be ranked.

\paragraph{Speech-text unification.}
Joint text-speech CM models are largely absent. Audio LLMs are a natural testbed, but they still struggle with Hindi--English ASR \citep{10889964}.

\paragraph{LLM-generated CM and its validity.}
LLMs now generate CM training text, including for CM ASR \citep{yong-etal-2023-prompting,11447403,alharbi24_syndata4genai}. Yet in minimal pairs, larger LLMs prefer natural switches to synthetic ones \citep{sterner-teufel-2025-minimal}. Generated text should therefore be checked for human acceptability.

\section*{Limitations}

Our framing treats the CM setting mainly as a site of transfer. This foregrounds engineering methods and leaves aside work on why and how people mix languages, which \citet{dogruoz-etal-2021-survey} survey.

Our labels were produced by an LLM from titles and abstracts. Scope was then checked by a keyword audit, and we read full texts only for papers we discuss in detail. A human evaluation of 100 random included papers found 2 incorrect and 3 partially incorrect labellings (Appendix~\ref{app:humaneval}). A few errors may remain in rare label classes, which a random sample covers thinly. The 1,681 ``maybe'' papers, which concern the CM setting but state no transfer, were not adjudicated, and some of them may involve transfer that the abstract does not mention.

Recall was estimated only indirectly, by comparing multiple search routes: Semantic Scholar queries, the ACL Anthology, and citation chasing. The last route added few new candidates, which is encouraging but not proof of coverage. No recall check against an independent list of papers was completed. Semantic Scholar covers IEEE and ISCA venues only in part, which matters most for speech, and about 29\% of speech candidates had no abstract, so their labels rest on titles alone. Text may therefore be better covered than speech.

Era and language-pair groups were assigned by rules from publication year and language codes, and may misplace papers with missing or unusual metadata. We cite only part of the papers in scope, so the choice of what to discuss reflects our selection. A paper's absence from the text does not mean it is unimportant. We make no claim about the relative size of effects across papers, because the reported numbers come from different data sets and settings.

\section*{Ethical Considerations}

This survey uses only publicly available papers, including arXiv preprints, with no personal data and no human participants. Code-mixing is tied to community identity, and many pairs are low-resource, so we caution that better tools for them should be built with speaker communities, and that automatic labels should be checked by humans, as we did on a random sample. We used AI assistants in preparing this paper. They helped with LaTeX formatting, with grammar and wording, and with searching for related literature. The LLM used to label papers is a separate part of our method, described in Appendix~\ref{app:search}.

% Comment out for anonymous review (ARR forbids acknowledgments there).
\section*{Acknowledgments}

We thank Uddisa Kkaarthiek, a BS student from IISER Bhopal, for assistance during paper collection and management.

\bibliography{main}

\begin{thebibliography}{386}
\providecommand{\natexlab}[1]{#1}

\bibitem[{Abdoli et~al.(2026)Abdoli, Al-Sumaidaee, Taylor, ElShiekh, and
  Rashad}]{abdoli2026benchmarking}
Sajjad Abdoli, Ghassan Al-Sumaidaee, Clayton~W Taylor, Ahmad ElShiekh, and
  Ahmed Rashad. 2026.
\newblock Benchmarking commercial {ASR} systems on code-switching speech:
  {Arabic}, {Persian}, and {German}.
\newblock \emph{arXiv preprint arXiv:2605.19069}.

\bibitem[{Adel et~al.(2013)Adel, Vu, Kraus, Schlippe, Li, and
  Schultz}]{6639306}
Heike Adel, Ngoc~Thang Vu, Franziska Kraus, Tim Schlippe, Haizhou Li, and Tanja
  Schultz. 2013.
\newblock \href {https://doi.org/10.1109/ICASSP.2013.6639306} {Recurrent neural
  network language modeling for code switching conversational speech}.
\newblock In \emph{2013 IEEE International Conference on Acoustics, Speech and
  Signal Processing}, pages 8411--8415.

\bibitem[{Aditya et~al.(2024)Aditya, Rohmatillah, Tai, and Chien}]{10446258}
Bobbi Aditya, Mahdin Rohmatillah, Liang-Hsuan Tai, and Jen-Tzung Chien. 2024.
\newblock \href {https://doi.org/10.1109/ICASSP48485.2024.10446258}
  {Attention-guided adaptation for code-switching speech recognition}.
\newblock In \emph{ICASSP 2024 - 2024 IEEE International Conference on
  Acoustics, Speech and Signal Processing (ICASSP)}, pages 10256--10260.

\bibitem[{Adouane and Bernardy(2020)}]{adouane-bernardy-2020-multi}
Wafia Adouane and Jean-Philippe Bernardy. 2020.
\newblock \href {https://aclanthology.org/2020.calcs-1.3/} {When is multi-task
  learning beneficial for low-resource noisy code-switched user-generated
  {A}lgerian texts?}
\newblock In \emph{Proceedings of the 4th Workshop on Computational Approaches
  to Code Switching}, pages 17--25, Marseille, France. European Language
  Resources Association.

\bibitem[{Agro et~al.(2026)Agro, Kulkarni, Kadaoui, Talat, and
  Aldarmaki}]{agro-etal-2026-code}
Maha~Tufail Agro, Atharva~A. Kulkarni, Karima Kadaoui, Zeerak Talat, and Hanan
  Aldarmaki. 2026.
\newblock \href {https://doi.org/10.63317/477upr9ikf9n} {Code-switching in
  end-to-end automatic speech recognition: A systematic literature review}.
\newblock In \emph{Proceedings of the Fifteenth Language Resources and
  Evaluation Conference}, pages 9790--9812, Palma de Mallorca, Spain. ELRA
  Language Resource Association.

\bibitem[{Ahmed and Tan(2012)}]{6473715}
Basem~H.A. Ahmed and Tien-Ping Tan. 2012.
\newblock \href {https://doi.org/10.1109/IALP.2012.28} {Automatic speech
  recognition of code switching speech using 1-best rescoring}.
\newblock In \emph{2012 International Conference on Asian Language Processing},
  pages 137--140.

\bibitem[{Al~Ali and Aldarmaki(2024)}]{al-ali-aldarmaki-2024-mixat}
Maryam~Khalifa Al~Ali and Hanan Aldarmaki. 2024.
\newblock \href {https://aclanthology.org/2024.sigul-1.26/} {{Mixat}: A data
  set of bilingual {Emirati}-{E}nglish speech}.
\newblock In \emph{Proceedings of the 3rd Annual Meeting of the Special
  Interest Group on Under-resourced Languages @ LREC-COLING 2024}, pages
  222--226, Torino, Italia. ELRA and ICCL.

\bibitem[{Al-Badrashiny et~al.(2014)Al-Badrashiny, Eskander, Habash, and
  Rambow}]{al-badrashiny-etal-2014-automatic}
Mohamed Al-Badrashiny, Ramy Eskander, Nizar Habash, and Owen Rambow. 2014.
\newblock \href {https://doi.org/10.3115/v1/W14-1604} {Automatic
  transliteration of {R}omanized dialectal {A}rabic}.
\newblock In \emph{Proceedings of the Eighteenth Conference on Computational
  Natural Language Learning}, pages 30--38, Ann Arbor, Michigan. Association
  for Computational Linguistics.

\bibitem[{Alam and ul~Hussain(2017)}]{8289449}
Mehreen Alam and Sibt ul~Hussain. 2017.
\newblock \href {https://doi.org/10.1109/INMIC.2017.8289449} {Sequence to
  sequence networks for {Roman}-{Urdu} to {Urdu} transliteration}.
\newblock In \emph{2017 International Multi-topic Conference (INMIC)}, pages
  1--7.

\bibitem[{Alastruey et~al.(2023)Alastruey, Sperber, Gollan, Telaar, Ng, and
  Agarwal}]{alastruey-etal-2023-towards}
Belen Alastruey, Matthias Sperber, Christian Gollan, Dominic Telaar, Tim Ng,
  and Aashish Agarwal. 2023.
\newblock \href {https://aclanthology.org/2023.calcs-1.2/} {Towards real-world
  streaming speech translation for code-switched speech}.
\newblock In \emph{Proceedings of the 6th Workshop on Computational Approaches
  to Linguistic Code-Switching}, pages 14--22, Singapore. Association for
  Computational Linguistics.

\bibitem[{Alharbi et~al.(2024)Alharbi, Binmuqbil, Ali, Aloraini, Bari,
  Alowisheq, and Alonaizan}]{alharbi24_syndata4genai}
Sadeen Alharbi, Reem Binmuqbil, Ahmed Ali, Raghad Aloraini, Saiful Bari, Areeb
  Alowisheq, and Yaser Alonaizan. 2024.
\newblock \href {https://doi.org/10.21437/SynData4GenAI.2024-6} {{Leveraging
  LLM for Augmenting Textual Data in Code-Switching ASR: Arabic as an
  Example}}.
\newblock In \emph{{Synthetic Data’s Transformative Role in Foundational
  Speech Models}}, pages 26--30.

\bibitem[{Ali et~al.(2021)Ali, Chowdhury, Hussein, and
  Hifny}]{ali21b_interspeech}
Ahmed Ali, Shammur~Absar Chowdhury, Amir Hussein, and Yasser Hifny. 2021.
\newblock \href {https://doi.org/10.21437/Interspeech.2021-2231} {{Arabic
  Code-Switching Speech Recognition Using Monolingual Data}}.
\newblock In \emph{{Interspeech 2021}}, pages 3475--3479.

\bibitem[{Appicharla et~al.(2021)Appicharla, Gupta, Ekbal, and
  Bhattacharyya}]{appicharla-etal-2021-iitp}
Ramakrishna Appicharla, Kamal~Kumar Gupta, Asif Ekbal, and Pushpak
  Bhattacharyya. 2021.
\newblock \href {https://doi.org/10.18653/v1/2021.calcs-1.5} {{IITP}-{MT} at
  {CALCS}2021: {E}nglish to {H}inglish neural machine translation using
  unsupervised synthetic code-mixed parallel corpus}.
\newblock In \emph{Proceedings of the Fifth Workshop on Computational
  Approaches to Linguistic Code-Switching}, pages 31--35, Online. Association
  for Computational Linguistics.

\bibitem[{Badino et~al.(2004)Badino, Barolo, and
  Quazza}]{badino04b_interspeech}
Leonardo Badino, Claudia Barolo, and Silvia Quazza. 2004.
\newblock \href {https://doi.org/10.21437/Interspeech.2004-312} {{A general
  approach to TTS reading of mixed-language texts}}.
\newblock In \emph{{Interspeech 2004}}, pages 849--852.

\bibitem[{Baghel et~al.(2023)Baghel, Ramoji, Sidharth, H, Singh, Jain, {Roy
  Chowdhuri}, Kulkarni, Padhi, Vijayasenan, and
  Ganapathy}]{baghel23_interspeech}
Shikha Baghel, Shreyas Ramoji, Sidharth, Ranjana H, Prachi Singh, Somil Jain,
  Pratik {Roy Chowdhuri}, Kaustubh Kulkarni, Swapnil Padhi, Deepu Vijayasenan,
  and Sriram Ganapathy. 2023.
\newblock \href {https://doi.org/10.21437/Interspeech.2023-2367} {{The DISPLACE
  Challenge 2023 - DIarization of SPeaker and LAnguage in Conversational
  Environments}}.
\newblock In \emph{{Interspeech 2023}}, pages 3562--3566.

\bibitem[{Baly et~al.(2017)Baly, El-Khoury, Moukalled, Aoun, Hajj, Shaban, and
  El-Hajj}]{BALY2017266}
Ramy Baly, Georges El-Khoury, Rawan Moukalled, Rita Aoun, Hazem Hajj,
  Khaled~Bashir Shaban, and Wassim El-Hajj. 2017.
\newblock \href {https://doi.org/10.1016/j.procs.2017.10.118} {Comparative
  evaluation of sentiment analysis methods across {Arabic} dialects}.
\newblock \emph{Procedia Computer Science}, 117:266--273.
\newblock Arabic Computational Linguistics.

\bibitem[{Banerjee et~al.(2025)Banerjee, Chatterjee, Kumar, Layek, Agrawal,
  Hazra, and Mukherjee}]{banerjee2025attributional}
Somnath Banerjee, Pratyush Chatterjee, Shanu Kumar, Sayan Layek, Parag Agrawal,
  Rima Hazra, and Animesh Mukherjee. 2025.
\newblock Attributional safety failures in large language models under
  code-mixed perturbations.
\newblock \emph{arXiv preprint arXiv:2505.14469}.

\bibitem[{Bansal et~al.(2020{\natexlab{a}})Bansal, Mukherjee, Satpal, and
  Mehta}]{bansal20_interspeech}
Shubham Bansal, Arijit Mukherjee, Sandeepkumar Satpal, and Rupeshkumar Mehta.
  2020{\natexlab{a}}.
\newblock \href {https://doi.org/10.21437/Interspeech.2020-2654} {{On Improving
  Code Mixed Speech Synthesis with Mixlingual Grapheme-to-Phoneme Model}}.
\newblock In \emph{{Interspeech 2020}}, pages 2957--2961.

\bibitem[{Bansal et~al.(2020{\natexlab{b}})Bansal, Garimella, Suhane, Patro,
  and Mukherjee}]{bansal-etal-2020-code}
Srijan Bansal, Vishal Garimella, Ayush Suhane, Jasabanta Patro, and Animesh
  Mukherjee. 2020{\natexlab{b}}.
\newblock \href {https://doi.org/10.18653/v1/2020.acl-main.96} {Code-switching
  patterns can be an effective route to improve performance of downstream {NLP}
  applications: A case study of humour, sarcasm and hate speech detection}.
\newblock In \emph{Proceedings of the 58th Annual Meeting of the Association
  for Computational Linguistics}, pages 1018--1023, Online. Association for
  Computational Linguistics.

\bibitem[{Barik et~al.(2019)Barik, Mahendra, and
  Adriani}]{barik-etal-2019-normalization}
Anab~Maulana Barik, Rahmad Mahendra, and Mirna Adriani. 2019.
\newblock \href {https://doi.org/10.18653/v1/D19-5554} {Normalization of
  {I}ndonesian-{E}nglish code-mixed {T}witter data}.
\newblock In \emph{Proceedings of the 5th Workshop on Noisy User-generated Text
  (W-NUT 2019)}, pages 417--424, Hong Kong, China. Association for
  Computational Linguistics.

\bibitem[{Betke and Wilson(2022)}]{betke2022scaling}
Margrit Betke and Rc~Wilson. 2022.
\newblock Scaling frame analysis to genuinely low-resource languages: A case
  study in {Swahili} and {Tamil}.
\newblock \emph{Bull. Comput. Data Sci.}, 3(1):11--21.

\bibitem[{Bharathi and Silvia(2021)}]{b-a-2021-ssncse}
B.~Bharathi and Agnusimmaculate Silvia. 2021.
\newblock \href {https://aclanthology.org/2021.dravidianlangtech-1.45/}
  {{SSNCSE}{\_}{NLP}@{D}ravidian{L}ang{T}ech-{EACL}2021: Offensive language
  identification on multilingual code mixing text}.
\newblock In \emph{Proceedings of the First Workshop on Speech and Language
  Technologies for Dravidian Languages}, pages 313--318, Kyiv. Association for
  Computational Linguistics.

\bibitem[{Bhat et~al.(2016)Bhat, Choudhury, and Bali}]{bhat2016grammatical}
Gayatri Bhat, Monojit Choudhury, and Kalika Bali. 2016.
\newblock Grammatical constraints on intra-sentential code-switching: From
  theories to working models.
\newblock \emph{arXiv preprint arXiv:1612.04538}.

\bibitem[{Bhat et~al.(2017)Bhat, Bhat, Shrivastava, and
  Sharma}]{bhat-etal-2017-joining}
Irshad Bhat, Riyaz~A. Bhat, Manish Shrivastava, and Dipti Sharma. 2017.
\newblock \href {https://aclanthology.org/E17-2052/} {Joining hands: Exploiting
  monolingual treebanks for parsing of code-mixing data}.
\newblock In \emph{Proceedings of the 15th Conference of the {E}uropean Chapter
  of the Association for Computational Linguistics: Volume 2, Short Papers},
  pages 324--330, Valencia, Spain. Association for Computational Linguistics.

\bibitem[{Bhat et~al.(2018)Bhat, Bhat, Shrivastava, and
  Sharma}]{bhat-etal-2018-universal}
Irshad Bhat, Riyaz~A. Bhat, Manish Shrivastava, and Dipti Sharma. 2018.
\newblock \href {https://doi.org/10.18653/v1/N18-1090} {{U}niversal
  {D}ependency parsing for {H}indi-{E}nglish code-switching}.
\newblock In \emph{Proceedings of the 2018 Conference of the North {A}merican
  Chapter of the Association for Computational Linguistics: Human Language
  Technologies, Volume 1 (Long Papers)}, pages 987--998, New Orleans,
  Louisiana. Association for Computational Linguistics.

\bibitem[{Bhattacharya et~al.(2024)Bhattacharya, Lin, Chen, and
  Hirschberg}]{bhattacharya24_interspeech}
Debasmita Bhattacharya, Eleanor Lin, Run Chen, and Julia Hirschberg. 2024.
\newblock \href {https://doi.org/10.21437/Interspeech.2024-1224} {{Switching
  Tongues, Sharing Hearts: Identifying the Relationship between Empathy and
  Code-switching in Speech}}.
\newblock In \emph{{Interspeech 2024}}, pages 492--496.

\bibitem[{Bhogale et~al.(2026{\natexlab{a}})Bhogale, Dhir, Walecha, Kaur,
  Chhabra, Pareek, Sidh, Manik, Jain, Singh et~al.}]{bhogale2026voice}
Kaushal Bhogale, Manas Dhir, Amritansh Walecha, Manmeet Kaur, Vanshika Chhabra,
  Aaditya Pareek, Hanuman Sidh, Mahima Manik, Sagar Jain, Bhaskar Singh, et~al.
  2026{\natexlab{a}}.
\newblock Voice of {India}: A large-scale benchmark for real-world speech
  recognition in {India}.
\newblock \emph{arXiv preprint arXiv:2604.19151}.

\bibitem[{Bhogale et~al.(2026{\natexlab{b}})Bhogale, Javed, John, Rathi,
  Padmanaban, Parasa, and Khapra}]{11464888}
Kaushal~Santosh Bhogale, Tahir Javed, Greeshma~Susan John, Dhruv Rathi,
  Akshayasree Padmanaban, Niharika Parasa, and Mitesh~M. Khapra.
  2026{\natexlab{b}}.
\newblock \href {https://doi.org/10.1109/ICASSP55912.2026.11464888} {Towards
  orthographically-informed evaluation of speech recognition systems for
  {Indian} languages}.
\newblock In \emph{ICASSP 2026 - 2026 IEEE International Conference on
  Acoustics, Speech and Signal Processing (ICASSP)}, pages 17092--17096.

\bibitem[{Bhowmick et~al.(2023)Bhowmick, Ganguli, Paul, Paul, and
  Sil}]{10.1145/3606695}
Rajat~Subhra Bhowmick, Isha Ganguli, Ananya Paul, Jayanta Paul, and Jaya Sil.
  2023.
\newblock \href {https://doi.org/10.1145/3606695} {Improving {Indic} code-mixed
  to monolingual translation using mixed script augmentation, generation \&
  transfer learning}.
\newblock \emph{ACM Trans. Asian Low-Resour. Lang. Inf. Process.}
\newblock Just Accepted.

\bibitem[{Biswas et~al.(2025)Biswas, Shevelev, Abdaoui, Tyagi, and
  Boumadane}]{biswas25_interspeech}
Astik Biswas, Oleg Shevelev, Amine Abdaoui, Vivek Tyagi, and Abdelmoumene
  Boumadane. 2025.
\newblock \href {https://doi.org/10.21437/Interspeech.2025-447} {{Adapting
  Whisper for low-resource Hindi-English Code-Mix speech with on-the-fly
  Augmentation \& LLM-Synthesised Data}}.
\newblock In \emph{{Interspeech 2025}}, pages 4293--4297.

\bibitem[{Blevins(2026)}]{blevins-2026-systematic}
Margaret Blevins. 2026.
\newblock \href {https://doi.org/10.63317/3bv9dmxr24p6} {Systematic
  normalization of spoken mixed-language, mixed-dialect data}.
\newblock In \emph{Proceedings of the First Workshop on Dialects in {NLP} {---}
  A Resource Perspective}, pages 58--69, Palma de Mallorca. Association for
  Computational Linguistics.

\bibitem[{Bougares et~al.(2025)Bougares, Mdhaffar, Elleuch, and
  Est{\`e}ve}]{bougares-etal-2025-tedxtn}
Fethi Bougares, Salima Mdhaffar, Haroun Elleuch, and Yannick Est{\`e}ve. 2025.
\newblock \href {https://doi.org/10.18653/v1/2025.arabicnlp-main.22}
  {{TED}x{TN}: A three-way speech translation corpus for code-switched
  {T}unisian {A}rabic - {E}nglish}.
\newblock In \emph{Proceedings of The Third Arabic Natural Language Processing
  Conference}, pages 278--287, Suzhou, China. Association for Computational
  Linguistics.

\bibitem[{Burchell et~al.(2024)Burchell, Birch, Thompson, and
  Heafield}]{burchell-etal-2024-code}
Laurie Burchell, Alexandra Birch, Robert Thompson, and Kenneth Heafield. 2024.
\newblock \href {https://doi.org/10.18653/v1/2024.eacl-long.38} {Code-switched
  language identification is harder than you think}.
\newblock In \emph{Proceedings of the 18th Conference of the European Chapter
  of the Association for Computational Linguistics (Volume 1: Long Papers)},
  pages 646--658, St. Julian{'}s, Malta. Association for Computational
  Linguistics.

\bibitem[{Cao et~al.(2020)Cao, Liu, Wu, Kang, Liu, Wu, Liu, Su, Yu, and
  Meng}]{9053094}
Yuewen Cao, Songxiang Liu, Xixin Wu, Shiyin Kang, Peng Liu, Zhiyong Wu, Xunying
  Liu, Dan Su, Dong Yu, and Helen Meng. 2020.
\newblock \href {https://doi.org/10.1109/ICASSP40776.2020.9053094}
  {Code-switched speech synthesis using bilingual phonetic posteriorgram with
  only monolingual corpora}.
\newblock In \emph{ICASSP 2020 - 2020 IEEE International Conference on
  Acoustics, Speech and Signal Processing (ICASSP)}, pages 7619--7623.

\bibitem[{Cao et~al.(2019)Cao, Wu, Liu, Yu, Li, Wu, Liu, and Meng}]{8682927}
Yuewen Cao, Xixin Wu, Songxiang Liu, Jianwei Yu, Xu~Li, Zhiyong Wu, Xunying
  Liu, and Helen Meng. 2019.
\newblock \href {https://doi.org/10.1109/ICASSP.2019.8682927} {End-to-end
  code-switched {TTS} with mix of monolingual recordings}.
\newblock In \emph{ICASSP 2019 - 2019 IEEE International Conference on
  Acoustics, Speech and Signal Processing (ICASSP)}, pages 6935--6939.

\bibitem[{{\c{C}}etino{\u{g}}lu(2017)}]{cetinoglu-2017-code}
{\"O}zlem {\c{C}}etino{\u{g}}lu. 2017.
\newblock \href {https://doi.org/10.18653/v1/W17-0804} {A code-switching corpus
  of {T}urkish-{G}erman conversations}.
\newblock In \emph{Proceedings of the 11th Linguistic Annotation Workshop},
  pages 34--40, Valencia, Spain. Association for Computational Linguistics.

\bibitem[{{\c{C}}etino{\u{g}}lu et~al.(2016){\c{C}}etino{\u{g}}lu, Schulz, and
  Vu}]{cetinoglu-etal-2016-challenges}
{\"O}zlem {\c{C}}etino{\u{g}}lu, Sarah Schulz, and Ngoc~Thang Vu. 2016.
\newblock \href {https://doi.org/10.18653/v1/W16-5801} {Challenges of
  computational processing of code-switching}.
\newblock In \emph{Proceedings of the Second Workshop on Computational
  Approaches to Code Switching}, pages 1--11, Austin, Texas. Association for
  Computational Linguistics.

\bibitem[{Chai et~al.(2025)Chai, Yang, Sun, Guo, Liu, Wang, Liang, Bai, Li,
  Peng et~al.}]{chai2025xcot}
Linzheng Chai, Jian Yang, Tao Sun, Hongcheng Guo, Jiaheng Liu, Bing Wang,
  Xinnian Liang, Jiaqi Bai, Tongliang Li, Qiyao Peng, et~al. 2025.
\newblock xcot: Cross-lingual instruction tuning for cross-lingual
  chain-of-thought reasoning.
\newblock In \emph{Proceedings of the AAAI Conference on Artificial
  Intelligence}, volume~39, pages 23550--23558.

\bibitem[{Chakravarthi et~al.(2022)Chakravarthi, Hande, Ponnusamy, Kumaresan,
  and Priyadharshini}]{CHAKRAVARTHI2022100119}
Bharathi~Raja Chakravarthi, Adeep Hande, Rahul Ponnusamy, Prasanna~Kumar
  Kumaresan, and Ruba Priyadharshini. 2022.
\newblock \href {https://doi.org/10.1016/j.jjimei.2022.100119} {How can we
  detect homophobia and transphobia? experiments in a multilingual code-mixed
  setting for social media governance}.
\newblock \emph{International Journal of Information Management Data Insights},
  2(2):100119.

\bibitem[{Chandu et~al.(2018)Chandu, Manzini, Singh, and
  Black}]{chandu-etal-2018-language}
Khyathi Chandu, Thomas Manzini, Sumeet Singh, and Alan~W. Black. 2018.
\newblock \href {https://doi.org/10.18653/v1/W18-3211} {Language informed
  modeling of code-switched text}.
\newblock In \emph{Proceedings of the Third Workshop on Computational
  Approaches to Linguistic Code-Switching}, pages 92--97, Melbourne, Australia.
  Association for Computational Linguistics.

\bibitem[{Chang et~al.(2019)Chang, Chuang, and Lee}]{chang19_interspeech}
Ching-Ting Chang, Shun-Po Chuang, and Hung-Yi Lee. 2019.
\newblock \href {https://doi.org/10.21437/Interspeech.2019-3214}
  {{Code-Switching Sentence Generation by Generative Adversarial Networks and
  its Application to Data Augmentation}}.
\newblock In \emph{{Interspeech 2019}}, pages 554--558.

\bibitem[{Changsong et~al.(2023)Changsong, Nga, Qi, and Siong}]{10390242}
Liu Changsong, Ho~Thi Nga, Yip~Jia Qi, and Chng~Eng Siong. 2023.
\newblock \href {https://doi.org/10.1109/ICAICTA59291.2023.10390242}
  {{Transformer}-based punctuation restoration models for {Indonesian} with
  {English} codeswitching speech transcripts}.
\newblock In \emph{2023 10th International Conference on Advanced Informatics:
  Concept, Theory and Application (ICAICTA)}, pages 1--6.

\bibitem[{Chatterjee et~al.(2022)Chatterjee, Sharma, Raj, and
  Ekbal}]{chatterjee-etal-2022-pacman}
Arindam Chatterjee, Chhavi Sharma, Ayush Raj, and Asif Ekbal. 2022.
\newblock \href {https://aclanthology.org/2022.icon-main.29/}
  {{PACMAN}:{PA}rallel {C}ode{M}ixed d{A}ta generatio{N} for {POS} tagging}.
\newblock In \emph{Proceedings of the 19th International Conference on Natural
  Language Processing (ICON)}, pages 234--244, New Delhi, India. Association
  for Computational Linguistics.

\bibitem[{Chen et~al.(2016)Chen, Pan, Zhao, and Yan}]{e99-d_10_2554}
Mengzhe Chen, Jielin Pan, Qingwei Zhao, and Yonghong Yan. 2016.
\newblock \href {https://doi.org/10.1587/transinf.2016SLL0004} {Multi-task
  learning in deep neural networks for {Mandarin}-{English} code-mixing speech
  recognition}.
\newblock \emph{IEICE TRANSACTIONS on Information}, E99-D(10):2554--2557.

\bibitem[{Cheng et~al.(2025)Cheng, Hua, Poh, and Ming}]{11255834}
Wilson Pang~Vui Cheng, Lim~Kong Hua, Yu~Yong Poh, and Lim~Tong Ming. 2025.
\newblock \href {https://doi.org/10.1109/ICCBE65177.2025.11255834} {{CS} in
  automatic speech recognition: A review of advances and challenges}.
\newblock In \emph{2025 IEEE International Conference on Computation, Big-Data
  and Engineering (ICCBE)}, pages 349--354.

\bibitem[{Cheng et~al.(2023)Cheng, Xu, Yao, Zhu, Li, Li, and
  Zou}]{cheng23b_interspeech}
Xuxin Cheng, Wanshi Xu, Ziyu Yao, Zhihong Zhu, Yaowei Li, Hongxiang Li, and
  Yuexian Zou. 2023.
\newblock \href {https://doi.org/10.21437/Interspeech.2023-46} {{FC-MTLF: A
  Fine- and Coarse-grained Multi-Task Learning Framework for Cross-Lingual
  Spoken Language Understanding}}.
\newblock In \emph{{Interspeech 2023}}, pages 690--694.

\bibitem[{Chou et~al.(2025)Chou, Hsu, Chung, Tseng, Cheng, Fu, Huang, and
  Lee}]{11434800}
Cheng-Kang Chou, Chan-Jan Hsu, Ho-Lam Chung, Liang-Hsuan Tseng, Hsi-Chun Cheng,
  Yu-Kuan Fu, Kuan~Po Huang, and Hung-Yi Lee. 2025.
\newblock \href {https://doi.org/10.1109/ASRU65441.2025.11434800} {A
  self-refining framework for enhancing {ASR} using {TTS}-synthesized data}.
\newblock In \emph{2025 IEEE Automatic Speech Recognition and Understanding
  Workshop (ASRU)}, pages 1--8.

\bibitem[{Choudhary et~al.(2023)Choudhary, Singh, Bindlish, and
  Shrivastava}]{10.1007/978-3-031-23804-8_9}
Nurendra Choudhary, Rajat Singh, Ishita Bindlish, and Manish Shrivastava. 2023.
\newblock Sentiment analysis of code-mixed languages leveraging resource rich
  languages.
\newblock In \emph{Computational Linguistics and Intelligent Text Processing},
  pages 104--114, Cham. Springer Nature Switzerland.

\bibitem[{Chowdhury et~al.(2020)Chowdhury, Samih, Eldesouki, and
  Ali}]{chowdhury20c_interspeech}
Shammur~A. Chowdhury, Younes Samih, Mohamed Eldesouki, and Ahmed Ali. 2020.
\newblock \href {https://doi.org/10.21437/Interspeech.2020-2271} {{Effects of
  Dialectal Code-Switching on Speech Modules: A Study Using Egyptian Arabic
  Broadcast Speech}}.
\newblock In \emph{{Interspeech 2020}}, pages 2382--2386.

\bibitem[{Chowdhury et~al.(2021)Chowdhury, Hussein, Abdelali, and
  Ali}]{chowdhury21_interspeech}
Shammur~Absar Chowdhury, Amir Hussein, Ahmed Abdelali, and Ahmed Ali. 2021.
\newblock \href {https://doi.org/10.21437/Interspeech.2021-1809} {{Towards One
  Model to Rule All: Multilingual Strategy for Dialectal Code-Switching Arabic
  ASR}}.
\newblock In \emph{{Interspeech 2021}}, pages 2466--2470.

\bibitem[{Chua et~al.(2024)Chua, Ghazi, Huang, Kamath, Kumar, Manurangsi,
  Sinha, Xie, and Zhang}]{chua2024crosslingual}
Lynn Chua, Badih Ghazi, Yangsibo Huang, Pritish Kamath, Ravi Kumar, Pasin
  Manurangsi, Amer Sinha, Chulin Xie, and Chiyuan Zhang. 2024.
\newblock Crosslingual capabilities and knowledge barriers in multilingual
  large language models.
\newblock \emph{arXiv preprint arXiv:2406.16135}.

\bibitem[{Chua et~al.(2023)Chua, Liu, Garcia, Woon, Wong, Zhang, Khudanpur,
  Khong, Dauwels, and Styles}]{chua23_interspeech}
Victoria Y.~H. Chua, Hexin Liu, Leibny~Paola Garcia, Fei~Ting Woon, Jinyi Wong,
  Xiangyu Zhang, Sanjeev Khudanpur, Andy W.~H. Khong, Justin Dauwels, and
  Suzy~J. Styles. 2023.
\newblock \href {https://doi.org/10.21437/Interspeech.2023-1446} {{MERLIon CCS
  Challenge: A English-Mandarin code-switching child-directed speech corpus for
  language identification and diarization}}.
\newblock In \emph{{Interspeech 2023}}, pages 4109--4113.

\bibitem[{Chuang et~al.(2020)Chuang, Sung, and Lee}]{9053775}
Shun-Po Chuang, Tzu-Wei Sung, and Hung-yi Lee. 2020.
\newblock \href {https://doi.org/10.1109/ICASSP40776.2020.9053775} {Training
  code-switching language model with monolingual data}.
\newblock In \emph{ICASSP 2020 - 2020 IEEE International Conference on
  Acoustics, Speech and Signal Processing (ICASSP)}, pages 7949--7953.

\bibitem[{Colombo et~al.(2021)Colombo, Chapuis, Labeau, and
  Clavel}]{colombo-etal-2021-code}
Pierre Colombo, Emile Chapuis, Matthieu Labeau, and Chlo{\'e} Clavel. 2021.
\newblock \href {https://doi.org/10.18653/v1/2021.emnlp-main.656}
  {Code-switched inspired losses for spoken dialog representations}.
\newblock In \emph{Proceedings of the 2021 Conference on Empirical Methods in
  Natural Language Processing}, pages 8320--8337, Online and Punta Cana,
  Dominican Republic. Association for Computational Linguistics.

\bibitem[{Dabre(2022)}]{dabre-2022-nict}
Raj Dabre. 2022.
\newblock \href {https://doi.org/10.18653/v1/2022.wmt-1.111} {{NICT} at
  {M}ix{MT} 2022: Synthetic code-mixed pre-training and multi-way fine-tuning
  for {H}inglish{--}{E}nglish translation}.
\newblock In \emph{Proceedings of the Seventh Conference on Machine Translation
  (WMT)}, pages 1122--1125, Abu Dhabi, United Arab Emirates (Hybrid).
  Association for Computational Linguistics.

\bibitem[{Dalmia et~al.(2021)Dalmia, Liu, Ronanki, and Kirchhoff}]{9413562}
Siddharth Dalmia, Yuzong Liu, Srikanth Ronanki, and Katrin Kirchhoff. 2021.
\newblock \href {https://doi.org/10.1109/ICASSP39728.2021.9413562}
  {{Transformer}-transducers for code-switched speech recognition}.
\newblock In \emph{ICASSP 2021 - 2021 IEEE International Conference on
  Acoustics, Speech and Signal Processing (ICASSP)}, pages 5859--5863.

\bibitem[{Darwish(2014)}]{darwish-2014-arabizi}
Kareem Darwish. 2014.
\newblock \href {https://doi.org/10.3115/v1/W14-3629} {{A}rabizi detection and
  conversion to {A}rabic}.
\newblock In \emph{Proceedings of the {EMNLP} 2014 Workshop on {A}rabic Natural
  Language Processing ({ANLP})}, pages 217--224, Doha, Qatar. Association for
  Computational Linguistics.

\bibitem[{Das et~al.(2026)Das, Parmar, Ramnath, and Verma}]{das2026indi}
Avisha Das, Mihir Parmar, Mohana Ramnath, and Pulkit Verma. 2026.
\newblock Indi-{RomCoM}: Code-mixed benchmark for evaluating {LLMs} on
  romanized {Indic}-{English} instructions.
\newblock \emph{arXiv preprint arXiv:2606.30790}.

\bibitem[{Das et~al.(2022{\natexlab{a}})Das, Banerjee, Saha, and
  Mukherjee}]{das-etal-2022-hate-speech}
Mithun Das, Somnath Banerjee, Punyajoy Saha, and Animesh Mukherjee.
  2022{\natexlab{a}}.
\newblock \href {https://doi.org/10.18653/v1/2022.aacl-main.23} {Hate speech
  and offensive language detection in {B}engali}.
\newblock In \emph{Proceedings of the 2nd Conference of the Asia-Pacific
  Chapter of the Association for Computational Linguistics and the 12th
  International Joint Conference on Natural Language Processing (Volume 1: Long
  Papers)}, pages 286--296, Online only. Association for Computational
  Linguistics.

\bibitem[{Das et~al.(2023)Das, Ranjan, Pathak, and
  Jyothi}]{das-etal-2023-improving}
Richeek Das, Sahasra Ranjan, Shreya Pathak, and Preethi Jyothi. 2023.
\newblock \href {https://doi.org/10.18653/v1/2023.acl-long.66} {Improving
  pretraining techniques for code-switched {NLP}}.
\newblock In \emph{Proceedings of the 61st Annual Meeting of the Association
  for Computational Linguistics (Volume 1: Long Papers)}, pages 1176--1191,
  Toronto, Canada. Association for Computational Linguistics.

\bibitem[{Das et~al.(2022{\natexlab{b}})Das, Williams, and
  Lai}]{das2022analysis}
Shuvayanti Das, Jennifer Williams, and Catherine Lai. 2022{\natexlab{b}}.
\newblock Analysis of voice conversion and code-switching synthesis using
  {VQ}-{VAE}.
\newblock \emph{arXiv preprint arXiv:2203.14640}.

\bibitem[{Dash et~al.(2025)Dash, Babu, Singaravel, and
  Balasubramanian}]{dash2025generative}
Puspita Dash, Sruthi Babu, Logeswari Singaravel, and Devadarshini
  Balasubramanian. 2025.
\newblock Generative {AI}-powered multilingual {ASR} for seamless
  language-mixing transcriptions.
\newblock \emph{Journal of Electrical Systems and Information Technology},
  12(1):42.

\bibitem[{Dewangan and Maurya(2026)}]{dewangan-maurya-2026-claocs}
Lipika Dewangan and Chandresh~Kumar Maurya. 2026.
\newblock \href {https://doi.org/10.18653/v1/2026.acl-long.2063}
  {{CLAOCS}-{TX}: Cross-lingual triplet extraction with aspect-opinion-aware
  code-switched prompting and {LLM}-guided contrastive distillation}.
\newblock In \emph{Proceedings of the 64th Annual Meeting of the {A}ssociation
  for {C}omputational {L}inguistics (Volume 1: Long Papers)}, pages
  44561--44577, San Diego, California, United States. Association for
  Computational Linguistics.

\bibitem[{Dhar et~al.(2018)Dhar, Kumar, and
  Shrivastava}]{dhar-etal-2018-enabling}
Mrinal Dhar, Vaibhav Kumar, and Manish Shrivastava. 2018.
\newblock \href {https://aclanthology.org/W18-3817/} {Enabling code-mixed
  translation: Parallel corpus creation and {MT} augmentation approach}.
\newblock In \emph{Proceedings of the First Workshop on Linguistic Resources
  for Natural Language Processing}, pages 131--140, Santa Fe, New Mexico, USA.
  Association for Computational Linguistics.

\bibitem[{Dhavale et~al.(2023)Dhavale, Dhavale, and Dhavale}]{10450050}
Aditi~Vikrant Dhavale, Amit~Vikrant Dhavale, and Sunita~Vikrant Dhavale. 2023.
\newblock \href {https://doi.org/10.1109/PuneCon58714.2023.10450050} {An
  intelligent software tool for audio signal-based {Hinglish} to {English}
  language translation}.
\newblock In \emph{2023 IEEE Pune Section International Conference (PuneCon)},
  pages 1--6.

\bibitem[{Dhawan et~al.(2023)Dhawan, Rekesh, and Ginsburg}]{dhawan2023unified}
Kunal Dhawan, Dima Rekesh, and Boris Ginsburg. 2023.
\newblock \href {https://aclanthology.org/2023.calcs-1.7/} {Unified model for
  code-switching speech recognition and language identification based on
  concatenated tokenizer}.
\newblock In \emph{Proceedings of the 6th Workshop on Computational Approaches
  to Linguistic Code-Switching}, pages 74--82, Singapore. Association for
  Computational Linguistics.

\bibitem[{Di et~al.(2023)Di, Song, Zhang, Zhang, and Wang}]{10.1145/3575803}
Donglin Di, Xianyang Song, Weinan Zhang, Yue Zhang, and Fanglin Wang. 2023.
\newblock \href {https://doi.org/10.1145/3575803} {Building dialogue
  understanding models for low-resource language {Indonesian} from scratch}.
\newblock \emph{ACM Trans. Asian Low-Resour. Lang. Inf. Process.}, 22(4).

\bibitem[{Ding et~al.(2024)Ding, Jia, Xu, Xi, Lai, and Ginsburg}]{10832233}
Wen Ding, Fei Jia, Hainan Xu, Yu~Xi, Junjie Lai, and Boris Ginsburg. 2024.
\newblock \href {https://doi.org/10.1109/SLT61566.2024.10832233} {Romanization
  encoding for multilingual {ASR}}.
\newblock In \emph{2024 IEEE Spoken Language Technology Workshop (SLT)}, pages
  468--475.

\bibitem[{Do et~al.(2024)Do, Lee, and Hwang}]{do-etal-2024-contrastivemix}
Junggeun Do, Jaeseong Lee, and Seung-won Hwang. 2024.
\newblock \href {https://doi.org/10.18653/v1/2024.naacl-short.17}
  {{C}ontrastive{M}ix: Overcoming code-mixing dilemma in cross-lingual transfer
  for information retrieval}.
\newblock In \emph{Proceedings of the 2024 Conference of the North American
  Chapter of the Association for Computational Linguistics: Human Language
  Technologies (Volume 2: Short Papers)}, pages 197--204, Mexico City, Mexico.
  Association for Computational Linguistics.

\bibitem[{Do{\u{g}}ru{\"o}z et~al.(2021)Do{\u{g}}ru{\"o}z, Sitaram, Bullock,
  and Toribio}]{dogruoz-etal-2021-survey}
A.~Seza Do{\u{g}}ru{\"o}z, Sunayana Sitaram, Barbara~E. Bullock, and
  Almeida~Jacqueline Toribio. 2021.
\newblock \href {https://doi.org/10.18653/v1/2021.acl-long.131} {A survey of
  code-switching: Linguistic and social perspectives for language
  technologies}.
\newblock In \emph{Proceedings of the 59th Annual Meeting of the Association
  for Computational Linguistics and the 11th International Joint Conference on
  Natural Language Processing (Volume 1: Long Papers)}, pages 1654--1666,
  Online. Association for Computational Linguistics.

\bibitem[{Duong et~al.(2017)Duong, Afshar, Estival, Pink, Cohen, and
  Johnson}]{duong-etal-2017-multilingual-semantic}
Long Duong, Hadi Afshar, Dominique Estival, Glen Pink, Philip Cohen, and Mark
  Johnson. 2017.
\newblock \href {https://doi.org/10.18653/v1/K17-1038} {Multilingual semantic
  parsing and code-switching}.
\newblock In \emph{Proceedings of the 21st Conference on Computational Natural
  Language Learning ({C}o{NLL} 2017)}, pages 379--389, Vancouver, Canada.
  Association for Computational Linguistics.

\bibitem[{Dutta et~al.(2025)Dutta, Ranjan, Jain, Singh, and Vatsa}]{10889964}
Bikash Dutta, Rishabh Ranjan, Akshat Jain, Richa Singh, and Mayank Vatsa. 2025.
\newblock \href {https://doi.org/10.1109/ICASSP49660.2025.10889964} {Can
  {RAG}-driven enhancements amplify audio {LLMs} for low-resource languages?}
\newblock In \emph{ICASSP 2025 - 2025 IEEE International Conference on
  Acoustics, Speech and Signal Processing (ICASSP)}, pages 1--5.

\bibitem[{Emond et~al.(2018)Emond, Ramabhadran, Roark, Moreno, and
  Ma}]{8639699}
Jesse Emond, Bhuvana Ramabhadran, Brian Roark, Pedro Moreno, and Min Ma. 2018.
\newblock \href {https://doi.org/10.1109/SLT.2018.8639699} {Transliteration
  based approaches to improve code-switched speech recognition performance}.
\newblock In \emph{2018 IEEE Spoken Language Technology Workshop (SLT)}, pages
  448--455.

\bibitem[{Emran et~al.(2024)Emran, Ganguly, Puspo, Raihan, and
  Goswami}]{emran2024masontigers}
Al~Nahian~Bin Emran, Amrita Ganguly, Sadiya Sayara~Chowdhury Puspo, Nishat
  Raihan, and Dhiman Goswami. 2024.
\newblock {MasonTigers} at {SemEval}-2024 task 10: Emotion discovery and flip
  reasoning in conversation with ensemble of {Transformers} and prompting.
\newblock \emph{arXiv preprint arXiv:2407.00581}.

\bibitem[{Fan et~al.(2023)Fan, Dong, Shen, Liang, Zhang, Lu, and
  Ma}]{fan2023language}
Zhiyun Fan, Linhao Dong, Chen Shen, Zhenlin Liang, Jun Zhang, Lu~Lu, and Zejun
  Ma. 2023.
\newblock Language-specific acoustic boundary learning for mandarin-english
  code-switching speech recognition.
\newblock \emph{arXiv preprint arXiv:2306.05279}.

\bibitem[{Farooq and Saz(2025)}]{11204332}
Muhammad~Umar Farooq and Oscar Saz. 2025.
\newblock \href {https://doi.org/10.1109/MLSP62443.2025.11204332} {{SAGE}:
  Spliced-audio generated data for enhancing foundational models in
  low-resource {Arabic}-{English} code-switched speech recognition}.
\newblock In \emph{2025 IEEE 35th International Workshop on Machine Learning
  for Signal Processing (MLSP)}, pages 1--6.

\bibitem[{Fehle et~al.(2026)Fehle, Hellwig, Kruschwitz, and
  Wolff}]{fehle2026zero}
Jakob Fehle, Nils~Constantin Hellwig, Udo Kruschwitz, and Christian Wolff.
  2026.
\newblock Zero-shot to full-resource: Cross-lingual transfer strategies for
  aspect-based sentiment analysis.
\newblock \emph{arXiv preprint arXiv:2604.26619}.

\bibitem[{Feng et~al.(2022)Feng, Li, and Koehn}]{feng-etal-2022-toward}
Yukun Feng, Feng Li, and Philipp Koehn. 2022.
\newblock \href {https://doi.org/10.18653/v1/2022.emnlp-main.400} {Toward the
  limitation of code-switching in cross-lingual transfer}.
\newblock In \emph{Proceedings of the 2022 Conference on Empirical Methods in
  Natural Language Processing}, pages 5966--5971, Abu Dhabi, United Arab
  Emirates. Association for Computational Linguistics.

\bibitem[{Fetahu et~al.(2021)Fetahu, Fang, Rokhlenko, and
  Malmasi}]{10.1145/3404835.3463102}
Besnik Fetahu, Anjie Fang, Oleg Rokhlenko, and Shervin Malmasi. 2021.
\newblock \href {https://doi.org/10.1145/3404835.3463102} {Gazetteer enhanced
  named entity recognition for code-mixed web queries}.
\newblock In \emph{Proceedings of the 44th International ACM SIGIR Conference
  on Research and Development in Information Retrieval}, SIGIR '21, page
  1677–1681, New York, NY, USA. Association for Computing Machinery.

\bibitem[{Gahoi et~al.(2022)Gahoi, Duneja, Padhi, Mangale, Rajput, Kamble,
  Sharma, and Varma}]{gahoi-etal-2022-gui}
Akshat Gahoi, Jayant Duneja, Anshul Padhi, Shivam Mangale, Saransh Rajput,
  Tanvi Kamble, Dipti Sharma, and Vasudev Varma. 2022.
\newblock \href {https://doi.org/10.18653/v1/2022.wmt-1.112} {Gui at {M}ix{MT}
  2022 : {E}nglish-{H}inglish : An {MT} approach for translation of code mixed
  data}.
\newblock In \emph{Proceedings of the Seventh Conference on Machine Translation
  (WMT)}, pages 1126--1130, Abu Dhabi, United Arab Emirates (Hybrid).
  Association for Computational Linguistics.

\bibitem[{Garg et~al.(2018{\natexlab{a}})Garg, Parekh, and
  Jyothi}]{garg-etal-2018-code}
Saurabh Garg, Tanmay Parekh, and Preethi Jyothi. 2018{\natexlab{a}}.
\newblock \href {https://doi.org/10.18653/v1/D18-1346} {Code-switched language
  models using dual {RNN}s and same-source pretraining}.
\newblock In \emph{Proceedings of the 2018 Conference on Empirical Methods in
  Natural Language Processing}, pages 3078--3083, Brussels, Belgium.
  Association for Computational Linguistics.

\bibitem[{Garg et~al.(2018{\natexlab{b}})Garg, Parekh, and
  Jyothi}]{garg18_interspeech}
Saurabh Garg, Tanmay Parekh, and Preethi Jyothi. 2018{\natexlab{b}}.
\newblock \href {https://doi.org/10.21437/Interspeech.2018-1343} {{Dual
  Language Models for Code Switched Speech Recognition}}.
\newblock In \emph{{Interspeech 2018}}, pages 2598--2602.

\bibitem[{Ghosal and Jain(2023)}]{10.1145/3576913}
Sayani Ghosal and Amita Jain. 2023.
\newblock \href {https://doi.org/10.1145/3576913} {{HateCircle} and
  unsupervised hate speech detection incorporating emotion and contextual
  semantics}.
\newblock \emph{ACM Trans. Asian Low-Resour. Lang. Inf. Process.}, 22(4).

\bibitem[{Gonen and Goldberg(2019)}]{gonen-goldberg-2019-language}
Hila Gonen and Yoav Goldberg. 2019.
\newblock \href {https://doi.org/10.18653/v1/D19-1427} {Language modeling for
  code-switching: Evaluation, integration of monolingual data, and
  discriminative training}.
\newblock In \emph{Proceedings of the 2019 Conference on Empirical Methods in
  Natural Language Processing and the 9th International Joint Conference on
  Natural Language Processing (EMNLP-IJCNLP)}, pages 4175--4185, Hong Kong,
  China. Association for Computational Linguistics.

\bibitem[{Guellil et~al.(2018{\natexlab{a}})Guellil, Adeel, Azouaou, Benali,
  Hachani, and Hussain}]{guellil-etal-2018-arabizi}
Imane Guellil, Ahsan Adeel, Faical Azouaou, Fodil Benali, Ala-eddine Hachani,
  and Amir Hussain. 2018{\natexlab{a}}.
\newblock \href {https://doi.org/10.18653/v1/W18-6249} {{A}rabizi sentiment
  analysis based on transliteration and automatic corpus annotation}.
\newblock In \emph{Proceedings of the 9th Workshop on Computational Approaches
  to Subjectivity, Sentiment and Social Media Analysis}, pages 335--341,
  Brussels, Belgium. Association for Computational Linguistics.

\bibitem[{Guellil et~al.(2018{\natexlab{b}})Guellil, Adeel, Azouaou, and
  Hussain}]{10.1007/978-3-030-00563-4_54}
Imane Guellil, Ahsan Adeel, Faical Azouaou, and Amir Hussain.
  2018{\natexlab{b}}.
\newblock {SentiALG}: Automated corpus annotation for {Algerian} sentiment
  analysis.
\newblock In \emph{Advances in Brain Inspired Cognitive Systems}, pages
  557--567, Cham. Springer International Publishing.

\bibitem[{Guo et~al.(2018)Guo, Xu, Xie, and Chng}]{guo18b_interspeech}
Pengcheng Guo, Haihua Xu, Lei Xie, and Eng~Siong Chng. 2018.
\newblock \href {https://doi.org/10.21437/Interspeech.2018-1974} {{Study of
  Semi-supervised Approaches to Improving English-Mandarin Code-Switching
  Speech Recognition}}.
\newblock In \emph{{Interspeech 2018}}, pages 1928--1932.

\bibitem[{Guo et~al.(2024)Guo, Zhang, Ding, Zhu, and Yu}]{10225623}
Zhiwei Guo, Qin Zhang, Feng Ding, Xiaogang Zhu, and Keping Yu. 2024.
\newblock \href {https://doi.org/10.1109/TCSS.2023.3298480} {A novel fake news
  detection model for context of mixed languages through multiscale
  {Transformer}}.
\newblock \emph{IEEE Transactions on Computational Social Systems},
  11(4):5079--5089.

\bibitem[{Gupta et~al.(2021{\natexlab{a}})Gupta, Vavre, and
  Sarawagi}]{gupta-etal-2021-training}
Abhirut Gupta, Aditya Vavre, and Sunita Sarawagi. 2021{\natexlab{a}}.
\newblock \href {https://doi.org/10.18653/v1/2021.naacl-main.459} {Training
  data augmentation for code-mixed translation}.
\newblock In \emph{Proceedings of the 2021 Conference of the North American
  Chapter of the Association for Computational Linguistics: Human Language
  Technologies}, pages 5760--5766, Online. Association for Computational
  Linguistics.

\bibitem[{Gupta et~al.(2021{\natexlab{b}})Gupta, Rallabandi, and
  Black}]{gupta-etal-2021-task}
Akshat Gupta, Sai~Krishna Rallabandi, and Alan~W Black. 2021{\natexlab{b}}.
\newblock \href {https://aclanthology.org/2021.dravidianlangtech-1.9/}
  {Task-specific pre-training and cross lingual transfer for sentiment analysis
  in {D}ravidian code-switched languages}.
\newblock In \emph{Proceedings of the First Workshop on Speech and Language
  Technologies for Dravidian Languages}, pages 73--79, Kyiv. Association for
  Computational Linguistics.

\bibitem[{Gupta et~al.(2024)Gupta, Bhogal, and Ghosh}]{gupta2024code}
Ayushman Gupta, Akhil Bhogal, and Kripabandhu Ghosh. 2024.
\newblock Code-mixer ya nahi: Novel approaches to measuring multilingual
  {LLMs'} code-mixing capabilities.
\newblock \emph{arXiv preprint arXiv:2410.11079}.

\bibitem[{Gupta et~al.(2020)Gupta, Ekbal, and
  Bhattacharyya}]{gupta-etal-2020-semi}
Deepak Gupta, Asif Ekbal, and Pushpak Bhattacharyya. 2020.
\newblock \href {https://doi.org/10.18653/v1/2020.findings-emnlp.206} {A
  semi-supervised approach to generate the code-mixed text using pre-trained
  encoder and transfer learning}.
\newblock In \emph{Findings of the Association for Computational Linguistics:
  EMNLP 2020}, pages 2267--2280, Online. Association for Computational
  Linguistics.

\bibitem[{Gupta et~al.(2026)Gupta, Jayarao, Dwivedi, and
  Varshney}]{gupta2026code}
Himanshu Gupta, Pratik Jayarao, Chaitanya Dwivedi, and Neeraj Varshney. 2026.
\newblock Code mixologist: A practitioner's guide to building code-mixed
  {LLMs}.
\newblock \emph{arXiv preprint arXiv:2602.11181}.

\bibitem[{Gupta et~al.(2018)Gupta, Chinnakotla, and
  Shrivastava}]{gupta-etal-2018-transliteration}
Vishal Gupta, Manoj Chinnakotla, and Manish Shrivastava. 2018.
\newblock \href {https://doi.org/10.18653/v1/W18-3205} {Transliteration better
  than translation? answering code-mixed questions over a knowledge base}.
\newblock In \emph{Proceedings of the Third Workshop on Computational
  Approaches to Linguistic Code-Switching}, pages 39--50, Melbourne, Australia.
  Association for Computational Linguistics.

\bibitem[{Haisa et~al.(2024)Haisa, Altenbek, and Li}]{haisa2024phrase}
Gulizada Haisa, Gulila Altenbek, and Wen Li. 2024.
\newblock Phrase based code-switching for cross-lingual question understanding.
\newblock \emph{Multimedia Tools and Applications}, 83(11):32159--32175.

\bibitem[{Hajbi et~al.(2024)Hajbi, Amezian, Moukhi, Korchiyne, and
  Chihab}]{HAJBI2024e02073}
Soufiane Hajbi, Omayma Amezian, Nawfal~El Moukhi, Redouan Korchiyne, and Younes
  Chihab. 2024.
\newblock \href {https://doi.org/10.1016/j.sciaf.2024.e02073} {{Moroccan}
  {Arabizi}-to-{Arabic} conversion using rule-based transliteration and
  weighted levenshtein algorithm}.
\newblock \emph{Scientific African}, 23:e02073.

\bibitem[{Hamed et~al.(2022)Hamed, Denisov, Li, Elmahdy, Abdennadher, and
  Vu}]{HAMED2022101278}
Injy Hamed, Pavel Denisov, Chia-Yu Li, Mohamed Elmahdy, Slim Abdennadher, and
  Ngoc~Thang Vu. 2022.
\newblock \href {https://doi.org/10.1016/j.csl.2021.101278} {Investigations on
  speech recognition systems for low-resource dialectal {Arabic}–{English}
  code-switching speech}.
\newblock \emph{Computer Speech \& Language}, 72:101278.

\bibitem[{Hamed et~al.(2023)Hamed, Hussein, Chellah, Chowdhury, Mubarak,
  Sitaram, Habash, and Ali}]{10023181}
Injy Hamed, Amir Hussein, Oumnia Chellah, Shammur Chowdhury, Hamdy Mubarak,
  Sunayana Sitaram, Nizar Habash, and Ahmed Ali. 2023.
\newblock \href {https://doi.org/10.1109/SLT54892.2023.10023181} {Benchmarking
  evaluation metrics for code-switching automatic speech recognition}.
\newblock In \emph{2022 IEEE Spoken Language Technology Workshop (SLT)}, pages
  999--1005.

\bibitem[{Hamed et~al.(2025)Hamed, Sabty, Abdennadher, Vu, Solorio, and
  Habash}]{hamed-etal-2025-survey}
Injy Hamed, Caroline Sabty, Slim Abdennadher, Ngoc~Thang Vu, Thamar Solorio,
  and Nizar Habash. 2025.
\newblock \href {https://aclanthology.org/2025.coling-main.307/} {A survey of
  code-switched {A}rabic {NLP}: Progress, challenges, and future directions}.
\newblock In \emph{Proceedings of the 31st International Conference on
  Computational Linguistics}, pages 4561--4585, Abu Dhabi, UAE. Association for
  Computational Linguistics.

\bibitem[{Hamed et~al.(2020)Hamed, Vu, and Abdennadher}]{hamed-etal-2020-arzen}
Injy Hamed, Ngoc~Thang Vu, and Slim Abdennadher. 2020.
\newblock \href {https://aclanthology.org/2020.lrec-1.523/} {{A}rz{E}n: A
  speech corpus for code-switched {E}gyptian {A}rabic-{E}nglish}.
\newblock In \emph{Proceedings of the Twelfth Language Resources and Evaluation
  Conference}, pages 4237--4246, Marseille, France. European Language Resources
  Association.

\bibitem[{Hande et~al.(2021)Hande, Priyadharshini, Sampath, Thamburaj,
  Chandran, and Chakravarthi}]{hande2021hope}
Adeep Hande, Ruba Priyadharshini, Anbukkarasi Sampath, Kingston~Pal Thamburaj,
  Prabakaran Chandran, and Bharathi~Raja Chakravarthi. 2021.
\newblock {Hope} speech detection in under-resourced kannada language.
\newblock \emph{arXiv preprint arXiv:2108.04616}.

\bibitem[{He and Chien(2023)}]{10317410}
Chun-Yi He and Jen-Tzung Chien. 2023.
\newblock \href {https://doi.org/10.1109/APSIPAASC58517.2023.10317410}
  {Learning adapters for code-switching speech recognition}.
\newblock In \emph{2023 Asia Pacific Signal and Information Processing
  Association Annual Summit and Conference (APSIPA ASC)}, pages 344--349.

\bibitem[{He et~al.(2012)He, Qian, Soong, and Zhao}]{he12_interspeech}
Ji~He, Yao Qian, Frank~K. Soong, and Sheng Zhao. 2012.
\newblock \href {https://doi.org/10.21437/Interspeech.2012-287} {{Turning a
  monolingual speaker into multilingual for a mixed-language TTS}}.
\newblock In \emph{{Interspeech 2012}}, pages 963--966.

\bibitem[{Hegde et~al.(2025)Hegde, Balouchzahi, Butt, Coelho, G, Kumar, D,
  H.~L., and Agrawal}]{10.1145/3734947.3735663}
Asha Hegde, Fazlourrahman Balouchzahi, Sabur Butt, Sharal Coelho, Kavya G,
  Harshitha~S Kumar, Sonith D, Shashirekha H.~L., and Ameeta Agrawal. 2025.
\newblock \href {https://doi.org/10.1145/3734947.3735663} {{CoLI}@{FIRE2024}:
  Findings of word-level code-mixed language identification in {Dravidian}
  languages}.
\newblock In \emph{Proceedings of the 16th Annual Meeting of the Forum for
  Information Retrieval Evaluation}, FIRE '24, page 7–10, New York, NY, USA.
  Association for Computing Machinery.

\bibitem[{Heredia et~al.(2026)Heredia, Labaka, Barnes, and
  Soroa}]{heredia-etal-2026-conditioning}
Maite Heredia, Gorka Labaka, Jeremy Barnes, and Aitor Soroa. 2026.
\newblock \href {https://doi.org/10.63317/443bxexszimg} {Conditioning {LLM}s to
  generate code-switched text}.
\newblock In \emph{Proceedings of the Fifteenth Language Resources and
  Evaluation Conference}, pages 8937--8953, Palma de Mallorca, Spain. ELRA
  Language Resource Association.

\bibitem[{Hidayatullah(2024)}]{10730138}
Ahmad~Fathan Hidayatullah. 2024.
\newblock \href {https://doi.org/10.1109/ICITISEE63424.2024.10730138}
  {Code-mixed sentiment analysis on {Indonesian}-{Javanese}-{English} text
  using {Transformer} models}.
\newblock In \emph{2024 8th International Conference on Information Technology,
  Information Systems and Electrical Engineering (ICITISEE)}, pages 340--345.

\bibitem[{Ho et~al.(2024)Ho, Chean, and Lim}]{10.1145/3651781.3651819}
Chuk~Fong Ho, Kae~Lun Chean, and Tong~Ming Lim. 2024.
\newblock \href {https://doi.org/10.1145/3651781.3651819} {Leveraging machine
  translation to enhance sentiment analysis on multilingual text}.
\newblock In \emph{Proceedings of the 2024 13th International Conference on
  Software and Computer Applications}, ICSCA '24, page 242–248, New York, NY,
  USA. Association for Computing Machinery.

\bibitem[{Hong et~al.(2025)Hong, Lee, Moon, and Lim}]{hong-etal-2025-migrate}
Seongtae Hong, Seungyoon Lee, Hyeonseok Moon, and Heuiseok Lim. 2025.
\newblock \href {https://aclanthology.org/2025.coling-main.617/} {{MIGRATE}:
  Cross-lingual adaptation of domain-specific {LLM}s through code-switching and
  embedding transfer}.
\newblock In \emph{Proceedings of the 31st International Conference on
  Computational Linguistics}, pages 9184--9193, Abu Dhabi, UAE. Association for
  Computational Linguistics.

\bibitem[{Hossain et~al.(2021)Hossain, Sharif, and
  Hoque}]{hossain-etal-2021-nlp-cuet}
Eftekhar Hossain, Omar Sharif, and Mohammed~Moshiul Hoque. 2021.
\newblock \href {https://aclanthology.org/2021.ltedi-1.25/}
  {{NLP}-{CUET}@{LT}-{EDI}-{EACL}2021: Multilingual code-mixed {Hope} speech
  detection using cross-lingual representation learner}.
\newblock In \emph{Proceedings of the First Workshop on Language Technology for
  Equality, Diversity and Inclusion}, pages 168--174, Kyiv. Association for
  Computational Linguistics.

\bibitem[{Hsu et~al.(2025)Hsu, Lin, Lin, Chen, Chung, Li, Chen, Yu, Lee, Chen
  et~al.}]{hsu2025breezyvoice}
Chan-Jan Hsu, Yi-Cheng Lin, Chia-Chun Lin, Wei-Chih Chen, Ho~Lam Chung, Chen-An
  Li, Yi-Chang Chen, Chien-Yu Yu, Ming-Ji Lee, Chien-Cheng Chen, et~al. 2025.
\newblock {BreezyVoice}: Adapting {TTS} for {Taiwanese} {Mandarin} with
  enhanced polyphone disambiguation--challenges and insights.
\newblock \emph{arXiv preprint arXiv:2501.17790}.

\bibitem[{Hu et~al.(2023)Hu, Sainath, Li, Zhang, Cheng, Wang, Zhang, and
  Liu}]{10389644}
Ke~Hu, Tara~N. Sainath, Bo~Li, Yu~Zhang, Yong Cheng, Tao Wang, Yujing Zhang,
  and Frederick Liu. 2023.
\newblock \href {https://doi.org/10.1109/ASRU57964.2023.10389644} {Improving
  multilingual and code-switching {ASR} using large language model generated
  text}.
\newblock In \emph{2023 IEEE Automatic Speech Recognition and Understanding
  Workshop (ASRU)}, pages 1--7.

\bibitem[{Huang et~al.(2022)Huang, Di, Wang, and
  Ouchi}]{10.1145/3490725.3490737}
Chao Huang, Hui Di, Lina Wang, and Kazushige Ouchi. 2022.
\newblock \href {https://doi.org/10.1145/3490725.3490737} {{ECO}-{DST}: An
  efficient cross-lingual dialogue state tracking framework}.
\newblock In \emph{Proceedings of the 2021 4th International Conference on
  Machine Learning and Machine Intelligence}, MLMI '21, page 77–82, New York,
  NY, USA. Association for Computing Machinery.

\bibitem[{Huang and Wu(2007)}]{4288089}
Chien-Lin Huang and Chung-Hsien Wu. 2007.
\newblock \href {https://doi.org/10.1109/TC.2007.1064} {Generation of phonetic
  units for mixed-language speech recognition based on acoustic and contextual
  analysis}.
\newblock \emph{IEEE Transactions on Computers}, 56(9):1225--1233.

\bibitem[{Huang et~al.(2025)Huang, Lu, Shan, Qu, Zhang, Guan, Hong, and
  Li}]{10890805}
Hukai Huang, Shenghui Lu, Yahui Shan, He~Qu, Fengrun Zhang, Wenhao Guan,
  Qingyang Hong, and Lin Li. 2025.
\newblock \href {https://doi.org/10.1109/ICASSP49660.2025.10890805} {Dynamic
  language group-based {MoE}: Enhancing code-switching speech recognition with
  hierarchical routing}.
\newblock In \emph{ICASSP 2025 - 2025 IEEE International Conference on
  Acoustics, Speech and Signal Processing (ICASSP)}, pages 1--5.

\bibitem[{Huang et~al.(2024)Huang, Yang, Fu, Dunbar, and Lee}]{10446737}
Kuan-Po Huang, Chih-Kai Yang, Yu-Kuan Fu, Ewan Dunbar, and Hung-Yi Lee. 2024.
\newblock \href {https://doi.org/10.1109/ICASSP48485.2024.10446737} {Zero
  resource code-switched speech benchmark using speech utterance pairs for
  multiple spoken languages}.
\newblock In \emph{ICASSP 2024 - 2024 IEEE International Conference on
  Acoustics, Speech and Signal Processing (ICASSP)}, pages 10006--10010.

\bibitem[{Huang et~al.(2021)Huang, Wang, Wang, Miao, Xu, and
  Zhang}]{app11199106}
Zheying Huang, Pei Wang, Jian Wang, Haoran Miao, Ji~Xu, and Pengyuan Zhang.
  2021.
\newblock \href {https://doi.org/10.3390/app11199106} {Improving {Transformer}
  based end-to-end code-switching speech recognition using language
  identification}.
\newblock \emph{Applied Sciences}, 11(19).

\bibitem[{Hussein et~al.(2021)Hussein, Chowdhury, Dehak, and
  Ali}]{hussein2021balanced}
Amir Hussein, Shammur Chowdhury, Najim Dehak, and Ahmed Ali. 2021.
\newblock Balanced end-to-end monolingual pre-training for low-resourced indic
  languages code-switching speech recognition.
\newblock \emph{arXiv preprint arXiv:2106.05885}.

\bibitem[{Hussein et~al.(2023)Hussein, Chowdhury, Abdelali, Dehak, Ali, and
  Khudanpur}]{10023313}
Amir Hussein, Shammur~Absar Chowdhury, Ahmed Abdelali, Najim Dehak, Ahmed Ali,
  and Sanjeev Khudanpur. 2023.
\newblock \href {https://doi.org/10.1109/SLT54892.2023.10023313} {Textual data
  augmentation for {Arabic}-{English} code-switching speech recognition}.
\newblock In \emph{2022 IEEE Spoken Language Technology Workshop (SLT)}, pages
  777--784.

\bibitem[{Hussein et~al.(2024{\natexlab{a}})Hussein, Raj, Wiesner, Povey,
  Garcia, and Khudanpur}]{hussein24_interspeech}
Amir Hussein, Desh Raj, Matthew Wiesner, Daniel Povey, Paola Garcia, and
  Sanjeev Khudanpur. 2024{\natexlab{a}}.
\newblock \href {https://doi.org/10.21437/Interspeech.2024-1418} {{Enhancing
  Neural Transducer for Multilingual ASR with Synchronized Language
  Diarization}}.
\newblock In \emph{{Interspeech 2024}}, pages 3994--3998.

\bibitem[{Hussein et~al.(2024{\natexlab{b}})Hussein, Zeinali, Klejch, Wiesner,
  Yan, Chowdhury, Ali, Watanabe, and Khudanpur}]{10446857}
Amir Hussein, Dorsa Zeinali, Ondřej Klejch, Matthew Wiesner, Brian Yan,
  Shammur Chowdhury, Ahmed Ali, Shinji Watanabe, and Sanjeev Khudanpur.
  2024{\natexlab{b}}.
\newblock \href {https://doi.org/10.1109/ICASSP48485.2024.10446857} {Speech
  collage: Code-switched audio generation by collaging monolingual corpora}.
\newblock In \emph{ICASSP 2024 - 2024 IEEE International Conference on
  Acoustics, Speech and Signal Processing (ICASSP)}, pages 12006--12010.

\bibitem[{Iliescu et~al.(2021)Iliescu, Grand, Qirko, and van~der
  Goot}]{iliescu-etal-2021-much}
Dana-Maria Iliescu, Rasmus Grand, Sara Qirko, and Rob van~der Goot. 2021.
\newblock \href {https://doi.org/10.18653/v1/2021.calcs-1.9} {Much gracias:
  Semi-supervised code-switch detection for {S}panish-{E}nglish: How far can we
  get?}
\newblock In \emph{Proceedings of the Fifth Workshop on Computational
  Approaches to Linguistic Code-Switching}, pages 65--71, Online. Association
  for Computational Linguistics.

\bibitem[{Ismail and Mars(2026)}]{ismail2026multi}
Samia~Ben Ismail and Abdelkarim Mars. 2026.
\newblock Multi-expert topic modeling for code-switched dialectal {Arabic}: A
  dynamic routing approach to {Tunisian} social media analysis.
\newblock In \emph{ICAART (4)}, pages 3815--3822.

\bibitem[{Iyer et~al.(2023)Iyer, Barba, Birch, Pan, and
  Navigli}]{iyer-etal-2023-code}
Vivek Iyer, Edoardo Barba, Alexandra Birch, Jeff Pan, and Roberto Navigli.
  2023.
\newblock \href {https://doi.org/10.18653/v1/2023.findings-emnlp.859}
  {Code-switching with word senses for pretraining in neural machine
  translation}.
\newblock In \emph{Findings of the Association for Computational Linguistics:
  EMNLP 2023}, pages 12889--12901, Singapore. Association for Computational
  Linguistics.

\bibitem[{Jamil and Rafi(2026)}]{jamil-rafi-2026-code}
Waleed Jamil and Saima Rafi. 2026.
\newblock \href {https://doi.org/10.18653/v1/2026.abjadnlp-1.37}
  {Code-switching as a safety failure mode in large language models: An
  empirical study of {R}oman {U}rdu across {E}nglish, mixed, and
  transliteration-only inputs}.
\newblock In \emph{Proceedings of the 2nd Workshop on {NLP} for Languages Using
  {A}rabic Script}, pages 295--300, Rabat, Morocco. Association for
  Computational Linguistics.

\bibitem[{Jasim et~al.(2020)Jasim, Namboodiri, and
  Jawahar}]{jasim-etal-2020-phraseout}
Binu Jasim, Vinay Namboodiri, and C~V Jawahar. 2020.
\newblock \href {https://aclanthology.org/2020.icon-main.63/} {{P}hrase{O}ut: A
  code mixed data augmentation method for {M}ultilingual{N}eural machine
  tranlsation}.
\newblock In \emph{Proceedings of the 17th International Conference on Natural
  Language Processing (ICON)}, pages 470--474, Indian Institute of Technology
  Patna, Patna, India. NLP Association of India (NLPAI).

\bibitem[{Jawahar et~al.(2021)Jawahar, Nagoudi, Abdul-Mageed, and
  Lakshmanan}]{jawahar-etal-2021-exploring}
Ganesh Jawahar, El~Moatez~Billah Nagoudi, Muhammad Abdul-Mageed, and Laks
  Lakshmanan, V.S. 2021.
\newblock \href {https://doi.org/10.18653/v1/2021.calcs-1.6} {Exploring
  text-to-text {Transformers} for {E}nglish to {H}inglish machine translation
  with synthetic code-mixing}.
\newblock In \emph{Proceedings of the Fifth Workshop on Computational
  Approaches to Linguistic Code-Switching}, pages 36--46, Online. Association
  for Computational Linguistics.

\bibitem[{Ji et~al.(2023)Ji, Wang, Li, Ye, and Zhang}]{ji-etal-2023-isotropic}
Yixin Ji, Jikai Wang, Juntao Li, Hai Ye, and Min Zhang. 2023.
\newblock \href {https://doi.org/10.18653/v1/2023.findings-emnlp.545}
  {Isotropic representation can improve zero-shot cross-lingual transfer on
  multilingual language models}.
\newblock In \emph{Findings of the Association for Computational Linguistics:
  EMNLP 2023}, pages 8104--8118, Singapore. Association for Computational
  Linguistics.

\bibitem[{Jiang et~al.(2026)Jiang, Zhu, Wang, Chen, Zhao, Qiu, Wang, Gao, Li,
  Yu et~al.}]{jiang2026towards}
Zixuan Jiang, Yanqiao Zhu, Peng Wang, Qinyuan Chen, Xinjian Zhao, Xipeng Qiu,
  Wupeng Wang, Zhifu Gao, Xiangang Li, Kai Yu, et~al. 2026.
\newblock Towards human-like interactive speech recognition with agentic
  correction and semantic evaluation.
\newblock \emph{arXiv preprint arXiv:2605.29430}.

\bibitem[{Jose et~al.(2020)Jose, Chakravarthi, Suryawanshi, Sherly, and
  McCrae}]{9074205}
Navya Jose, Bharathi~Raja Chakravarthi, Shardul Suryawanshi, Elizabeth Sherly,
  and John~P. McCrae. 2020.
\newblock \href {https://doi.org/10.1109/ICACCS48705.2020.9074205} {A survey of
  current datasets for code-switching research}.
\newblock In \emph{2020 6th International Conference on Advanced Computing and
  Communication Systems (ICACCS)}, pages 136--141.

\bibitem[{Joshy and James(2025)}]{11306433}
Amlu~Anna Joshy and Shakespeare James. 2025.
\newblock \href {https://doi.org/10.1109/CONECCT65861.2025.11306433} {Sentiment
  analysis of {Malayalam}-{English} code-mixed data using {GPT}-based
  augmentation}.
\newblock In \emph{2025 IEEE International Conference on Electronics, Computing
  and Communication Technologies (CONECCT)}, pages 1--6.

\bibitem[{Kadaoui et~al.(2024)Kadaoui, Ali, Toyin, Mohammed, and
  Aldarmaki}]{kadaoui-etal-2024-polywer}
Karima Kadaoui, Maryam~Al Ali, Hawau~Olamide Toyin, Ibrahim Mohammed, and Hanan
  Aldarmaki. 2024.
\newblock \href {https://doi.org/10.18653/v1/2024.findings-emnlp.356}
  {{P}oly{WER}: A holistic evaluation framework for code-switched speech
  recognition}.
\newblock In \emph{Findings of the Association for Computational Linguistics:
  EMNLP 2024}, pages 6144--6153, Miami, Florida, USA. Association for
  Computational Linguistics.

\bibitem[{Kang(2024)}]{kang-2024-covoswitch}
Yeeun Kang. 2024.
\newblock \href {https://doi.org/10.18653/v1/2024.acl-srw.40}
  {{C}o{V}o{S}witch: Machine translation of synthetic code-switched text based
  on intonation units}.
\newblock In \emph{Proceedings of the 62nd Annual Meeting of the Association
  for Computational Linguistics (Volume 4: Student Research Workshop)}, pages
  345--357, Bangkok, Thailand. Association for Computational Linguistics.

\bibitem[{Kargaran et~al.(2024)Kargaran, Yvon, and
  Sch{\"u}tze}]{kargaran2024masklid}
Amir~Hossein Kargaran, Fran{\c{c}}ois Yvon, and Hinrich Sch{\"u}tze. 2024.
\newblock {MaskLID}: Code-switching language identification through iterative
  masking.
\newblock In \emph{Proceedings of the 62nd Annual Meeting of the Association
  for Computational Linguistics (Volume 2: Short Papers)}, pages 459--469.

\bibitem[{Kartik et~al.(2024)Kartik, Soni, Kunchukuttan, Chakraborty, and
  Akhtar}]{kartik-etal-2024-synthetic}
Kartik, Sanjana Soni, Anoop Kunchukuttan, Tanmoy Chakraborty, and Md~Shad
  Akhtar. 2024.
\newblock \href {https://aclanthology.org/2024.lrec-main.1345/} {Synthetic data
  generation and joint learning for robust code-mixed translation}.
\newblock In \emph{Proceedings of the 2024 Joint International Conference on
  Computational Linguistics, Language Resources and Evaluation (LREC-COLING
  2024)}, pages 15480--15492, Torino, Italia. ELRA and ICCL.

\bibitem[{Khan et~al.(2022)Khan, Kanade, Budhrani, Jhanglani, and
  Xu}]{khan-etal-2022-sit}
Abdul Khan, Hrishikesh Kanade, Girish Budhrani, Preet Jhanglani, and Jia Xu.
  2022.
\newblock \href {https://doi.org/10.18653/v1/2022.wmt-1.114} {{SIT} at
  {M}ix{MT} 2022: Fluent translation built on giant pre-trained models}.
\newblock In \emph{Proceedings of the Seventh Conference on Machine Translation
  (WMT)}, pages 1136--1144, Abu Dhabi, United Arab Emirates (Hybrid).
  Association for Computational Linguistics.

\bibitem[{Khanuja et~al.(2020)Khanuja, Dandapat, Srinivasan, Sitaram, and
  Choudhury}]{khanuja-etal-2020-gluecos}
Simran Khanuja, Sandipan Dandapat, Anirudh Srinivasan, Sunayana Sitaram, and
  Monojit Choudhury. 2020.
\newblock \href {https://doi.org/10.18653/v1/2020.acl-main.329} {{GLUEC}o{S}:
  An evaluation benchmark for code-switched {NLP}}.
\newblock In \emph{Proceedings of the 58th Annual Meeting of the Association
  for Computational Linguistics}, pages 3575--3585, Online. Association for
  Computational Linguistics.

\bibitem[{Khassanov et~al.(2019)Khassanov, Xu, Pham, Zeng, Chng, Ni, and
  Ma}]{khassanov19_interspeech}
Yerbolat Khassanov, Haihua Xu, Van~Tung Pham, Zhiping Zeng, Eng~Siong Chng,
  Chongjia Ni, and Bin Ma. 2019.
\newblock \href {https://doi.org/10.21437/Interspeech.2019-1867} {{Constrained
  Output Embeddings for End-to-End Code-Switching Speech Recognition with Only
  Monolingual Data}}.
\newblock In \emph{{Interspeech 2019}}, pages 2160--2164.

\bibitem[{Khatri et~al.(2023)Khatri, Srivastava, and
  Vig}]{khatri-etal-2023-translate}
Jyotsana Khatri, Vivek Srivastava, and Lovekesh Vig. 2023.
\newblock \href {https://doi.org/10.18653/v1/2023.ijcnlp-short.10} {Can you
  translate for me? code-switched machine translation with large language
  models}.
\newblock In \emph{Proceedings of the 13th International Joint Conference on
  Natural Language Processing and the 3rd Conference of the Asia-Pacific
  Chapter of the Association for Computational Linguistics (Volume 2: Short
  Papers)}, pages 83--92, Nusa Dua, Bali. Association for Computational
  Linguistics.

\bibitem[{Kim et~al.(2024)Kim, Lee, Kim, and Ko}]{10627333}
Gwantae Kim, Bokyeung Lee, Donghyeon Kim, and Hanseok Ko. 2024.
\newblock \href {https://doi.org/10.1109/ICASSPW62465.2024.10627333} {Gated
  low-rank adaptation for personalized code-switching automatic speech
  recognition on the low-spec devices}.
\newblock In \emph{2024 IEEE International Conference on Acoustics, Speech, and
  Signal Processing Workshops (ICASSPW)}, pages 760--764.

\bibitem[{Kim(2026)}]{article}
Heesoo Kim. 2026.
\newblock \href {https://doi.org/10.29403/LI.30.1.2} {Emotion classification in
  {LLM}-based code-switched text: Linguistic constraints and directional
  effects in {English}-{Spanish} code switching}.
\newblock \emph{Language and Information}, 30:25--46.

\bibitem[{Kodali et~al.(2025)Kodali, Goel, Asapu, Bonagiri, Govil, Choudhury,
  Kumaraguru, and Shrivastava}]{10.1145/3748312}
Prashant Kodali, Anmol Goel, Likhith Asapu, Vamshi~Krishna Bonagiri, Anirudh
  Govil, Monojit Choudhury, Ponnurangam Kumaraguru, and Manish Shrivastava.
  2025.
\newblock \href {https://doi.org/10.1145/3748312} {From human judgements to
  predictive models: Unravelling acceptability in code-mixed sentences}.
\newblock \emph{ACM Trans. Asian Low-Resour. Lang. Inf. Process.}, 24(9).

\bibitem[{Kodali et~al.(2026)Kodali, Shivkumar, Joshi, Choudhury, Kumaraguru,
  and Shrivastava}]{10.1145/3799830.3799852}
Prashant Kodali, Vaishnavi Shivkumar, Swarang Joshi, Monojit Choudhury,
  Ponnurangam Kumaraguru, and Manish Shrivastava. 2026.
\newblock \href {https://doi.org/10.1145/3799830.3799852} {Adapting
  multilingual models to code-mixed tasks via model merging}.
\newblock In \emph{Proceedings of the 13th ACM IKDD International Conference on
  Data Science}, CODS '25, page 199–207, New York, NY, USA. Association for
  Computing Machinery.

\bibitem[{Krasitskii et~al.(2025)Krasitskii, Kolesnikova, Chanona~Hernandez,
  Sidorov, and Gelbukh}]{krasitskii-etal-2025-advancing}
Mikhail Krasitskii, Olga Kolesnikova, Liliana Chanona~Hernandez, Grigori
  Sidorov, and Alexander Gelbukh. 2025.
\newblock \href {https://doi.org/10.18653/v1/2025.nlp4dh-1.27} {Advancing
  sentiment analysis in {T}amil-{E}nglish code-mixed texts: Challenges and
  {Transformer}-based solutions}.
\newblock In \emph{Proceedings of the 5th International Conference on Natural
  Language Processing for Digital Humanities}, pages 305--312, Albuquerque,
  USA. Association for Computational Linguistics.

\bibitem[{Krishnan et~al.(2021)Krishnan, Anastasopoulos, Purohit, and
  Rangwala}]{krishnan-etal-2021-multilingual}
Jitin Krishnan, Antonios Anastasopoulos, Hemant Purohit, and Huzefa Rangwala.
  2021.
\newblock \href {https://doi.org/10.18653/v1/2021.mrl-1.18} {Multilingual
  code-switching for zero-shot cross-lingual intent prediction and slot
  filling}.
\newblock In \emph{Proceedings of the 1st Workshop on Multilingual
  Representation Learning}, pages 211--223, Punta Cana, Dominican Republic.
  Association for Computational Linguistics.

\bibitem[{Kronis et~al.(2024)Kronis, Salimbajevs, and
  Pinnis}]{kronis-etal-2024-code}
Martins Kronis, Askars Salimbajevs, and M{\={a}}rcis Pinnis. 2024.
\newblock \href {https://aclanthology.org/2024.lrec-main.308/} {Code-mixed text
  augmentation for {L}atvian {ASR}}.
\newblock In \emph{Proceedings of the 2024 Joint International Conference on
  Computational Linguistics, Language Resources and Evaluation (LREC-COLING
  2024)}, pages 3469--3479, Torino, Italia. ELRA and ICCL.

\bibitem[{Kuckreja et~al.(2026)Kuckreja, Gupta, Hamed, Solorio, Khan, and
  Dhall}]{ICLR2026_3122aaa2}
Kartik Kuckreja, Parul Gupta, Injy Hamed, Thamar Solorio, Muhammad~Haris Khan,
  and Abhinav Dhall. 2026.
\newblock \href
  {https://proceedings.iclr.cc/paper_files/paper/2026/file/3122aaa22b2fe83f9cead1a696f65ceb-Paper-Conference.pdf}
  {Tell me habibi, is it real or fake?}
\newblock In \emph{International Conference on Learning Representations},
  volume 2026, pages 28903--28926.

\bibitem[{Kulkarni et~al.(2023)Kulkarni, Kulkarni, Couceiro, and
  Aldarmaki}]{kulkarni2023adapting}
Atharva Kulkarni, Ajinkya Kulkarni, Miguel Couceiro, and Hanan Aldarmaki. 2023.
\newblock Adapting the adapters for code-switching in multilingual {ASR}.
\newblock \emph{arXiv preprint arXiv:2310.07423}.

\bibitem[{Kumar(2024)}]{kumar2024two}
Rohit Kumar. 2024.
\newblock A two-stage transliteration approach to improve performance of a
  multilingual {ASR}.
\newblock \emph{arXiv preprint arXiv:2410.14709}.

\bibitem[{Kumar et~al.(2024)Kumar, Akhtar, Cambria, and
  Chakraborty}]{kumar-etal-2024-semeval}
Shivani Kumar, Md.~Shad Akhtar, Erik Cambria, and Tanmoy Chakraborty. 2024.
\newblock \href {https://doi.org/10.18653/v1/2024.semeval-1.270} {{S}em{E}val
  2024 - task 10: Emotion discovery and reasoning its flip in conversation
  ({ED}i{R}e{F})}.
\newblock In \emph{Proceedings of the 18th International Workshop on Semantic
  Evaluation (SemEval-2024)}, pages 1933--1946, Mexico City, Mexico.
  Association for Computational Linguistics.

\bibitem[{Kumar et~al.(2022)Kumar, Murthy, and
  Dhamecha}]{kumar-etal-2022-utilizing}
Vishwajeet Kumar, Rudra Murthy, and Tejas Dhamecha. 2022.
\newblock \href {https://doi.org/10.18653/v1/2022.findings-emnlp.283} {On
  utilizing constituent language resources to improve downstream tasks in
  {H}inglish}.
\newblock In \emph{Findings of the Association for Computational Linguistics:
  EMNLP 2022}, pages 3859--3865, Abu Dhabi, United Arab Emirates. Association
  for Computational Linguistics.

\bibitem[{Kumaresan et~al.(2022)Kumaresan, Premjith, Sakuntharaj, Thavareesan,
  Navaneethakrishnan, Madasamy, Chakravarthi, and
  McCrae}]{10.1145/3503162.3503179}
Prasanna~Kumar Kumaresan, Premjith, Ratnasingam Sakuntharaj, Sajeetha
  Thavareesan, Subalalitha Navaneethakrishnan, Anand~Kumar Madasamy,
  Bharathi~Raja Chakravarthi, and John~P. McCrae. 2022.
\newblock \href {https://doi.org/10.1145/3503162.3503179} {Findings of shared
  task on offensive language identification in {Tamil} and {Malayalam}}.
\newblock In \emph{Proceedings of the 13th Annual Meeting of the Forum for
  Information Retrieval Evaluation}, FIRE '21, page 16–18, New York, NY, USA.
  Association for Computing Machinery.

\bibitem[{Kuwanto et~al.(2026)Kuwanto, Agarwal, Winata, and
  Wijaya}]{kuwanto-etal-2026-linguistics}
Garry Kuwanto, Chaitanya Agarwal, Genta~Indra Winata, and Derry~Tanti Wijaya.
  2026.
\newblock \href {https://doi.org/10.18653/v1/2026.cdl-1.1} {Linguistics theory
  meets {LLM}: Code-switched text generation via equivalence constrained large
  language models}.
\newblock In \emph{Proceedings of the 1st Workshop on Computational
  Developmental Linguistics ({CDL})}, pages 1--14, Grand Hyatt Manchester San
  Diego, 1 Market Pl, San Diego, CA 92101. Association for Computational
  Linguistics.

\bibitem[{Kuwanto et~al.(2023)Kuwanto, Aky{\"u}rek, Tourni, Li, Jones, and
  Wijaya}]{kuwanto2023low}
Garry Kuwanto, Afra~Feyza Aky{\"u}rek, Isidora~Chara Tourni, Siyang Li, Alex
  Jones, and Derry Wijaya. 2023.
\newblock \href {https://doi.org/10.1007/978-981-99-7025-4_39} {Low-resource
  machine translation training curriculum fit for low-resource languages}.
\newblock In \emph{PRICAI 2023: Trends in Artificial Intelligence}, Lecture
  Notes in Computer Science, pages 453--458. Springer.

\bibitem[{Lai et~al.(2021)Lai, Huang, Jing, Chen, Xu, and
  Liu}]{lai-etal-2021-saliency-based}
Siyu Lai, Hui Huang, Dong Jing, Yufeng Chen, Jinan Xu, and Jian Liu. 2021.
\newblock \href {https://doi.org/10.18653/v1/2021.findings-emnlp.55}
  {Saliency-based multi-view mixed language training for zero-shot
  cross-lingual classification}.
\newblock In \emph{Findings of the Association for Computational Linguistics:
  EMNLP 2021}, pages 599--610, Punta Cana, Dominican Republic. Association for
  Computational Linguistics.

\bibitem[{Laureano De~Leon et~al.(2025)Laureano De~Leon, Abbas,
  Tayyar~Madabushi, and Lee}]{laureano-de-leon-etal-2025-evaluating}
Frances~Adriana Laureano De~Leon, Asim Abbas, Harish Tayyar~Madabushi, and Mark
  Lee. 2025.
\newblock \href {https://aclanthology.org/2025.ranlp-1.75/} {Evaluating large
  language models on multiword expressions in multilingual and code-switched
  contexts}.
\newblock In \emph{Proceedings of the 15th International Conference on Recent
  Advances in Natural Language Processing - Natural Language Processing in the
  Generative AI Era}, pages 644--653, Varna, Bulgaria. INCOMA Ltd., Shoumen,
  Bulgaria.

\bibitem[{Lee et~al.(2021)Lee, Lee, Lee, Chun, and
  Hwang}]{10.1145/3459637.3482176}
Dohyeon Lee, Jaeseong Lee, Gyewon Lee, Byung-gon Chun, and Seung-won Hwang.
  2021.
\newblock \href {https://doi.org/10.1145/3459637.3482176} {{SCOPA}: Soft
  code-switching and pairwise alignment for zero-shot cross-lingual transfer}.
\newblock In \emph{Proceedings of the 30th ACM International Conference on
  Information \& Knowledge Management}, CIKM '21, page 3176–3180, New York,
  NY, USA. Association for Computing Machinery.

\bibitem[{Lee et~al.(2019)Lee, Yue, and Li}]{lee19d_interspeech}
Grandee Lee, Xianghu Yue, and Haizhou Li. 2019.
\newblock \href {https://doi.org/10.21437/Interspeech.2019-1382}
  {{Linguistically Motivated Parallel Data Augmentation for Code-Switch
  Language Modeling}}.
\newblock In \emph{{Interspeech 2019}}, pages 3730--3734.

\bibitem[{Lee et~al.(2024)Lee, Jung, and Hwang}]{lee-etal-2024-commit}
Jaeseong Lee, YeonJoon Jung, and Seung-won Hwang. 2024.
\newblock \href {https://doi.org/10.18653/v1/2024.findings-naacl.198}
  {{COMMIT}: Code-mixing {E}nglish-centric large language model for
  multilingual instruction tuning}.
\newblock In \emph{Findings of the Association for Computational Linguistics:
  NAACL 2024}, pages 3130--3137, Mexico City, Mexico. Association for
  Computational Linguistics.

\bibitem[{Lee et~al.(2025{\natexlab{a}})Lee, Woo, Ko, and
  Son}]{lee-etal-2025-controlling}
Nahyun Lee, Yeongseo Woo, Hyunwoo Ko, and Guijin Son. 2025{\natexlab{a}}.
\newblock \href {https://doi.org/10.18653/v1/2025.acl-srw.81} {Controlling
  language confusion in multilingual {LLM}s}.
\newblock In \emph{Proceedings of the 63rd Annual Meeting of the Association
  for Computational Linguistics (Volume 4: Student Research Workshop)}, pages
  1026--1035, Vienna, Austria. Association for Computational Linguistics.

\bibitem[{Lee et~al.(2025{\natexlab{b}})Lee, Choi, Kim, and Kang}]{lee2025sage}
Sangmin Lee, Woongjib Choi, Jihyun Kim, and Hong-Goo Kang. 2025{\natexlab{b}}.
\newblock {SAGE}-{LD}: Towards scalable and generalizable end-to-end language
  diarization via simulated data augmentation.
\newblock \emph{arXiv preprint arXiv:2510.00582}.

\bibitem[{Lee et~al.(2025{\natexlab{c}})Lee, Chung, Um, and
  Kang}]{lee-etal-2025-unicom}
Sangmin Lee, Woojin Chung, Seyun Um, and Hong-Goo Kang. 2025{\natexlab{c}}.
\newblock \href {https://doi.org/10.18653/v1/2025.findings-emnlp.715}
  {{U}ni{C}o{M}: A universal code-switching speech generator}.
\newblock In \emph{Findings of the Association for Computational Linguistics:
  EMNLP 2025}, pages 13273--13288, Suzhou, China. Association for Computational
  Linguistics.

\bibitem[{Lee and Wang(2015)}]{lee-wang-2015-emotion}
Sophia Lee and Zhongqing Wang. 2015.
\newblock \href {https://doi.org/10.18653/v1/W15-3116} {Emotion in
  code-switching texts: Corpus construction and analysis}.
\newblock In \emph{Proceedings of the Eighth {SIGHAN} Workshop on {C}hinese
  Language Processing}, pages 91--99, Beijing, China. Association for
  Computational Linguistics.

\bibitem[{Lehečka et~al.(2024)Lehečka, Psutka, Smidl, Ircing, and
  Psutka}]{lehecka24_interspeech}
Jan Lehečka, Josef~V. Psutka, Lubos Smidl, Pavel Ircing, and Josef Psutka.
  2024.
\newblock \href {https://doi.org/10.21437/Interspeech.2024-472} {{A Comparative
  Analysis of Bilingual and Trilingual Wav2Vec Models for Automatic Speech
  Recognition in Multilingual Oral History Archives}}.
\newblock In \emph{{Interspeech 2024}}, pages 1285--1289.

\bibitem[{Leventhal et~al.(2026)Leventhal, Diarra, Coulibaly, and
  Kamat{\'e}}]{leventhal-etal-2026-kunnafonidilaw}
Michael Leventhal, Yacouba Diarra, Nouhoum Coulibaly, and Panga~Azazia
  Kamat{\'e}. 2026.
\newblock \href {https://doi.org/10.18653/v1/2026.africanlp-main.18}
  {Kunnafonidilaw ka cadeau: an {ASR} dataset of present-day {B}ambara}.
\newblock In \emph{Proceedings of the 7th Workshop on {A}frican Natural
  Language Processing ({A}frica{NLP} 2026)}, pages 190--196, Rabat, Morocco.
  Association for Computational Linguistics.

\bibitem[{Li et~al.(2019{\natexlab{a}})Li, Zhang, Sainath, Wu, and
  Chan}]{8682674}
Bo~Li, Yu~Zhang, Tara Sainath, Yonghui Wu, and William Chan.
  2019{\natexlab{a}}.
\newblock \href {https://doi.org/10.1109/ICASSP.2019.8682674} {Bytes are all
  you need: End-to-end multilingual speech recognition and synthesis with
  bytes}.
\newblock In \emph{ICASSP 2019 - 2019 IEEE International Conference on
  Acoustics, Speech and Signal Processing (ICASSP)}, pages 5621--5625.

\bibitem[{Li et~al.(2024{\natexlab{a}})Li, Huang, Ching, Dai, and
  Chen}]{li-etal-2024-prealign}
Jiahuan Li, Shujian Huang, Aarron Ching, Xinyu Dai, and Jiajun Chen.
  2024{\natexlab{a}}.
\newblock \href {https://doi.org/10.18653/v1/2024.emnlp-main.572}
  {{P}re{A}lign: Boosting cross-lingual transfer by early establishment of
  multilingual alignment}.
\newblock In \emph{Proceedings of the 2024 Conference on Empirical Methods in
  Natural Language Processing}, pages 10246--10257, Miami, Florida, USA.
  Association for Computational Linguistics.

\bibitem[{Li et~al.(2019{\natexlab{b}})Li, Li, Ye, Zhao, and Gong}]{8683223}
Ke~Li, Jinyu Li, Guoli Ye, Rui Zhao, and Yifan Gong. 2019{\natexlab{b}}.
\newblock \href {https://doi.org/10.1109/ICASSP.2019.8683223} {Towards
  code-switching {ASR} for end-to-end {CTC} models}.
\newblock In \emph{ICASSP 2019 - 2019 IEEE International Conference on
  Acoustics, Speech and Signal Processing (ICASSP)}, pages 6076--6080.

\bibitem[{Li et~al.(2022)Li, Luo, Hu, Liu, Zhang, Li, and Hong}]{10037997}
Song Li, Haoneng Luo, Wenxuan Hu, Yuan Liu, Shiliang Zhang, Lin Li, and
  Qingyang Hong. 2022.
\newblock \href {https://doi.org/10.1109/ISCSLP57327.2022.10037997} {Towards
  language-universal {Mandarin}-{English} speech recognition with unsupervised
  label synchronous adaptation}.
\newblock In \emph{2022 13th International Symposium on Chinese Spoken Language
  Processing (ISCSLP)}, pages 16--20.

\bibitem[{Li et~al.(2025)Li, Xin, Miao, Long, and Ungar}]{li-etal-2025-impact}
Yihao Li, Jiayi Xin, Miranda~Muqing Miao, Qi~Long, and Lyle Ungar. 2025.
\newblock \href {https://doi.org/10.18653/v1/2025.emnlp-main.1654} {The impact
  of language mixing on bilingual {LLM} reasoning}.
\newblock In \emph{Proceedings of the 2025 Conference on Empirical Methods in
  Natural Language Processing}, pages 32531--32548, Suzhou, China. Association
  for Computational Linguistics.

\bibitem[{Li and Fung(2013)}]{6639094}
Ying Li and Pascale Fung. 2013.
\newblock \href {https://doi.org/10.1109/ICASSP.2013.6639094} {Improved mixed
  language speech recognition using asymmetric acoustic model and language
  model with code-switch inversion constraints}.
\newblock In \emph{2013 IEEE International Conference on Acoustics, Speech and
  Signal Processing}, pages 7368--7372.

\bibitem[{Li and Fung(2014)}]{li-fung-2014-language}
Ying Li and Pascale Fung. 2014.
\newblock \href {https://doi.org/10.3115/v1/D14-1098} {Language modeling with
  functional head constraint for code switching speech recognition}.
\newblock In \emph{Proceedings of the 2014 Conference on Empirical Methods in
  Natural Language Processing ({EMNLP})}, pages 907--916, Doha, Qatar.
  Association for Computational Linguistics.

\bibitem[{Li et~al.(2023)Li, Liu, Wu, and Meng}]{li23z_interspeech}
Yunxiang Li, Pengfei Liu, Xixin Wu, and Helen Meng. 2023.
\newblock \href {https://doi.org/10.21437/Interspeech.2023-1675}
  {{PunCantonese: A Benchmark Corpus for Low-Resource Cantonese Punctuation
  Restoration from Speech Transcripts}}.
\newblock In \emph{{Interspeech 2023}}, pages 2183--2187.

\bibitem[{Li et~al.(2024{\natexlab{b}})Li, Hu, Chen, Chen, Guo, and
  Zhang}]{li2024improving}
Zhuoran Li, Chunming Hu, Junfan Chen, Zhijun Chen, Xiaohui Guo, and Richong
  Zhang. 2024{\natexlab{b}}.
\newblock Improving zero-shot cross-lingual transfer via progressive
  code-switching.
\newblock \emph{arXiv preprint arXiv:2406.13361}.

\bibitem[{Liang et~al.(2022{\natexlab{a}})Liang, Shou, Pei, Gong, Zuo, Zuo, and
  Jiang}]{liang-etal-2022-label}
Shining Liang, Linjun Shou, Jian Pei, Ming Gong, Wanli Zuo, Xianglin Zuo, and
  Daxin Jiang. 2022{\natexlab{a}}.
\newblock \href {https://doi.org/10.18653/v1/2022.emnlp-main.673} {Label-aware
  multi-level contrastive learning for cross-lingual spoken language
  understanding}.
\newblock In \emph{Proceedings of the 2022 Conference on Empirical Methods in
  Natural Language Processing}, pages 9903--9918, Abu Dhabi, United Arab
  Emirates. Association for Computational Linguistics.

\bibitem[{Liang et~al.(2022{\natexlab{b}})Liang, Chen, Yu, Zhu, Xu, Gao, and
  Xie}]{liang2022npu}
Yuhao Liang, Peikun Chen, Fan Yu, Xinfa Zhu, Tianyi Xu, Yingying Gao, and Lei
  Xie. 2022{\natexlab{b}}.
\newblock The {NPU}-{ASLP} system for the {ISCSLP} 2022 magichub code-swiching
  {ASR} challenge.
\newblock In \emph{2022 13th International Symposium on Chinese Spoken Language
  Processing (ISCSLP)}, pages 532--536. IEEE.

\bibitem[{Liang et~al.(2023)Liang, Song, Ma, Du, Yu, and
  Chen}]{liang2023improving}
Zheng Liang, Zheshu Song, Ziyang Ma, Chenpeng Du, Kai Yu, and Xie Chen. 2023.
\newblock Improving code-switching and named entity recognition in asr with
  speech editing based data augmentation.
\newblock \emph{arXiv preprint arXiv:2306.08588}.

\bibitem[{Litschko et~al.(2023)Litschko, Artemova, and
  Plank}]{litschko-etal-2023-boosting}
Robert Litschko, Ekaterina Artemova, and Barbara Plank. 2023.
\newblock \href {https://doi.org/10.18653/v1/2023.findings-acl.193} {Boosting
  zero-shot cross-lingual retrieval by training on artificially code-switched
  data}.
\newblock In \emph{Findings of the Association for Computational Linguistics:
  ACL 2023}, pages 3096--3108, Toronto, Canada. Association for Computational
  Linguistics.

\bibitem[{Liu and Cao(2021)}]{9414158}
Guoyu Liu and Lixin Cao. 2021.
\newblock \href {https://doi.org/10.1109/ICASSP39728.2021.9414158} {Code-switch
  speech rescoring with monolingual data}.
\newblock In \emph{ICASSP 2021 - 2021 IEEE International Conference on
  Acoustics, Speech and Signal Processing (ICASSP)}, pages 6229--6233.

\bibitem[{Liu et~al.(2021{\natexlab{a}})Liu, Perera, Zhang, Dauwels, Khong,
  Khudanpur, and Styles}]{liu21d_interspeech}
Hexin Liu, Leibny Paola~García Perera, Xinyi Zhang, Justin Dauwels, Andy~W.H.
  Khong, Sanjeev Khudanpur, and Suzy~J. Styles. 2021{\natexlab{a}}.
\newblock \href {https://doi.org/10.21437/Interspeech.2021-82} {{End-to-End
  Language Diarization for Bilingual Code-Switching Speech}}.
\newblock In \emph{{Interspeech 2021}}, pages 1489--1493.

\bibitem[{Liu et~al.(2026)Liu, Zhang, Zhang, Zhang, Shi, Chng, and
  Li}]{11447403}
Hexin Liu, Haoyang Zhang, Qiquan Zhang, Xiangyu Zhang, Dongyuan Shi, Eng~Siong
  Chng, and Haizhou Li. 2026.
\newblock \href {https://doi.org/10.1109/TASLPRO.2026.3675776} {Code-switching
  speech recognition under the lens: Model- and data-centric perspectives}.
\newblock \emph{IEEE Transactions on Audio, Speech and Language Processing},
  34:1853--1865.

\bibitem[{Liu et~al.(2024)Liu, Shi, Wang, Zhang, and
  Schütze}]{doi:10.3233/FAIA240950}
Yongkang Liu, Feng Shi, Daling Wang, Yifei Zhang, and Hinrich Schütze. 2024.
\newblock \href {https://doi.org/10.3233/FAIA240950} {{ChatZero}: Zero-shot
  cross-lingual dialogue generation via pseudo-target language}.
\newblock In \emph{ECAI 2024}, pages 3867--3874. SAGE Publications.

\bibitem[{Liu et~al.(2021{\natexlab{b}})Liu, Winata, and
  Fung}]{liu-etal-2021-continual}
Zihan Liu, Genta~Indra Winata, and Pascale Fung. 2021{\natexlab{b}}.
\newblock \href {https://doi.org/10.18653/v1/2021.findings-acl.239} {Continual
  mixed-language pre-training for extremely low-resource neural machine
  translation}.
\newblock In \emph{Findings of the Association for Computational Linguistics:
  ACL-IJCNLP 2021}, pages 2706--2718, Online. Association for Computational
  Linguistics.

\bibitem[{Liu et~al.(2019)Liu, Winata, Lin, Xu, and
  Fung}]{liu2019attentioninformedmixedlanguagetrainingzeroshot}
Zihan Liu, Genta~Indra Winata, Zhaojiang Lin, Peng Xu, and Pascale Fung. 2019.
\newblock \href {https://arxiv.org/abs/1911.09273} {Attention-informed
  mixed-language training for zero-shot cross-lingual task-oriented dialogue
  systems}.
\newblock \emph{Preprint}, arXiv:1911.09273.

\bibitem[{Lodagala et~al.(2025)Lodagala, Alkanhal, Izham, Mehta, Chowdhury,
  Makki, Hussein, Henter, and Ali}]{lodagala25_interspeech}
Vasista~Sai Lodagala, Lamya Alkanhal, Daniel Izham, Shivam Mehta, Shammur~Absar
  Chowdhury, Aqeelah Makki, Hamdy~S. Hussein, Gustav~Eje Henter, and Ahmed Ali.
  2025.
\newblock \href {https://doi.org/10.21437/Interspeech.2025-2573} {{SawtArabi: A
  Benchmark Corpus for Arabic TTS. Standard, Dialectal and Code-Switching}}.
\newblock In \emph{{Interspeech 2025}}, pages 4793--4797.

\bibitem[{Lu et~al.(2023)Lu, Lu, and Zhang}]{lu-etal-2023-take}
Jinliang Lu, Yu~Lu, and Jiajun Zhang. 2023.
\newblock \href {https://doi.org/10.18653/v1/2023.findings-emnlp.190} {Take a
  closer look at multilinguality! improve multilingual pre-training using
  monolingual corpora only}.
\newblock In \emph{Findings of the Association for Computational Linguistics:
  EMNLP 2023}, pages 2891--2907, Singapore. Association for Computational
  Linguistics.

\bibitem[{Lu et~al.(2020)Lu, Huang, Li, Guo, and Qian}]{lu20f_interspeech}
Yizhou Lu, Mingkun Huang, Hao Li, Jiaqi Guo, and Yanmin Qian. 2020.
\newblock \href {https://doi.org/10.21437/Interspeech.2020-2485} {{Bi-Encoder
  Transformer Network for Mandarin-English Code-Switching Speech Recognition
  Using Mixture of Experts}}.
\newblock In \emph{{Interspeech 2020}}, pages 4766--4770.

\bibitem[{Luo et~al.(2018)Luo, Jiang, Zhao, Gong, Zou, and Li}]{luo2018towards}
Ne~Luo, Dongwei Jiang, Shuaijiang Zhao, Caixia Gong, Wei Zou, and Xiangang Li.
  2018.
\newblock Towards end-to-end code-switching speech recognition.
\newblock \emph{arXiv preprint arXiv:1810.13091}.

\bibitem[{Lyu et~al.(2013{\natexlab{a}})Lyu, Chng, and Li}]{6639083}
Dau-Cheng Lyu, Eng-Siong Chng, and Haizhou Li. 2013{\natexlab{a}}.
\newblock \href {https://doi.org/10.1109/ICASSP.2013.6639083} {Language
  diarization for code-switch conversational speech}.
\newblock In \emph{2013 IEEE International Conference on Acoustics, Speech and
  Signal Processing}, pages 7314--7318.

\bibitem[{Lyu et~al.(2013{\natexlab{b}})Lyu, Chng, and Li}]{6625316}
Dau-Cheng Lyu, Eng-Siong Chng, and Haizhou Li. 2013{\natexlab{b}}.
\newblock \href {https://doi.org/10.1109/ChinaSIP.2013.6625316} {Language
  diarization for conversational code-switch speech with pronunciation
  dictionary adaptation}.
\newblock In \emph{2013 IEEE China Summit and International Conference on
  Signal and Information Processing}, pages 147--150.

\bibitem[{Ma et~al.(2023)Ma, Wang, Li, Yang, Du, and Fu}]{10389662}
Guodong Ma, Wenxuan Wang, Yuke Li, Yuting Yang, Binbin Du, and Haoran Fu. 2023.
\newblock \href {https://doi.org/10.1109/ASRU57964.2023.10389662}
  {{LAE}-{ST}-{MOE}: Boosted language-aware encoder using speech translation
  auxiliary task for {E2E} code-switching {ASR}}.
\newblock In \emph{2023 IEEE Automatic Speech Recognition and Understanding
  Workshop (ASRU)}, pages 1--8.

\bibitem[{Madhani et~al.(2023)Madhani, Khapra, and
  Kunchukuttan}]{madhani-etal-2023-bhasa}
Yash Madhani, Mitesh~M. Khapra, and Anoop Kunchukuttan. 2023.
\newblock \href {https://doi.org/10.18653/v1/2023.acl-short.71}
  {Bhasa-{A}bhijnaanam: Native-script and romanized language identification for
  22 {I}ndic languages}.
\newblock In \emph{Proceedings of the 61st Annual Meeting of the Association
  for Computational Linguistics (Volume 2: Short Papers)}, pages 816--826,
  Toronto, Canada. Association for Computational Linguistics.

\bibitem[{Madhumani et~al.(2020)Madhumani, Shah, Abraham, Joshi, and
  Sitaram}]{madhumani2020learning}
Gurunath~Reddy Madhumani, Sanket Shah, Basil Abraham, Vikas Joshi, and Sunayana
  Sitaram. 2020.
\newblock Learning not to discriminate: Task agnostic learning for improving
  monolingual and code-switched speech recognition.
\newblock \emph{arXiv preprint arXiv:2006.05257}.

\bibitem[{Mahata et~al.(2019)Mahata, Mandal, Das, and
  Bandyopadhyay}]{10.1145/3368567.3368579}
Sainik~Kumar Mahata, Soumil Mandal, Dipankar Das, and Sivaji Bandyopadhyay.
  2019.
\newblock \href {https://doi.org/10.1145/3368567.3368579} {Code-mixed to
  monolingual translation framework}.
\newblock In \emph{Proceedings of the 11th Annual Meeting of the Forum for
  Information Retrieval Evaluation}, FIRE '19, page 30–35, New York, NY, USA.
  Association for Computing Machinery.

\bibitem[{Maimaiti et~al.(2025)Maimaiti, Zheng, Zhang, Zhang, Luo, and
  Huang}]{MAIMAITI2025113919}
Mieradilijiang Maimaiti, Yuanhang Zheng, Ji~Zhang, Yue Zhang, Wenpei Luo, and
  Kaiyu Huang. 2025.
\newblock \href {https://doi.org/10.1016/j.knosys.2025.113919} {Improving
  cross-lingual representation for semantic retrieval with code-switching}.
\newblock \emph{Knowledge-Based Systems}, 325:113919.

\bibitem[{Manghat et~al.(2020)Manghat, Manghat, and
  Schultz}]{manghat20_interspeech}
Sreeja Manghat, Sreeram Manghat, and Tanja Schultz. 2020.
\newblock \href {https://doi.org/10.21437/Interspeech.2020-1936}
  {{Malayalam-English Code-Switched: Grapheme to Phoneme System}}.
\newblock In \emph{{Interspeech 2020}}, pages 4133--4137.

\bibitem[{Mao and Zhang(2023)}]{10393998}
Tianjun Mao and Chenghong Zhang. 2023.
\newblock \href {https://doi.org/10.1109/SMC53992.2023.10393998} {{CMM}:
  Code-switching with manifold mixup for cross-lingual spoken language
  understanding}.
\newblock In \emph{2023 IEEE International Conference on Systems, Man, and
  Cybernetics (SMC)}, pages 2151--2156.

\bibitem[{Masmoudi et~al.(2019)Masmoudi, Khmekhem, Khrouf, and
  Belguith}]{10.1145/3364319}
Abir Masmoudi, Mariem~Ellouze Khmekhem, Mourad Khrouf, and Lamia~Hadrich
  Belguith. 2019.
\newblock \href {https://doi.org/10.1145/3364319} {Transliteration of {Arabizi}
  into {Arabic} script for {Tunisian} dialect}.
\newblock \emph{ACM Trans. Asian Low-Resour. Lang. Inf. Process.}, 19(2).

\bibitem[{Mathur et~al.(2018{\natexlab{a}})Mathur, Sawhney, Ayyar, and
  Shah}]{mathur-etal-2018-offend}
Puneet Mathur, Ramit Sawhney, Meghna Ayyar, and Rajiv Shah. 2018{\natexlab{a}}.
\newblock \href {https://doi.org/10.18653/v1/W18-5118} {Did you offend me?
  classification of offensive tweets in {H}inglish language}.
\newblock In \emph{Proceedings of the 2nd Workshop on Abusive Language Online
  ({ALW}2)}, pages 138--148, Brussels, Belgium. Association for Computational
  Linguistics.

\bibitem[{Mathur et~al.(2018{\natexlab{b}})Mathur, Shah, Sawhney, and
  Mahata}]{mathur-etal-2018-detecting}
Puneet Mathur, Rajiv Shah, Ramit Sawhney, and Debanjan Mahata.
  2018{\natexlab{b}}.
\newblock \href {https://doi.org/10.18653/v1/W18-3504} {Detecting offensive
  tweets in {H}indi-{E}nglish code-switched language}.
\newblock In \emph{Proceedings of the Sixth International Workshop on Natural
  Language Processing for Social Media}, pages 18--26, Melbourne, Australia.
  Association for Computational Linguistics.

\bibitem[{Mavandadi et~al.(2023)Mavandadi, Li, Zhang, Farris, Sainath, and
  Strohman}]{10023346}
Sepand Mavandadi, Bo~Li, Chao Zhang, Brian Farris, Tara~N. Sainath, and Trevor
  Strohman. 2023.
\newblock \href {https://doi.org/10.1109/SLT54892.2023.10023346} {A truly
  multilingual first pass and monolingual second pass streaming on-device {ASR}
  system}.
\newblock In \emph{2022 IEEE Spoken Language Technology Workshop (SLT)}, pages
  838--845.

\bibitem[{May et~al.(2014)May, Benjira, and Echihabi}]{may-etal-2014-arabizi}
Jonathan May, Yassine Benjira, and Abdessamad Echihabi. 2014.
\newblock \href {https://aclanthology.org/2014.amta-researchers.25/} {An
  {A}rabizi-{E}nglish social media statistical machine translation system}.
\newblock In \emph{Proceedings of the 11th Conference of the Association for
  Machine Translation in the Americas: MT Researchers Track}, pages 329--341,
  Vancouver, Canada. Association for Machine Translation in the Americas.

\bibitem[{Mazumder et~al.(2025{\natexlab{a}})Mazumder, Kumar, and
  Patro}]{10.1145/3726866}
Debajyoti Mazumder, Aakash Kumar, and Jasabanta Patro. 2025{\natexlab{a}}.
\newblock \href {https://doi.org/10.1145/3726866} {Improving code-mixed hate
  detection by native sample mixing: A case study for {Hindi}-{English}
  code-mixed scenario}.
\newblock \emph{ACM Trans. Asian Low-Resour. Lang. Inf. Process.}, 24(5).

\bibitem[{Mazumder et~al.(2025{\natexlab{b}})Mazumder, Kumar, and
  Patro}]{mazumder-etal-2025-revealing}
Debajyoti Mazumder, Aakash Kumar, and Jasabanta Patro. 2025{\natexlab{b}}.
\newblock \href {https://doi.org/10.18653/v1/2025.findings-emnlp.1308}
  {Revealing the impact of synthetic native samples and multi-tasking
  strategies in {H}indi-{E}nglish code-mixed humour and sarcasm detection}.
\newblock In \emph{Findings of the Association for Computational Linguistics:
  EMNLP 2025}, pages 24077--24107, Suzhou, China. Association for Computational
  Linguistics.

\bibitem[{Mazumder et~al.(2026{\natexlab{a}})Mazumder, Pathak, Kodali, Joshi,
  Agarwal, and Patro}]{mazumder2026evaluating}
Debajyoti Mazumder, Divyansh Pathak, Prashant Kodali, Aditya Joshi, Akshay
  Agarwal, and Jasabanta Patro. 2026{\natexlab{a}}.
\newblock Evaluating cross-lingual knowledge consistency in code-mixed
  vis-a-vis {Indian} languages using {IndicKLAR}.
\newblock \emph{arXiv preprint arXiv:2605.29637}.

\bibitem[{Mazumder et~al.(2026{\natexlab{b}})Mazumder, Pathak, Kodali, and
  Patro}]{mazumder2026neither}
Debajyoti Mazumder, Divyansh Pathak, Prashant Kodali, and Jasabanta Patro.
  2026{\natexlab{b}}.
\newblock Neither here nor there: Cross-lingual representation dynamics of
  code-mixed text in multilingual encoders.
\newblock \emph{arXiv preprint arXiv:2603.19771}.

\bibitem[{Micallef et~al.(2024)Micallef, Habash, Borg, Eryani, and
  Bouamor}]{micallef-etal-2024-cross}
Kurt Micallef, Nizar Habash, Claudia Borg, Fadhl Eryani, and Houda Bouamor.
  2024.
\newblock \href {https://doi.org/10.18653/v1/2024.eacl-long.61} {Cross-lingual
  transfer from related languages: Treating low-resource {M}altese as
  multilingual code-switching}.
\newblock In \emph{Proceedings of the 18th Conference of the European Chapter
  of the Association for Computational Linguistics (Volume 1: Long Papers)},
  pages 1014--1025, St. Julian{'}s, Malta. Association for Computational
  Linguistics.

\bibitem[{Mitra et~al.(2026)Mitra, Chakravarty, Mondal, and
  Bandyopadhyay}]{mitra-etal-2026-llm}
Anuran Mitra, Anirvan Chakravarty, Tapabrata Mondal, and Sivaji Bandyopadhyay.
  2026.
\newblock \href {https://doi.org/10.63317/5k9xcfdyrtev} {From {LLM} prompts to
  acoustic baselines: A scalable pipeline for under-resourced disfluent
  code-mixed speech}.
\newblock In \emph{Proceedings of the SIGUL 2026 Joint Workshop with ELE,
  EURALI, and DCLRL "Towards Inclusivity and Equality: Language Resources and
  Technologies for Under-Resourced and Endangered Languages}, pages 222--232,
  Palma, Mallorca, Spain. European Language Resources Association (ELRA).

\bibitem[{Mohamed et~al.(2026)Mohamed, Zhang, Vazirgiannis, and
  Shang}]{mohamed-etal-2026-lost}
Amr Mohamed, Yang Zhang, Michalis Vazirgiannis, and Guokan Shang. 2026.
\newblock \href {https://doi.org/10.18653/v1/2026.acl-long.2080} {Lost in the
  mix: Evaluating {LLM} understanding of code-switched text}.
\newblock In \emph{Proceedings of the 64th Annual Meeting of the {A}ssociation
  for {C}omputational {L}inguistics (Volume 1: Long Papers)}, pages
  44922--44938, San Diego, California, United States. Association for
  Computational Linguistics.

\bibitem[{{Mohana Priya} et~al.(2025){Mohana Priya}, Shrinithi, Nithish, and
  Pranesh}]{mohanapriya-etal-2025-comparative}
K.~T. {Mohana Priya}, G.~Shrinithi, P.~Nithish, and A.~C. Pranesh. 2025.
\newblock \href {https://doi.org/10.63458/ijerst.v3i1.101} {Comparative
  analysis of transformer models for sentiment classification in code- mixed
  indic languages}.
\newblock \emph{International Journal of Engineering Research and Sustainable
  Technologies (IJERST)}, 3(1):1–9.

\bibitem[{Mondal et~al.(2022)Mondal, Ritika, Pathak, Jyothi, and
  Raghuveer}]{mondal-etal-2022-cocoa}
Sneha Mondal, Ritika, Shreya Pathak, Preethi Jyothi, and Aravindan Raghuveer.
  2022.
\newblock \href {https://doi.org/10.18653/v1/2022.emnlp-main.158} {{C}o{C}oa:
  An encoder-decoder model for controllable code-switched generation}.
\newblock In \emph{Proceedings of the 2022 Conference on Empirical Methods in
  Natural Language Processing}, pages 2466--2479, Abu Dhabi, United Arab
  Emirates. Association for Computational Linguistics.

\bibitem[{Muller et~al.(2020)Muller, Sagot, and Seddah}]{muller2020can}
Benjamin Muller, Benoit Sagot, and Djam{\'e} Seddah. 2020.
\newblock Can multilingual language models transfer to an unseen dialect? a
  case study on north african arabizi.
\newblock \emph{arXiv preprint arXiv:2005.00318}.

\bibitem[{N and Sooda(2025)}]{11167161}
Shruthi N and Kavitha Sooda. 2025.
\newblock \href {https://doi.org/10.1109/CONIT65521.2025.11167161} {{IndicBART}
  for translating code-mixed {Kannada}-{English} sentences into {Kannada}: An
  encoder-decoder {Transformer} approach}.
\newblock In \emph{2025 5th International Conference on Intelligent
  Technologies (CONIT)}, pages 1--6.

\bibitem[{Nagar et~al.(2025)Nagar, Khvatskii, Sokol, and
  Chawla}]{nagar2025breaking}
Tanay Nagar, Grigorii Khvatskii, Anna Sokol, and Nitesh~V Chawla. 2025.
\newblock Breaking language barriers: Equitable performance in multilingual
  language models.
\newblock \emph{arXiv preprint arXiv:2508.12662}.

\bibitem[{Nakayama et~al.(2019)Nakayama, Kano, Tjandra, Sakti, and
  Nakamura}]{9060847}
Sahoko Nakayama, Takatomo Kano, Andros Tjandra, Sakriani Sakti, and Satoshi
  Nakamura. 2019.
\newblock \href {https://doi.org/10.1109/O-COCOSDA46868.2019.9060847}
  {Recognition and translation of code-switching speech utterances}.
\newblock In \emph{2019 22nd Conference of the Oriental COCOSDA International
  Committee for the Co-ordination and Standardisation of Speech Databases and
  Assessment Techniques (O-COCOSDA)}, pages 1--6.

\bibitem[{Naowarat et~al.(2021)Naowarat, Kongthaworn, Karunratanakul, Wu, and
  Chuangsuwanich}]{9413806}
Burin Naowarat, Thananchai Kongthaworn, Korrawe Karunratanakul, Sheng~Hui Wu,
  and Ekapol Chuangsuwanich. 2021.
\newblock \href {https://doi.org/10.1109/ICASSP39728.2021.9413806} {Reducing
  spelling inconsistencies in code-switching {ASR} using contextualized {CTC}
  loss}.
\newblock In \emph{ICASSP 2021 - 2021 IEEE International Conference on
  Acoustics, Speech and Signal Processing (ICASSP)}, pages 6239--6243.

\bibitem[{Nayak and Joshi(2022)}]{nayak-joshi-2022-l3cube}
Ravindra Nayak and Raviraj Joshi. 2022.
\newblock \href {https://aclanthology.org/2022.wildre-1.2/}
  {{L}3{C}ube-{H}ing{C}orpus and {H}ing{BERT}: A code mixed {H}indi-{E}nglish
  dataset and {BERT} language models}.
\newblock In \emph{Proceedings of the WILDRE-6 Workshop within the 13th
  Language Resources and Evaluation Conference}, pages 7--12, Marseille,
  France. European Language Resources Association.

\bibitem[{Nekvinda and Dušek(2020)}]{nekvinda20_interspeech}
Tomáš Nekvinda and Ondřej Dušek. 2020.
\newblock \href {https://doi.org/10.21437/Interspeech.2020-2679} {{One Model,
  Many Languages: Meta-Learning for Multilingual Text-to-Speech}}.
\newblock In \emph{{Interspeech 2020}}, pages 2972--2976.

\bibitem[{Ng and Chan(2024)}]{ng2024talking}
Lynnette Hui~Xian Ng and Luo~Qi Chan. 2024.
\newblock What talking you?: Translating code-mixed messaging texts to english.
\newblock \emph{arXiv preprint arXiv:2411.05253}.

\bibitem[{Nguyen and Bryant(2020)}]{nguyen-bryant-2020-canvec}
Li~Nguyen and Christopher Bryant. 2020.
\newblock \href {https://aclanthology.org/2020.lrec-1.507/} {{C}an{VEC} {--}
  the {C}anberra {V}ietnamese-{E}nglish code-switching natural speech corpus}.
\newblock In \emph{Proceedings of the Twelfth Language Resources and Evaluation
  Conference}, pages 4121--4129, Marseille, France. European Language Resources
  Association.

\bibitem[{Oh et~al.(2026)Oh, Yoo, Haznitrama, and Oh}]{oh-etal-2026-ola}
Juhyun Oh, Haneul Yoo, Faiz~Ghifari Haznitrama, and Alice Oh. 2026.
\newblock \href {https://doi.org/10.18653/v1/2026.acl-long.2162} {{OLA}: Output
  language alignment in code-switched {LLM} interactions}.
\newblock In \emph{Proceedings of the 64th Annual Meeting of the {A}ssociation
  for {C}omputational {L}inguistics (Volume 1: Long Papers)}, pages
  46603--46624, San Diego, California, United States. Association for
  Computational Linguistics.

\bibitem[{Ojo et~al.(2025)Ojo, Kamel, and Adelani}]{ojo2025divers}
Jessica Ojo, Zina Kamel, and David~Ifeoluwa Adelani. 2025.
\newblock Divers-bench: Evaluating language identification across domain shifts
  and code-switching.
\newblock \emph{arXiv preprint arXiv:2509.17768}.

\bibitem[{Okocha and Grant(2026)}]{okocha2026beyond}
Chibuzor Okocha and Christan~Earl Grant. 2026.
\newblock Beyond word error rate: A switch aware evaluation of {ASR} and audio
  language models on {English} {Yoruba} code-switched speech.
\newblock \emph{arXiv preprint arXiv:2609.11786}.

\bibitem[{Ou and Xu(2024)}]{10692736}
Fangwei Ou and Jinan Xu. 2024.
\newblock \href {https://doi.org/10.1109/ICNLP60986.2024.10692736} {Enhancing
  multilingual universal sentence embeddings by monolingual contrastive
  learning}.
\newblock In \emph{2024 6th International Conference on Natural Language
  Processing (ICNLP)}, pages 36--40.

\bibitem[{Paik et~al.(2026{\natexlab{a}})Paik, Kim, Lee, Ahn, and
  Kim}]{paik-etal-2026-hike}
Gio Paik, Yongbeom Kim, Soungmin Lee, Sangmin Ahn, and Chan~Woo Kim.
  2026{\natexlab{a}}.
\newblock \href {https://doi.org/10.18653/v1/2026.findings-eacl.33} {{H}i{KE}:
  Hierarchical evaluation framework for {K}orean-{E}nglish code-switching
  speech recognition}.
\newblock In \emph{Findings of the {A}ssociation for {C}omputational
  {L}inguistics: {EACL} 2026}, pages 673--681, Rabat, Morocco. Association for
  Computational Linguistics.

\bibitem[{Paik et~al.(2026{\natexlab{b}})Paik, Shin, and Lee}]{paik2026towards}
Gio Paik, Hyunseo Shin, and Soungmin Lee. 2026{\natexlab{b}}.
\newblock Towards truly multilingual {ASR}: Generalizing code-switching {ASR}
  to unseen language pairs.
\newblock \emph{arXiv preprint arXiv:2606.05846}.

\bibitem[{Pamungkas et~al.(2024)Pamungkas, Purworini, Priyawati, and
  Chasana}]{10.1145/3639233.3639247}
Endang~Wahyu Pamungkas, Dian Purworini, Diah Priyawati, and Rona Rizkhy~Bunga
  Chasana. 2024.
\newblock \href {https://doi.org/10.1145/3639233.3639247} {Exploring the impact
  of lexicon-based knowledge transfer for hate speech detection in indonesia
  code-mixed languages}.
\newblock In \emph{Proceedings of the 2023 7th International Conference on
  Natural Language Processing and Information Retrieval}, NLPIR '23, page
  85–90, New York, NY, USA. Association for Computing Machinery.

\bibitem[{Pan et~al.(2021)Pan, Hang, Qi, Shah, Potdar, and
  Yu}]{pan-etal-2021-multilingual}
Lin Pan, Chung-Wei Hang, Haode Qi, Abhishek Shah, Saloni Potdar, and Mo~Yu.
  2021.
\newblock \href {https://doi.org/10.18653/v1/2021.naacl-main.20} {Multilingual
  {BERT} post-pretraining alignment}.
\newblock In \emph{Proceedings of the 2021 Conference of the North American
  Chapter of the Association for Computational Linguistics: Human Language
  Technologies}, pages 210--219, Online. Association for Computational
  Linguistics.

\bibitem[{Pandey et~al.(2017)Pandey, Srivastava, and Gangashetty}]{8300583}
Ayushi Pandey, Brij Mohan~Lai Srivastava, and Suryakanth~V Gangashetty. 2017.
\newblock \href {https://doi.org/10.1109/IALP.2017.8300583} {Adapting
  monolingual resources for code-mixed hindi-english speech recognition}.
\newblock In \emph{2017 International Conference on Asian Language Processing
  (IALP)}, pages 218--221.

\bibitem[{Peng et~al.(2014)Peng, Wang, and Dredze}]{peng-etal-2014-learning}
Nanyun Peng, Yiming Wang, and Mark Dredze. 2014.
\newblock \href {https://doi.org/10.3115/v1/P14-2110} {Learning polylingual
  topic models from code-switched social media documents}.
\newblock In \emph{Proceedings of the 52nd Annual Meeting of the Association
  for Computational Linguistics (Volume 2: Short Papers)}, pages 674--679,
  Baltimore, Maryland. Association for Computational Linguistics.

\bibitem[{Peng et~al.(2023)Peng, Yan, Watanabe, and
  Harwath}]{peng2023prompting}
Puyuan Peng, Brian Yan, Shinji Watanabe, and David Harwath. 2023.
\newblock Prompting the hidden talent of web-scale speech models for zero-shot
  task generalization.
\newblock \emph{arXiv preprint arXiv:2305.11095}.

\bibitem[{Peng et~al.(2022)Peng, Liu, Zhang, Xu, He, Huang, and
  Chng}]{peng2022internal}
Yizhou Peng, Yufei Liu, Jicheng Zhang, Haihua Xu, Yi~He, Hao Huang, and
  Eng~Siong Chng. 2022.
\newblock Internal language model estimation based language model fusion for
  cross-domain code-switching speech recognition.
\newblock \emph{arXiv preprint arXiv:2207.04176}.

\bibitem[{P{\'e}rez et~al.(2022)P{\'e}rez, Furman, Alonso~Alemany, and
  Luque}]{perez-etal-2022-robertuito}
Juan~Manuel P{\'e}rez, Dami{\'a}n~Ariel Furman, Laura Alonso~Alemany, and
  Franco~M. Luque. 2022.
\newblock \href {https://aclanthology.org/2022.lrec-1.785/} {{R}o{BERT}uito: a
  pre-trained language model for social media text in {S}panish}.
\newblock In \emph{Proceedings of the Thirteenth Language Resources and
  Evaluation Conference}, pages 7235--7243, Marseille, France. European
  Language Resources Association.

\bibitem[{Pires et~al.(2019)Pires, Schlinger, and
  Garrette}]{pires-etal-2019-multilingual}
Telmo Pires, Eva Schlinger, and Dan Garrette. 2019.
\newblock \href {https://doi.org/10.18653/v1/P19-1493} {How multilingual is
  multilingual {BERT}?}
\newblock In \emph{Proceedings of the 57th Annual Meeting of the Association
  for Computational Linguistics}, pages 4996--5001, Florence, Italy.
  Association for Computational Linguistics.

\bibitem[{Pratapa et~al.(2018{\natexlab{a}})Pratapa, Bhat, Choudhury, Sitaram,
  Dandapat, and Bali}]{pratapa-etal-2018-language}
Adithya Pratapa, Gayatri Bhat, Monojit Choudhury, Sunayana Sitaram, Sandipan
  Dandapat, and Kalika Bali. 2018{\natexlab{a}}.
\newblock \href {https://doi.org/10.18653/v1/P18-1143} {Language modeling for
  code-mixing: The role of linguistic theory based synthetic data}.
\newblock In \emph{Proceedings of the 56th Annual Meeting of the Association
  for Computational Linguistics (Volume 1: Long Papers)}, pages 1543--1553,
  Melbourne, Australia. Association for Computational Linguistics.

\bibitem[{Pratapa and Choudhury(2021)}]{pratapa-choudhury-2021-comparing}
Adithya Pratapa and Monojit Choudhury. 2021.
\newblock \href {https://doi.org/10.18653/v1/2021.wnut-1.18} {Comparing
  grammatical theories of code-mixing}.
\newblock In \emph{Proceedings of the Seventh Workshop on Noisy User-generated
  Text (W-NUT 2021)}, pages 158--167, Online. Association for Computational
  Linguistics.

\bibitem[{Pratapa et~al.(2018{\natexlab{b}})Pratapa, Choudhury, and
  Sitaram}]{pratapa-etal-2018-word}
Adithya Pratapa, Monojit Choudhury, and Sunayana Sitaram. 2018{\natexlab{b}}.
\newblock \href {https://doi.org/10.18653/v1/D18-1344} {Word embeddings for
  code-mixed language processing}.
\newblock In \emph{Proceedings of the 2018 Conference on Empirical Methods in
  Natural Language Processing}, pages 3067--3072, Brussels, Belgium.
  Association for Computational Linguistics.

\bibitem[{Prioleau and Aryal(2023)}]{10590279}
Howard Prioleau and Saurav~K. Aryal. 2023.
\newblock \href {https://doi.org/10.1109/CSCI62032.2023.00036} {Benchmarking
  current state-of-the-art {Transformer} models on token level language
  identification and language pair identification}.
\newblock In \emph{2023 International Conference on Computational Science and
  Computational Intelligence (CSCI)}, pages 193--199.

\bibitem[{Priyadharshini et~al.(2023)Priyadharshini, Chakravarthi, Subramanian,
  Chinnaudayar~Navaneethakrishnan, Shanmugavadivel, B, Murugappan, Kumaresan,
  Sai~Prashanth, Sai Rishith~Reddy, and
  Chandu}]{priyadharshini-etal-2023-overview}
Ruba Priyadharshini, Bharathi~Raja Chakravarthi, Malliga Subramanian,
  Subalalitha Chinnaudayar~Navaneethakrishnan, Kogilavani Shanmugavadivel,
  Premjith B, Abirami Murugappan, Prasanna~Kumar Kumaresan, Karnati
  Sai~Prashanth, Mangamuru Sai Rishith~Reddy, and Janakiram Chandu. 2023.
\newblock \href {https://aclanthology.org/2023.dravidianlangtech-1.11/}
  {Overview of shared-task on abusive comment detection in {T}amil and
  {T}elugu}.
\newblock In \emph{Proceedings of the Third Workshop on Speech and Language
  Technologies for Dravidian Languages}, pages 80--87, Varna, Bulgaria. INCOMA
  Ltd., Shoumen, Bulgaria.

\bibitem[{Qin et~al.(2024)Qin, Chen, Zhou, Chen, Li, Liao, Li, Che, and
  Yu}]{qin2024multilingual}
Libo Qin, Qiguang Chen, Yuhang Zhou, Zhi Chen, Yinghui Li, Lizi Liao, Min Li,
  Wanxiang Che, and Philip~S Yu. 2024.
\newblock Multilingual large language model: A survey of resources, taxonomy
  and frontiers.
\newblock \emph{arXiv preprint arXiv:2404.04925}.

\bibitem[{Qin et~al.(2020)Qin, Ni, Zhang, and Che}]{ijcai2020p533}
Libo Qin, Minheng Ni, Yue Zhang, and Wanxiang Che. 2020.
\newblock \href {https://doi.org/10.24963/ijcai.2020/533} {{CoSDA}-{ML}:
  Multi-lingual code-switching data augmentation for zero-shot cross-lingual
  {NLP}}.
\newblock In \emph{Proceedings of the Twenty-Ninth International Joint
  Conference on Artificial Intelligence, {IJCAI-20}}, pages 3853--3860.
  International Joint Conferences on Artificial Intelligence Organization.
\newblock Main track.

\bibitem[{Raheem et~al.(2022)Raheem, Elrashid, Johnson, and
  Kreutzer}]{raheem-etal-2022-domain}
Lekan Raheem, Maab Elrashid, Melvin Johnson, and Julia Kreutzer. 2022.
\newblock \href {https://doi.org/10.18653/v1/2022.wmt-1.117} {Domain curricula
  for code-switched {MT} at {M}ix{MT} 2022}.
\newblock In \emph{Proceedings of the Seventh Conference on Machine Translation
  (WMT)}, pages 1162--1169, Abu Dhabi, United Arab Emirates (Hybrid).
  Association for Computational Linguistics.

\bibitem[{Raihan et~al.(2023)Raihan, Goswami, and Mahmud}]{raihan2023mixed}
Md~Nishat Raihan, Dhiman Goswami, and Antara Mahmud. 2023.
\newblock Mixed-distil-bert: Code-mixed language modeling for bangla, english,
  and hindi.
\newblock \emph{arXiv preprint arXiv:2309.10272}.

\bibitem[{Raj~Khan et~al.(2021)Raj~Khan, Gupta, and
  Ekbal}]{raj-khan-etal-2021-towards-developing}
Humair Raj~Khan, Deepak Gupta, and Asif Ekbal. 2021.
\newblock \href {https://doi.org/10.18653/v1/2021.findings-emnlp.151} {Towards
  developing a multilingual and code-mixed visual question answering system by
  knowledge distillation}.
\newblock In \emph{Findings of the Association for Computational Linguistics:
  EMNLP 2021}, pages 1753--1767, Punta Cana, Dominican Republic. Association
  for Computational Linguistics.

\bibitem[{Rallabandi and Black(2017)}]{rallabandi17_interspeech}
SaiKrishna Rallabandi and Alan~W. Black. 2017.
\newblock \href {https://doi.org/10.21437/Interspeech.2017-1244} {{On Building
  Mixed Lingual Speech Synthesis Systems}}.
\newblock In \emph{{Interspeech 2017}}, pages 52--56.

\bibitem[{Ran et~al.(2024)Ran, Zhang, Wang, Xu, and Zhang}]{10651379}
Xuqiao Ran, You Zhang, Jin Wang, Dan Xu, and Xuejie Zhang. 2024.
\newblock \href {https://doi.org/10.1109/IJCNN60899.2024.10651379}
  {{LoRA}-enhanced language alignments for robust code-mixed text
  representation}.
\newblock In \emph{2024 International Joint Conference on Neural Networks
  (IJCNN)}, pages 1--8.

\bibitem[{Ranjan et~al.(2016)Ranjan, Raja, Priyadharshini, and
  Balabantaray}]{7918035}
Prakash Ranjan, Bharathi Raja, Ruba Priyadharshini, and Rakesh~Chandra
  Balabantaray. 2016.
\newblock \href {https://doi.org/10.1109/IC3I.2016.7918035} {A comparative
  study on code-mixed data of {Indian} social media vs formal text}.
\newblock In \emph{2016 2nd International Conference on Contemporary Computing
  and Informatics (IC3I)}, pages 608--611.

\bibitem[{Reddy and Rao(2018)}]{8554693}
M.~Kiran Reddy and K.~Sreenivasa Rao. 2018.
\newblock \href {https://doi.org/10.1109/ICACCI.2018.8554693} {{DNN}-based
  bilingual ({Telugu}-{Hindi}) polyglot speech synthesis}.
\newblock In \emph{2018 International Conference on Advances in Computing,
  Communications and Informatics (ICACCI)}, pages 1808--1811.

\bibitem[{Rizvi et~al.(2021)Rizvi, Srinivasan, Ganu, Choudhury, and
  Sitaram}]{rizvi-etal-2021-gcm}
Mohd Sanad~Zaki Rizvi, Anirudh Srinivasan, Tanuja Ganu, Monojit Choudhury, and
  Sunayana Sitaram. 2021.
\newblock \href {https://doi.org/10.18653/v1/2021.eacl-demos.24} {{GCM}: A
  toolkit for generating synthetic code-mixed text}.
\newblock In \emph{Proceedings of the 16th Conference of the European Chapter
  of the Association for Computational Linguistics: System Demonstrations},
  pages 205--211, Online. Association for Computational Linguistics.

\bibitem[{Robinson et~al.(2022)Robinson, Hogan, Fulda, and
  Mortensen}]{robinson-etal-2022-data}
Nathaniel Robinson, Cameron Hogan, Nancy Fulda, and David~R. Mortensen. 2022.
\newblock \href {https://aclanthology.org/2022.loresmt-1.5/} {Data-adaptive
  transfer learning for translation: A case study in {H}aitian and {Jamaican}}.
\newblock In \emph{Proceedings of the Fifth Workshop on Technologies for
  Machine Translation of Low-Resource Languages (LoResMT 2022)}, pages 35--42,
  Gyeongju, Republic of Korea. Association for Computational Linguistics.

\bibitem[{Rooryck et~al.(2026)Rooryck, Cai, Belinkov, Alvarez-Melis, and
  Brantley}]{rooryck2026feeding}
Dries Rooryck, Alex Cai, Yonatan Belinkov, David Alvarez-Melis, and Kiant{\'e}
  Brantley. 2026.
\newblock Feeding {BabyLMs} macaroni: Code-switching curricula cause
  cross-lingual convergence.
\newblock \emph{arXiv preprint arXiv:2609.30535}.

\bibitem[{Sai and Sharma(2021)}]{sai-sharma-2021-towards}
Siva Sai and Yashvardhan Sharma. 2021.
\newblock \href {https://aclanthology.org/2021.dravidianlangtech-1.3/} {Towards
  offensive language identification for {D}ravidian languages}.
\newblock In \emph{Proceedings of the First Workshop on Speech and Language
  Technologies for Dravidian Languages}, pages 18--27, Kyiv. Association for
  Computational Linguistics.

\bibitem[{Sailor et~al.(2021)Sailor, T, Agrawal, Jain, and
  Pandey}]{sailor21_interspeech}
Hardik Sailor, Kiran~Praveen T, Vikas Agrawal, Abhinav Jain, and Abhishek
  Pandey. 2021.
\newblock \href {https://doi.org/10.21437/Interspeech.2021-1578} {{SRI-B
  End-to-End System for Multilingual and Code-Switching ASR Challenges for Low
  Resource Indian Languages}}.
\newblock In \emph{{Interspeech 2021}}, pages 2456--2460.

\bibitem[{Samanta et~al.(2019{\natexlab{a}})Samanta, Ganguly, and
  Chakrabarti}]{samanta-etal-2019-improved}
Bidisha Samanta, Niloy Ganguly, and Soumen Chakrabarti. 2019{\natexlab{a}}.
\newblock \href {https://doi.org/10.18653/v1/P19-1343} {Improved sentiment
  detection via label transfer from monolingual to synthetic code-switched
  text}.
\newblock In \emph{Proceedings of the 57th Annual Meeting of the Association
  for Computational Linguistics}, pages 3528--3537, Florence, Italy.
  Association for Computational Linguistics.

\bibitem[{Samanta et~al.(2019{\natexlab{b}})Samanta, Reddy, Jagirdar, Ganguly,
  and Chakrabarti}]{ijcai2019p719}
Bidisha Samanta, Sharmila Reddy, Hussain Jagirdar, Niloy Ganguly, and Soumen
  Chakrabarti. 2019{\natexlab{b}}.
\newblock \href {https://doi.org/10.24963/ijcai.2019/719} {A deep generative
  model for code switched text}.
\newblock In \emph{Proceedings of the Twenty-Eighth International Joint
  Conference on Artificial Intelligence, {IJCAI-19}}, pages 5175--5181.
  International Joint Conferences on Artificial Intelligence Organization.

\bibitem[{Santy et~al.(2021)Santy, Srinivasan, and
  Choudhury}]{santy-etal-2021-bertologicomix}
Sebastin Santy, Anirudh Srinivasan, and Monojit Choudhury. 2021.
\newblock \href {https://aclanthology.org/2021.adaptnlp-1.12/}
  {{BERT}ologi{C}o{M}ix: How does code-mixing interact with multilingual
  {BERT}?}
\newblock In \emph{Proceedings of the Second Workshop on Domain Adaptation for
  NLP}, pages 111--121, Kyiv, Ukraine. Association for Computational
  Linguistics.

\bibitem[{Senevirathna et~al.(2025)Senevirathna, Adhikari, Navojith, Rizvi,
  Kasthurirathna, and Abeywardhana}]{11361448}
Prasanthika Senevirathna, Nipuna Adhikari, Thamindu Navojith, Azmarah Rizvi,
  Dharshana Kasthurirathna, and Lakmini Abeywardhana. 2025.
\newblock \href {https://doi.org/10.1109/ICAC69156.2025.11361448} {Explainable
  multilingual sentiment analysis for {Sinhala}, {English} and code-mixed
  banking reviews}.
\newblock In \emph{2025 7th International Conference on Advancements in
  Computing (ICAC)}, pages 1--6.

\bibitem[{Seong et~al.(2025)Seong, Kim, Ryu, and Lee}]{10929894}
Hyeonggi Seong, Nam-Joon Kim, Hyun~Gon Ryu, and Hyuk-Jae Lee. 2025.
\newblock \href {https://doi.org/10.1109/ICCE63647.2025.10929894} {Extending
  {Whisper} for {Korean}-{English} code-switching speech recognition}.
\newblock In \emph{2025 IEEE International Conference on Consumer Electronics
  (ICCE)}, pages 1--4.

\bibitem[{Sequiera et~al.(2014)Sequiera, Rao, and
  Shambavi}]{10.1145/2824864.2824871}
Royal~Denzil Sequiera, Shashank~S Rao, and B.~R. Shambavi. 2014.
\newblock \href {https://doi.org/10.1145/2824864.2824871} {Word-level language
  identification and back transliteration of romanized text}.
\newblock In \emph{Proceedings of the 6th Annual Meeting of the Forum for
  Information Retrieval Evaluation}, FIRE '14, page 70–73, New York, NY, USA.
  Association for Computing Machinery.

\bibitem[{Shah et~al.(2020)Shah, Abraham, Sitaram, Joshi
  et~al.}]{shah2020learning}
Sanket Shah, Basil Abraham, Sunayana Sitaram, Vikas Joshi, et~al. 2020.
\newblock Learning to recognize code-switched speech without forgetting
  monolingual speech recognition.
\newblock \emph{arXiv preprint arXiv:2006.00782}.

\bibitem[{Shan et~al.(2019)Shan, Weng, Wang, Su, Luo, Yu, and Xie}]{8682850}
Changhao Shan, Chao Weng, Guangsen Wang, Dan Su, Min Luo, Dong Yu, and Lei Xie.
  2019.
\newblock \href {https://doi.org/10.1109/ICASSP.2019.8682850} {Investigating
  end-to-end speech recognition for {Mandarin}-english code-switching}.
\newblock In \emph{ICASSP 2019 - 2019 IEEE International Conference on
  Acoustics, Speech and Signal Processing (ICASSP)}, pages 6056--6060.

\bibitem[{Shankar et~al.(2024)Shankar, Jyothi, and
  Bhattacharyya}]{shankar-etal-2024-context}
Bhavani Shankar, Preethi Jyothi, and Pushpak Bhattacharyya. 2024.
\newblock \href {https://doi.org/10.18653/v1/2024.acl-long.228} {In-context
  mixing ({ICM}): Code-mixed prompts for multilingual {LLM}s}.
\newblock In \emph{Proceedings of the 62nd Annual Meeting of the Association
  for Computational Linguistics (Volume 1: Long Papers)}, pages 4162--4176,
  Bangkok, Thailand. Association for Computational Linguistics.

\bibitem[{Shao et~al.(2026)Shao, Tang, Zhang, Sevegnani, Stenetorp, Yang, and
  Lu}]{shao-etal-2026-role}
Jiandong Shao, Raphael Tang, Crystina Zhang, Karin Sevegnani, Pontus Stenetorp,
  Jianfei Yang, and Yao Lu. 2026.
\newblock \href {https://doi.org/10.18653/v1/2026.acl-long.1706} {The role of
  mixed-language documents for multilingual large language model pretraining}.
\newblock In \emph{Proceedings of the 64th Annual Meeting of the {A}ssociation
  for {C}omputational {L}inguistics (Volume 1: Long Papers)}, pages
  36807--36818, San Diego, California, United States. Association for
  Computational Linguistics.

\bibitem[{Sharif et~al.(2021)Sharif, Hossain, and Hoque}]{sharif-etal-2021-nlp}
Omar Sharif, Eftekhar Hossain, and Mohammed~Moshiul Hoque. 2021.
\newblock \href {https://aclanthology.org/2021.dravidianlangtech-1.35/}
  {{NLP}-{CUET}@{D}ravidian{L}ang{T}ech-{EACL}2021: Offensive language
  detection from multilingual code-mixed text using {Transformers}}.
\newblock In \emph{Proceedings of the First Workshop on Speech and Language
  Technologies for Dravidian Languages}, pages 255--261, Kyiv. Association for
  Computational Linguistics.

\bibitem[{Sharma et~al.(2022)Sharma, Kabra, and Jain}]{SHARMA2022102760}
Arushi Sharma, Anubha Kabra, and Minni Jain. 2022.
\newblock \href {https://doi.org/10.1016/j.ipm.2021.102760} {Ceasing hate with
  {MoH}: Hate speech detection in {Hindi}–{English} code-switched language}.
\newblock \emph{Information Processing \& Management}, 59(1):102760.

\bibitem[{Sharma et~al.(2020)Sharma, Abraham, Taneja, and
  Jyothi}]{sharma20c_interspeech}
Yash Sharma, Basil Abraham, Karan Taneja, and Preethi Jyothi. 2020.
\newblock \href {https://doi.org/10.21437/Interspeech.2020-2402} {{Improving
  Low Resource Code-Switched ASR Using Augmented Code-Switched TTS}}.
\newblock In \emph{{Interspeech 2020}}, pages 4771--4775.

\bibitem[{Shastri et~al.(2022)Shastri, Patil, Wanere, Mahajan, Bhatt, and
  Sailor}]{shastri2022adversarial}
Parth Shastri, Chirag Patil, Poorval Wanere, Shrinivas Mahajan, Abhishek Bhatt,
  and Hardik Sailor. 2022.
\newblock Adversarial synthesis based data-augmentation for code-switched
  spoken language identification.
\newblock \emph{arXiv preprint arXiv:2205.15747}.

\bibitem[{Shazal et~al.(2020)Shazal, Usman, and
  Habash}]{shazal-etal-2020-unified}
Ali Shazal, Aiza Usman, and Nizar Habash. 2020.
\newblock \href {https://aclanthology.org/2020.wanlp-1.15/} {A unified model
  for {A}rabizi detection and transliteration using sequence-to-sequence
  models}.
\newblock In \emph{Proceedings of the Fifth Arabic Natural Language Processing
  Workshop}, pages 167--177, Barcelona, Spain (Online). Association for
  Computational Linguistics.

\bibitem[{Sheng et~al.(2025)Sheng, Han, Li, Zhang, Huang, Lang, and
  Liu}]{sheng-etal-2025-test}
Dongming Sheng, Kexin Han, Hao Li, Yan Zhang, Yucheng Huang, Jun Lang, and
  Wenqiang Liu. 2025.
\newblock \href {https://doi.org/10.18653/v1/2025.naacl-long.260} {Test-time
  code-switching for cross-lingual aspect sentiment triplet extraction}.
\newblock In \emph{Proceedings of the 2025 Conference of the Nations of the
  Americas Chapter of the Association for Computational Linguistics: Human
  Language Technologies (Volume 1: Long Papers)}, pages 5041--5053,
  Albuquerque, New Mexico. Association for Computational Linguistics.

\bibitem[{Sheokand and Sawant(2025)}]{sheokand2025codemixbench}
Manik Sheokand and Parth Sawant. 2025.
\newblock {CodeMixBench}: Evaluating large language models on code generation
  with code-mixed prompts.
\newblock \emph{arXiv preprint arXiv:2505.05063}.

\bibitem[{Sheth et~al.(2026)Sheth, Sinha, Patil, Beniwal, and
  Singh}]{sheth-etal-2026-beyond}
Rajvee Sheth, Samridhi~Raj Sinha, Mahavir Patil, Himanshu Beniwal, and Mayank
  Singh. 2026.
\newblock \href {https://doi.org/10.18653/v1/2026.acl-long.386} {Beyond
  monolingual assumptions: A survey on code-switched {NLP} in the era of large
  language models across modalities}.
\newblock In \emph{Proceedings of the 64th Annual Meeting of the {A}ssociation
  for {C}omputational {L}inguistics (Volume 1: Long Papers)}, pages 8519--8566,
  San Diego, California, United States. Association for Computational
  Linguistics.

\bibitem[{Si et~al.(2019)Si, Datta, Banerjee, and Naskar}]{8944868}
Shukrity Si, Anisha Datta, Somnath Banerjee, and Sudip~Kumar Naskar. 2019.
\newblock \href {https://doi.org/10.1109/ICCCNT45670.2019.8944868} {Aggression
  detection on multilingual social media text}.
\newblock In \emph{2019 10th International Conference on Computing,
  Communication and Networking Technologies (ICCCNT)}, pages 1--5.

\bibitem[{Sifat et~al.(2025)Sifat, Arko, Afrin, Kabir, Akand, and
  Ahmmed}]{11525909}
Hasin~Almas Sifat, Koushik~Biswas Arko, Abida Afrin, K.~M.~Tahsin Kabir,
  Abdullah~Rakib Akand, and Md.~Mortuza Ahmmed. 2025.
\newblock \href {https://doi.org/10.1109/WIECON-ECE69386.2025.11525909}
  {Capturing dialectal variation in code-mixed {Banglish} through multi-task
  {Transformers}}.
\newblock In \emph{2025 IEEE International Women in Engineering (WIE)
  Conference on Electrical and Computer Engineering (WIECON-ECE)}, pages
  614--619.

\bibitem[{Singh et~al.(2025{\natexlab{a}})Singh, Demuynck, and
  Arora}]{singh25d_interspeech}
Anup Singh, Kris Demuynck, and Vipul Arora. 2025{\natexlab{a}}.
\newblock \href {https://doi.org/10.21437/Interspeech.2025-2722}
  {{Language-Agnostic Speech Tokenizer for Spoken Term Detection with Efficient
  Retrieval}}.
\newblock In \emph{{Interspeech 2025}}, pages 2630--2634.

\bibitem[{Singh(2025)}]{10.1145/3746027.3764194}
Maneesh~Kumar Singh. 2025.
\newblock \href {https://doi.org/10.1145/3746027.3764194} {Sovereign \& shared:
  Frugally scalable multilingual-multimodal {AI} for bharat}.
\newblock In \emph{Proceedings of the 33rd ACM International Conference on
  Multimedia}, MM '25, page 14373, New York, NY, USA. Association for Computing
  Machinery.

\bibitem[{Singh and Lefever(2020)}]{singh-lefever-2020-sentiment}
Pranaydeep Singh and Els Lefever. 2020.
\newblock \href {https://aclanthology.org/2020.calcs-1.6/} {Sentiment analysis
  for {H}inglish code-mixed tweets by means of cross-lingual word embeddings}.
\newblock In \emph{Proceedings of the 4th Workshop on Computational Approaches
  to Code Switching}, pages 45--51, Marseille, France. European Language
  Resources Association.

\bibitem[{Singh et~al.(2023)Singh, Choudhary, and
  Shrivastava}]{10.1007/978-3-031-23793-5_30}
Rajat Singh, Nurendra Choudhary, and Manish Shrivastava. 2023.
\newblock Automatic normalization of word variations in code-mixed social
  media text.
\newblock In \emph{Computational Linguistics and Intelligent Text Processing},
  pages 371--381, Cham. Springer Nature Switzerland.

\bibitem[{Singh et~al.(2025{\natexlab{b}})Singh, Singh, and
  Kadyan}]{SINGH2025111886}
Shruti Singh, Muskaan Singh, and Virender Kadyan. 2025{\natexlab{b}}.
\newblock \href {https://doi.org/10.1016/j.dib.2025.111886} {{HiACC}:
  {Hinglish} adult \& children code-switched corpus}.
\newblock \emph{Data in Brief}, 62:111886.

\bibitem[{Sitaram and Black(2016)}]{sitaram-black-2016-speech}
Sunayana Sitaram and Alan~W Black. 2016.
\newblock \href {https://aclanthology.org/L16-1546/} {Speech synthesis of
  code-mixed text}.
\newblock In \emph{Proceedings of the Tenth International Conference on
  Language Resources and Evaluation ({LREC}'16)}, pages 3422--3428,
  Portoro{\v{z}}, Slovenia. European Language Resources Association (ELRA).

\bibitem[{Sitaram et~al.(2019)Sitaram, Chandu, Rallabandi, and
  Black}]{sitaram2019survey}
Sunayana Sitaram, Khyathi~Raghavi Chandu, Sai~Krishna Rallabandi, and Alan~W
  Black. 2019.
\newblock A survey of code-switched speech and language processing.
\newblock \emph{arXiv preprint arXiv:1904.00784}.

\bibitem[{Sitaram et~al.(2016)Sitaram, Rallabandi, Rijhwani, and
  Black}]{sitaram16_ssw}
Sunayana Sitaram, Sai~Krishna Rallabandi, Shruti Rijhwani, and Alan~W. Black.
  2016.
\newblock \href {https://doi.org/10.21437/SSW.2016-13} {{Experiments with
  Cross-lingual Systems for Synthesis of Code-Mixed Text}}.
\newblock In \emph{{9th ISCA Workshop on Speech Synthesis Workshop (SSW 9)}},
  pages 76--81.

\bibitem[{Sivasankaran et~al.(2018)Sivasankaran, Srivastava, Sitaram, Bali, and
  Choudhury}]{sivasankaran-etal-2018-phone}
Sunit Sivasankaran, Brij Mohan~Lal Srivastava, Sunayana Sitaram, Kalika Bali,
  and Monojit Choudhury. 2018.
\newblock \href {https://doi.org/10.18653/v1/W18-3202} {Phone merging for
  code-switched speech recognition}.
\newblock In \emph{Proceedings of the Third Workshop on Computational
  Approaches to Linguistic Code-Switching}, pages 11--19, Melbourne, Australia.
  Association for Computational Linguistics.

\bibitem[{Slottje et~al.(2021)Slottje, Wotherspoon, Hartmann, Snover, and
  Kimball}]{slottje2021using}
Andrew Slottje, Shannon Wotherspoon, William Hartmann, Matthew Snover, and Owen
  Kimball. 2021.
\newblock Using heterogeneity in semi-supervised transcription hypotheses to
  improve code-switched speech recognition.
\newblock \emph{arXiv preprint arXiv:2106.07699}.

\bibitem[{{\v{S}}m{\'\i}d et~al.(2026){\v{S}}m{\'\i}d, P{\v{r}}ib{\'a}{\v{n}},
  and Kr{\'a}l}]{vsmid2026generative}
Jakub {\v{S}}m{\'\i}d, Pavel P{\v{r}}ib{\'a}{\v{n}}, and Pavel Kr{\'a}l. 2026.
\newblock Generative models enhanced by sequence labelling and aspect-code
  switching improve cross-lingual aspect-based sentiment analysis.
\newblock \emph{arXiv preprint arXiv:2608.30425}.

\bibitem[{Solorio and Liu(2008)}]{solorio-liu-2008-learning}
Thamar Solorio and Yang Liu. 2008.
\newblock \href {https://aclanthology.org/D08-1102/} {Learning to predict
  code-switching points}.
\newblock In \emph{Proceedings of the 2008 Conference on Empirical Methods in
  Natural Language Processing}, pages 973--981, Honolulu, Hawaii. Association
  for Computational Linguistics.

\bibitem[{Somani et~al.(2025)Somani, Kumar, and Malapati}]{11104253}
Aditya Somani, S.~R.~Mithun Kumar, and Aruna Malapati. 2025.
\newblock \href {https://doi.org/10.1109/ACCESS.2025.3594135} {Where does
  {mBERT} understand code-mixing? layer-dependent performance on semantic
  tasks}.
\newblock \emph{IEEE Access}, 13:135950--135956.

\bibitem[{Song et~al.(2025)Song, Huang, Zhou, and
  Ma}]{song-etal-2025-multilingual}
Jiayang Song, Yuheng Huang, Zhehua Zhou, and Lei Ma. 2025.
\newblock \href {https://doi.org/10.18653/v1/2025.findings-naacl.191}
  {Multilingual blending: Large language model safety alignment evaluation with
  language mixture}.
\newblock In \emph{Findings of the Association for Computational Linguistics:
  NAACL 2025}, pages 3433--3449, Albuquerque, New Mexico. Association for
  Computational Linguistics.

\bibitem[{Song et~al.(2019)Song, Zhang, Yu, Luo, Wang, and
  Zhang}]{song-etal-2019-code}
Kai Song, Yue Zhang, Heng Yu, Weihua Luo, Kun Wang, and Min Zhang. 2019.
\newblock \href {https://doi.org/10.18653/v1/N19-1044} {Code-switching for
  enhancing {NMT} with pre-specified translation}.
\newblock In \emph{Proceedings of the 2019 Conference of the North {A}merican
  Chapter of the Association for Computational Linguistics: Human Language
  Technologies, Volume 1 (Long and Short Papers)}, pages 449--459, Minneapolis,
  Minnesota. Association for Computational Linguistics.

\bibitem[{Song et~al.(2017)Song, Zou, Huang, Chen, and Liu}]{8300538}
Xiao Song, Yuexian Zou, Shilei Huang, Shaobin Chen, and Yi~Liu. 2017.
\newblock \href {https://doi.org/10.1109/IALP.2017.8300538} {Investigating
  multi-task learning for automatic speech recognition with code-switching
  between mandarin and english}.
\newblock In \emph{2017 International Conference on Asian Language Processing
  (IALP)}, pages 27--30.

\bibitem[{Song et~al.(2024)Song, Ma, Yang, Zhuo, and
  Chen}]{song2024comparative}
Zheshu Song, Ziyang Ma, Yifan Yang, Jianheng Zhuo, and Xie Chen. 2024.
\newblock A comparative study of llm-based asr and whisper in low resource and
  code switching scenario.
\newblock \emph{arXiv preprint arXiv:2412.00721}.

\bibitem[{Sravani and Mamidi(2023)}]{sravani-mamidi-2023-enhancing}
Dama Sravani and Radhika Mamidi. 2023.
\newblock \href {https://doi.org/10.18653/v1/2023.conll-1.15} {Enhancing
  code-mixed text generation using synthetic data filtering in neural machine
  translation}.
\newblock In \emph{Proceedings of the 27th Conference on Computational Natural
  Language Learning (CoNLL)}, pages 211--220, Singapore. Association for
  Computational Linguistics.

\bibitem[{Sreeram and Sinha(2017)}]{sreeram2017language}
Ganji Sreeram and Rohit Sinha. 2017.
\newblock Language modeling for code-switched data: Challenges and approaches.
\newblock \emph{arXiv preprint arXiv:1711.03541}.

\bibitem[{Srivastava and Sitaram(2018)}]{srivastava18_interspeech}
Brij Mohan~Lal Srivastava and Sunayana Sitaram. 2018.
\newblock \href {https://doi.org/10.21437/Interspeech.2018-1171} {{Homophone
  Identification and Merging for Code-switched Speech Recognition}}.
\newblock In \emph{{Interspeech 2018}}, pages 1943--1947.

\bibitem[{Srivastava et~al.(2026)Srivastava, Sharma, Mehrotra, Rajput, and
  Singh}]{11679710}
Garima Srivastava, Soniya Sharma, Tushar Mehrotra, Gaurav~Kr. Rajput, and
  Pradeep~Kumar Singh. 2026.
\newblock \href {https://doi.org/10.1109/GTSS68743.2026.11679710} {A
  robustness-centric evaluation framework for multilingual sentiment analysis
  under code-mixing}.
\newblock In \emph{2026 International Conference on Green Technologies and
  Sustainable Solutions (GTSS)}, pages 804--809.

\bibitem[{Srivastava and Singh(2020)}]{srivastava-singh-2020-phinc}
Vivek Srivastava and Mayank Singh. 2020.
\newblock \href {https://doi.org/10.18653/v1/2020.wnut-1.7} {{PHINC}: A
  parallel {H}inglish social media code-mixed corpus for machine translation}.
\newblock In \emph{Proceedings of the Sixth Workshop on Noisy User-generated
  Text (W-NUT 2020)}, pages 41--49, Online. Association for Computational
  Linguistics.

\bibitem[{Srivastava and Singh(2021)}]{srivastava-singh-2021-hinge}
Vivek Srivastava and Mayank Singh. 2021.
\newblock \href {https://doi.org/10.18653/v1/2021.eval4nlp-1.20} {{H}in{GE}: A
  dataset for generation and evaluation of code-mixed {H}inglish text}.
\newblock In \emph{Proceedings of the 2nd Workshop on Evaluation and Comparison
  of NLP Systems}, pages 200--208, Punta Cana, Dominican Republic. Association
  for Computational Linguistics.

\bibitem[{Srivastava and Singh(2022)}]{srivastava-singh-2022-overview}
Vivek Srivastava and Mayank Singh. 2022.
\newblock \href {https://doi.org/10.18653/v1/2022.wmt-1.74} {Overview and
  results of {M}ix{MT} shared-task at {WMT} 2022}.
\newblock In \emph{Proceedings of the Seventh Conference on Machine Translation
  (WMT)}, pages 806--811, Abu Dhabi, United Arab Emirates (Hybrid). Association
  for Computational Linguistics.

\bibitem[{Stephan and Jayakumar(2025)}]{10.3844/jcssp.2025.1176.1186}
Jose Stephan and Muthayyan~Kamalam Jayakumar. 2025.
\newblock \href {https://doi.org/10.3844/jcssp.2025.1176.1186} {Enhancing
  {Indian} language speech recognition systems with language-independent
  phonetic script: An experimental exploration}.
\newblock \emph{Journal of Computer Science}, 21(5):1176--1186.

\bibitem[{Sterner and Teufel(2025)}]{sterner-teufel-2025-minimal}
Igor Sterner and Simone Teufel. 2025.
\newblock \href {https://doi.org/10.18653/v1/2025.acl-long.910} {Minimal
  pair-based evaluation of code-switching}.
\newblock In \emph{Proceedings of the 63rd Annual Meeting of the Association
  for Computational Linguistics (Volume 1: Long Papers)}, pages 18575--18598,
  Vienna, Austria. Association for Computational Linguistics.

\bibitem[{Suhasini and Bharathi(2026)}]{suhasini2026parameter}
S~Suhasini and B~Bharathi. 2026.
\newblock Parameter-efficient {ASR} for {Tamil}-{English} code-mixed speech
  using {LoRA} and {QLoRA} techniques.
\newblock \emph{EURASIP Journal on Audio, Speech, and Music Processing},
  2026(1):37.

\bibitem[{Sullivan et~al.(2026)Sullivan, Talafha, Ashraf, Bougares, Elleuch,
  Zhang, Elmadany, Mohamed, Mdhaffar, Est{\`e}ve et~al.}]{sullivan2026nadi}
Peter Sullivan, Bashar Talafha, Ahmed Ashraf, Fethi Bougares, Haroun Elleuch,
  Chiyu Zhang, AbdelRahim Elmadany, Youssef Mohamed, Salima Mdhaffar, Yannick
  Est{\`e}ve, et~al. 2026.
\newblock {NADI} 2026: The second multidialectal {Arabic} speech processing
  shared task.
\newblock \emph{arXiv preprint arXiv:2609.27086}.

\bibitem[{Suresh et~al.(2025)Suresh, Surana, Hao, and
  Chng}]{suresh-etal-2025-cs}
Sathya~Krishnan Suresh, Tanmay Surana, Lim~Zhi Hao, and Eng~Siong Chng. 2025.
\newblock \href {https://doi.org/10.18653/v1/2025.newsum-main.3} {{CS}-sum: A
  benchmark for code-switching dialogue summarization and the limits of large
  language models}.
\newblock In \emph{Proceedings of The 5th New Frontiers in Summarization
  Workshop}, pages 31--47, Hybrid. Association for Computational Linguistics.

\bibitem[{Talafha et~al.(2021)Talafha, Abuammar, and Al-Ayyoub}]{IJECE22767}
Bashar Talafha, Analle Abuammar, and Mahmoud Al-Ayyoub. 2021.
\newblock \href {https://doi.org/10.11591/ijece.v11i3.pp2327-2334} {Atar:
  Attention-based {LSTM} for {Arabizi} transliteration}.
\newblock \emph{International Journal of Electrical and Computer Engineering
  (IJECE)}, 11(3):2327--2334.

\bibitem[{Tan and Joty(2021)}]{tan-joty-2021-code}
Samson Tan and Shafiq Joty. 2021.
\newblock \href {https://doi.org/10.18653/v1/2021.naacl-main.282} {Code-mixing
  on sesame street: Dawn of the adversarial polyglots}.
\newblock In \emph{Proceedings of the 2021 Conference of the North American
  Chapter of the Association for Computational Linguistics: Human Language
  Technologies}, pages 3596--3616, Online. Association for Computational
  Linguistics.

\bibitem[{Tang et~al.(2020)Tang, Tang, and Yuan}]{9220904}
Tiancheng Tang, Xinhuai Tang, and Tianyi Yuan. 2020.
\newblock \href {https://doi.org/10.1109/ACCESS.2020.3030468} {Fine-tuning
  {BERT} for multi-label sentiment analysis in unbalanced code-switching text}.
\newblock \emph{IEEE Access}, 8:193248--193256.

\bibitem[{Tareq et~al.(2023)Tareq, Islam, Deb, Rahman, and Mahmud}]{10129187}
Mohammad Tareq, Md.~Fokhrul Islam, Swakshar Deb, Sejuti Rahman, and Abdullah~Al
  Mahmud. 2023.
\newblock \href {https://doi.org/10.1109/ACCESS.2023.3277787}
  {Data-augmentation for {Bangla}-{English} code-mixed sentiment analysis:
  Enhancing cross linguistic contextual understanding}.
\newblock \emph{IEEE Access}, 11:51657--51671.

\bibitem[{Tarunesh et~al.(2021)Tarunesh, Kumar, and
  Jyothi}]{tarunesh-etal-2021-machine}
Ishan Tarunesh, Syamantak Kumar, and Preethi Jyothi. 2021.
\newblock \href {https://doi.org/10.18653/v1/2021.acl-long.245} {From machine
  translation to code-switching: Generating high-quality code-switched text}.
\newblock In \emph{Proceedings of the 59th Annual Meeting of the Association
  for Computational Linguistics and the 11th International Joint Conference on
  Natural Language Processing (Volume 1: Long Papers)}, pages 3154--3169,
  Online. Association for Computational Linguistics.

\bibitem[{Tatariya et~al.(2023)Tatariya, Lent, and
  de~Lhoneux}]{tatariya-etal-2023-transfer}
Kushal Tatariya, Heather Lent, and Miryam de~Lhoneux. 2023.
\newblock \href {https://doi.org/10.18653/v1/2023.wassa-1.32} {Transfer
  learning for code-mixed data: Do pretraining languages matter?}
\newblock In \emph{Proceedings of the 13th Workshop on Computational Approaches
  to Subjectivity, Sentiment, {\&} Social Media Analysis}, pages 365--378,
  Toronto, Canada. Association for Computational Linguistics.

\bibitem[{Terblanche et~al.(2024)Terblanche, Olaleye, and
  Marivate}]{terblanche-etal-2024-prompting}
Michelle Terblanche, Kayode Olaleye, and Vukosi Marivate. 2024.
\newblock \href {https://aclanthology.org/2024.sigul-1.33/} {Prompting towards
  alleviating code-switched data scarcity in under-resourced languages with
  {GPT} as a pivot}.
\newblock In \emph{Proceedings of the 3rd Annual Meeting of the Special
  Interest Group on Under-resourced Languages @ LREC-COLING 2024}, pages
  272--282, Torino, Italia. ELRA and ICCL.

\bibitem[{Tian et~al.(2022)Tian, Yu, Zhang, Weng, Zou, and Yu}]{tian2022lae}
Jinchuan Tian, Jianwei Yu, Chunlei Zhang, Chao Weng, Yuexian Zou, and Dong Yu.
  2022.
\newblock Lae: Language-aware encoder for monolingual and multilingual asr.
\newblock \emph{arXiv preprint arXiv:2206.02093}.

\bibitem[{Tsai et~al.(2019)Tsai, Riesa, Johnson, Arivazhagan, Li, and
  Archer}]{tsai-etal-2019-small}
Henry Tsai, Jason Riesa, Melvin Johnson, Naveen Arivazhagan, Xin Li, and Amelia
  Archer. 2019.
\newblock \href {https://doi.org/10.18653/v1/D19-1374} {Small and practical
  {BERT} models for sequence labeling}.
\newblock In \emph{Proceedings of the 2019 Conference on Empirical Methods in
  Natural Language Processing and the 9th International Joint Conference on
  Natural Language Processing (EMNLP-IJCNLP)}, pages 3632--3636, Hong Kong,
  China. Association for Computational Linguistics.

\bibitem[{Tseng et~al.(2024)Tseng, Chen, Chang, Lee, Huang, and Lee}]{10832290}
Liang-Hsuan Tseng, Zih-Ching Chen, Wei-Shun Chang, Cheng-Kuang Lee, Tsung-Ren
  Huang, and Hung-Yi Lee. 2024.
\newblock \href {https://doi.org/10.1109/SLT61566.2024.10832290} {Leave no
  knowledge behind during knowledge distillation: Towards practical and
  effective knowledge distillation for code-switching {ASR} using realistic
  data}.
\newblock In \emph{2024 IEEE Spoken Language Technology Workshop (SLT)}, pages
  118--125.

\bibitem[{Ugan et~al.(2025{\natexlab{a}})Ugan, Pham, Bärmann, and
  Waibel}]{10889660}
Enes~Yavuz Ugan, Ngoc-Quan Pham, Leonard Bärmann, and Alex Waibel.
  2025{\natexlab{a}}.
\newblock \href {https://doi.org/10.1109/ICASSP49660.2025.10889660} {{PIER}: A
  novel metric for evaluating what matters in code-switching}.
\newblock In \emph{ICASSP 2025 - 2025 IEEE International Conference on
  Acoustics, Speech and Signal Processing (ICASSP)}, pages 1--5.

\bibitem[{Ugan et~al.(2025{\natexlab{b}})Ugan, Pham, and
  Waibel}]{ugan2025adapting}
Enes~Yavuz Ugan, Ngoc-Quan Pham, and Alexander Waibel. 2025{\natexlab{b}}.
\newblock Adapting language balance in code-switching speech.
\newblock \emph{arXiv preprint arXiv:2510.18724}.

\bibitem[{Ullah and Ahmed(2020)}]{ullah2020code}
Asad Ullah and Tauseef Ahmed. 2020.
\newblock Code switching language model using monolingual training data.
\newblock \emph{arXiv preprint arXiv:2012.12543}.

\bibitem[{van~der Goot and
  {\c{C}}etino{\u{g}}lu(2021)}]{van-der-goot-cetinoglu-2021-lexical}
Rob van~der Goot and {\"O}zlem {\c{C}}etino{\u{g}}lu. 2021.
\newblock \href {https://doi.org/10.18653/v1/2021.eacl-main.200} {Lexical
  normalization for code-switched data and its effect on {POS} tagging}.
\newblock In \emph{Proceedings of the 16th Conference of the European Chapter
  of the Association for Computational Linguistics: Main Volume}, pages
  2352--2365, Online. Association for Computational Linguistics.

\bibitem[{van Vüren and Niesler(2023)}]{10022517}
Joshua~Jansen van Vüren and Thomas Niesler. 2023.
\newblock \href {https://doi.org/10.1109/SLT54892.2023.10022517} {Code-switched
  language modelling using a code predictive lstm in under-resourced {South}
  {African} languages}.
\newblock In \emph{2022 IEEE Spoken Language Technology Workshop (SLT)}, pages
  785--791.

\bibitem[{{van Vüren} and Niesler(2023)}]{vanvuren23_interspeech}
Joshua~Jansen {van Vüren} and Thomas Niesler. 2023.
\newblock \href {https://doi.org/10.21437/Interspeech.2023-1841} {{Improving
  Under-Resourced Code-Switched Speech Recognition: Large Pre-trained Models or
  Architectural Interventions}}.
\newblock In \emph{{Interspeech 2023}}, pages 1439--1443.

\bibitem[{Vijesh et~al.(2026)Vijesh, Gothai, and
  Student}]{Vijesh2026TanglishGuardBA}
Vivina Vijesh, Dr.~E. Gothai, and M.~Student. 2026.
\newblock \href {https://api.semanticscholar.org/CorpusID:290709412}
  {{TanglishGuard}: Benchmarking {AI} safety guardrails on {Tamil}-{English}
  code-mixed prompts}.
\newblock \emph{International Journal of Innovative Science and Research
  Technology}.

\bibitem[{Vilares et~al.(2015)Vilares, Alonso, and
  G{\'o}mez-Rodr{\'i}guez}]{vilares-etal-2015-sentiment}
David Vilares, Miguel~A. Alonso, and Carlos G{\'o}mez-Rodr{\'i}guez. 2015.
\newblock \href {https://doi.org/10.18653/v1/W15-2902} {Sentiment analysis on
  monolingual, multilingual and code-switching {T}witter corpora}.
\newblock In \emph{Proceedings of the 6th Workshop on Computational Approaches
  to Subjectivity, Sentiment and Social Media Analysis}, pages 2--8, Lisboa,
  Portugal. Association for Computational Linguistics.

\bibitem[{Vishnumolakala et~al.(2026)Vishnumolakala, Doddavarapu, Mangira, and
  Uchoi}]{11565004}
Mahesh~Babu Vishnumolakala, Pavan Venkat~Kumar Doddavarapu, Sasivardhan
  Mangira, and Enjula Uchoi. 2026.
\newblock \href {https://doi.org/10.1109/ICICI68867.2026.11565004} {Zero-shot
  and few-shot cross-linguistic transfer for token-level {Hinglish} language
  identification using multilingual {Transformers}}.
\newblock In \emph{2026 International Conference on Innovations in
  Computational Intelligence (ICICI)}, pages 1--6.

\bibitem[{Vishwakarma and Kumar(2026)}]{11542676}
Shashank Vishwakarma and Rakesh Kumar. 2026.
\newblock \href {https://doi.org/10.1109/ICISESSC68634.2026.11542676}
  {{LoRA}-based benchmarking of multilingual {BERT} models for
  {Hindi}-{English} code-mixed text}.
\newblock In \emph{2026 International Conference on Intelligent Systems in
  Engineering, Secured Systems and Cybersecurity (ICISESSC)}, pages 998--1001.

\bibitem[{Visser et~al.(2025)Visser, Grobler, and Dunaiski}]{visser2025code}
Ruan Visser, Trienko Grobler, and Marcel Dunaiski. 2025.
\newblock Code-switch pretraining for improved cross-lingual alignment in
  low-resource languages.
\newblock In \emph{Southern African Conference for Artificial Intelligence
  Research}, pages 239--253. Springer.

\bibitem[{Vu et~al.(2012)Vu, Lyu, Weiner, Telaar, Schlippe, Blaicher, Chng,
  Schultz, and Li}]{6289015}
Ngoc~Thang Vu, Dau-Cheng Lyu, Jochen Weiner, Dominic Telaar, Tim Schlippe,
  Fabian Blaicher, Eng-Siong Chng, Tanja Schultz, and Haizhou Li. 2012.
\newblock \href {https://doi.org/10.1109/ICASSP.2012.6289015} {A first speech
  recognition system for {Mandarin}-{English} code-switch conversational
  speech}.
\newblock In \emph{2012 IEEE International Conference on Acoustics, Speech and
  Signal Processing (ICASSP)}, pages 4889--4892.

\bibitem[{Wang et~al.(2023{\natexlab{a}})Wang, Huang, Chang, and
  Chen}]{wang-etal-2023-self-augmentation}
Fei Wang, Kuan-Hao Huang, Kai-Wei Chang, and Muhao Chen. 2023{\natexlab{a}}.
\newblock \href {https://doi.org/10.18653/v1/2023.ijcnlp-short.1}
  {Self-augmentation improves zero-shot cross-lingual transfer}.
\newblock In \emph{Proceedings of the 13th International Joint Conference on
  Natural Language Processing and the 3rd Conference of the Asia-Pacific
  Chapter of the Association for Computational Linguistics (Volume 2: Short
  Papers)}, pages 1--9, Nusa Dua, Bali. Association for Computational
  Linguistics.

\bibitem[{Wang et~al.(2025{\natexlab{a}})Wang, Wan, Zheng, Liu, Zhou, Li, and
  Xie}]{10890869}
He~Wang, Xucheng Wan, Naijun Zheng, Kai Liu, Huan Zhou, Guojian Li, and Lei
  Xie. 2025{\natexlab{a}}.
\newblock \href {https://doi.org/10.1109/ICASSP49660.2025.10890869} {{CAMEL}:
  Cross-attention enhanced mixture-of-experts and language bias for
  code-switching speech recognition}.
\newblock In \emph{ICASSP 2025 - 2025 IEEE International Conference on
  Acoustics, Speech and Signal Processing (ICASSP)}, pages 1--5.

\bibitem[{Wang et~al.(2023{\natexlab{b}})Wang, Ma, Li, and
  Du}]{wang2023languageroutingmixtureexpertsmultilingual}
Wenxuan Wang, Guodong Ma, Yuke Li, and Binbin Du. 2023{\natexlab{b}}.
\newblock \href {https://arxiv.org/abs/2307.05956} {Language-routing mixture of
  experts for multilingual and code-switching speech recognition}.
\newblock \emph{Preprint}, arXiv:2307.05956.

\bibitem[{Wang et~al.(2025{\natexlab{b}})Wang, Li, Zhou, Weng, Wang, Huang,
  Han, Feng, Deng, and Huang}]{wang-etal-2025-investigating-scaling}
Zhijun Wang, Jiahuan Li, Hao Zhou, Rongxiang Weng, Jingang Wang, Xin Huang, Xue
  Han, Junlan Feng, Chao Deng, and Shujian Huang. 2025{\natexlab{b}}.
\newblock \href {https://doi.org/10.18653/v1/2025.findings-acl.575}
  {Investigating and scaling up code-switching for multilingual language model
  pre-training}.
\newblock In \emph{Findings of the Association for Computational Linguistics:
  ACL 2025}, pages 11032--11046, Vienna, Austria. Association for Computational
  Linguistics.

\bibitem[{Wang et~al.(2026)Wang, Liu, Zhou, Wei, Yang, and
  Huang}]{wang2026efficient}
Zhijun Wang, Junxiao Liu, Hao Zhou, Hao-Ran Wei, Baosong Yang, and Shujian
  Huang. 2026.
\newblock Efficient multilingual reasoning transfer via progressive
  code-switching.
\newblock \emph{arXiv preprint arXiv:2607.00485}.

\bibitem[{Weller et~al.(2022)Weller, Sperber, Pires, Setiawan, Gollan, Telaar,
  and Paulik}]{weller-etal-2022-end}
Orion Weller, Matthias Sperber, Telmo Pires, Hendra Setiawan, Christian Gollan,
  Dominic Telaar, and Matthias Paulik. 2022.
\newblock \href {https://doi.org/10.18653/v1/2022.findings-acl.113} {End-to-end
  speech translation for code switched speech}.
\newblock In \emph{Findings of the Association for Computational Linguistics:
  ACL 2022}, pages 1435--1448, Dublin, Ireland. Association for Computational
  Linguistics.

\bibitem[{White et~al.(2008)White, Khudanpur, and Baker}]{white08_interspeech}
Christopher~M. White, Sanjeev Khudanpur, and James~K. Baker. 2008.
\newblock \href {https://doi.org/10.21437/Interspeech.2008-667} {{An
  investigation of acoustic models for multilingual code-switching}}.
\newblock In \emph{{Interspeech 2008}}, pages 2691--2694.

\bibitem[{Whitehouse et~al.(2022)Whitehouse, Christopoulou, and
  Iacobacci}]{whitehouse-etal-2022-entitycs}
Chenxi Whitehouse, Fenia Christopoulou, and Ignacio Iacobacci. 2022.
\newblock \href {https://doi.org/10.18653/v1/2022.findings-emnlp.499}
  {{E}ntity{CS}: Improving zero-shot cross-lingual transfer with entity-centric
  code switching}.
\newblock In \emph{Findings of the Association for Computational Linguistics:
  EMNLP 2022}, pages 6698--6714, Abu Dhabi, United Arab Emirates. Association
  for Computational Linguistics.

\bibitem[{Wick et~al.(2016)Wick, Kanani, and Pocock}]{3016100.3016300}
Michael Wick, Pallika Kanani, and Adam Pocock. 2016.
\newblock Minimally-constrained multilingual embeddings via artificial
  code-switching.
\newblock In \emph{Proceedings of the Thirtieth AAAI Conference on Artificial
  Intelligence}, AAAI'16, page 2849–2855. AAAI Press.

\bibitem[{Wiesner et~al.(2021)Wiesner, Sarma, Arora, Raj, Gao, Huang, Preet,
  Johnson, Iqbal, Goel, Trmal, Perera, and Khudanpur}]{wiesner21_interspeech}
Matthew Wiesner, Mousmita Sarma, Ashish Arora, Desh Raj, Dongji Gao, Ruizhe
  Huang, Supreet Preet, Moris Johnson, Zikra Iqbal, Nagendra Goel, Jan Trmal,
  Leibny Paola~García Perera, and Sanjeev Khudanpur. 2021.
\newblock \href {https://doi.org/10.21437/Interspeech.2021-2127} {{Training
  Hybrid Models on Noisy Transliterated Transcripts for Code-Switched Speech
  Recognition}}.
\newblock In \emph{{Interspeech 2021}}, pages 2906--2910.

\bibitem[{Winata et~al.(2023)Winata, Aji, Yong, and
  Solorio}]{winata-etal-2023-decades}
Genta Winata, Alham~Fikri Aji, Zheng~Xin Yong, and Thamar Solorio. 2023.
\newblock \href {https://doi.org/10.18653/v1/2023.findings-acl.185} {The
  decades progress on code-switching research in {NLP}: A systematic survey on
  trends and challenges}.
\newblock In \emph{Findings of the Association for Computational Linguistics:
  ACL 2023}, pages 2936--2978, Toronto, Canada. Association for Computational
  Linguistics.

\bibitem[{Winata et~al.(2020)Winata, Cahyawijaya, Lin, Liu, Xu, and
  Fung}]{winata-etal-2020-meta}
Genta~Indra Winata, Samuel Cahyawijaya, Zhaojiang Lin, Zihan Liu, Peng Xu, and
  Pascale Fung. 2020.
\newblock \href {https://doi.org/10.18653/v1/2020.acl-main.348} {Meta-transfer
  learning for code-switched speech recognition}.
\newblock In \emph{Proceedings of the 58th Annual Meeting of the Association
  for Computational Linguistics}, pages 3770--3776, Online. Association for
  Computational Linguistics.

\bibitem[{Winata et~al.(2021)Winata, Cahyawijaya, Liu, Lin, Madotto, and
  Fung}]{winata-etal-2021-multilingual}
Genta~Indra Winata, Samuel Cahyawijaya, Zihan Liu, Zhaojiang Lin, Andrea
  Madotto, and Pascale Fung. 2021.
\newblock \href {https://doi.org/10.18653/v1/2021.calcs-1.20} {Are multilingual
  models effective in code-switching?}
\newblock In \emph{Proceedings of the Fifth Workshop on Computational
  Approaches to Linguistic Code-Switching}, pages 142--153, Online. Association
  for Computational Linguistics.

\bibitem[{Winata et~al.(2019{\natexlab{a}})Winata, Lin, and
  Fung}]{winata-etal-2019-learning}
Genta~Indra Winata, Zhaojiang Lin, and Pascale Fung. 2019{\natexlab{a}}.
\newblock \href {https://doi.org/10.18653/v1/W19-4320} {Learning multilingual
  meta-embeddings for code-switching named entity recognition}.
\newblock In \emph{Proceedings of the 4th Workshop on Representation Learning
  for NLP (RepL4NLP-2019)}, pages 181--186, Florence, Italy. Association for
  Computational Linguistics.

\bibitem[{Winata et~al.(2019{\natexlab{b}})Winata, Lin, Shin, Liu, and
  Fung}]{winata-etal-2019-hierarchical}
Genta~Indra Winata, Zhaojiang Lin, Jamin Shin, Zihan Liu, and Pascale Fung.
  2019{\natexlab{b}}.
\newblock \href {https://doi.org/10.18653/v1/D19-1360} {Hierarchical
  meta-embeddings for code-switching named entity recognition}.
\newblock In \emph{Proceedings of the 2019 Conference on Empirical Methods in
  Natural Language Processing and the 9th International Joint Conference on
  Natural Language Processing (EMNLP-IJCNLP)}, pages 3541--3547, Hong Kong,
  China. Association for Computational Linguistics.

\bibitem[{Winata et~al.(2018{\natexlab{a}})Winata, Madotto, Wu, and
  Fung}]{winata-etal-2018-code}
Genta~Indra Winata, Andrea Madotto, Chien-Sheng Wu, and Pascale Fung.
  2018{\natexlab{a}}.
\newblock \href {https://doi.org/10.18653/v1/W18-3207} {Code-switching language
  modeling using syntax-aware multi-task learning}.
\newblock In \emph{Proceedings of the Third Workshop on Computational
  Approaches to Linguistic Code-Switching}, pages 62--67, Melbourne, Australia.
  Association for Computational Linguistics.

\bibitem[{Winata et~al.(2018{\natexlab{b}})Winata, Madotto, Wu, and
  Fung}]{winata2018learn}
Genta~Indra Winata, Andrea Madotto, Chien-Sheng Wu, and Pascale Fung.
  2018{\natexlab{b}}.
\newblock Learn to code-switch: Data augmentation using copy mechanism on
  language modeling.
\newblock \emph{arXiv preprint arXiv:1810.10254}.

\bibitem[{Winata et~al.(2019{\natexlab{c}})Winata, Madotto, Wu, and
  Fung}]{winata-etal-2019-code}
Genta~Indra Winata, Andrea Madotto, Chien-Sheng Wu, and Pascale Fung.
  2019{\natexlab{c}}.
\newblock \href {https://doi.org/10.18653/v1/K19-1026} {Code-switched language
  models using neural based synthetic data from parallel sentences}.
\newblock In \emph{Proceedings of the 23rd Conference on Computational Natural
  Language Learning (CoNLL)}, pages 271--280, Hong Kong, China. Association for
  Computational Linguistics.

\bibitem[{Wu et~al.(2012)Wu, Shen, and Yang}]{6289009}
Chung-Hsien Wu, Han-Ping Shen, and Yan-Ting Yang. 2012.
\newblock \href {https://doi.org/10.1109/ICASSP.2012.6289009} {Phone set
  construction based on context-sensitive articulatory attributes for
  code-switching speech recognition}.
\newblock In \emph{2012 IEEE International Conference on Acoustics, Speech and
  Signal Processing (ICASSP)}, pages 4865--4868.

\bibitem[{Wu et~al.(2014)Wu, Shen, and Yang}]{6763085}
Chung-Hsien Wu, Han-Ping Shen, and Yan-Ting Yang. 2014.
\newblock \href {https://doi.org/10.1109/TASLP.2014.2310353}
  {{Chinese}-{English} phone set construction for code-switching {ASR} using
  acoustic and {DNN}-extracted articulatory features}.
\newblock \emph{IEEE/ACM Transactions on Audio, Speech, and Language
  Processing}, 22(4):858--862.

\bibitem[{Xiang et~al.(2023)Xiang, Zhang, Di, Huang, Li, Ouchi, Chen, and
  Xu}]{yu-etal-2023-improving}
Yu~Xiang, Ting Zhang, Hui Di, Hui Huang, Chunyou Li, Kazushige Ouchi, Yufeng
  Chen, and Jinan Xu. 2023.
\newblock \href {https://aclanthology.org/2023.ccl-1.54/} {Improving zero-shot
  cross-lingual dialogue state tracking via contrastive learning}.
\newblock In \emph{Proceedings of the 22nd Chinese National Conference on
  Computational Linguistics}, pages 624--625, Harbin, China. Chinese
  Information Processing Society of China.

\bibitem[{Xie et~al.(2025)Xie, Babbar, Chen, and Turura}]{xie2025enhancing}
Katherine Xie, Nitya Babbar, Vicky Chen, and Yoanna Turura. 2025.
\newblock Enhancing multilingual language models for code-switched input data.
\newblock \emph{arXiv preprint arXiv:2503.07990}.

\bibitem[{Xing and Tsang(2024)}]{10534816}
Bowen Xing and Ivor~W. Tsang. 2024.
\newblock \href {https://doi.org/10.1109/TPAMI.2024.3402746} {{HC}$^{2}$l:
  Hybrid and cooperative contrastive learning for cross-lingual spoken language
  understanding}.
\newblock \emph{IEEE Transactions on Pattern Analysis and Machine
  Intelligence}, 46(12):8094--8105.

\bibitem[{Xu et~al.(2025)Xu, Tan, Wang, and Chen}]{11434663}
Jing Xu, Daxin Tan, Jiaqi Wang, and Xiao Chen. 2025.
\newblock \href {https://doi.org/10.1109/ASRU65441.2025.11434663} {Enhancing
  code-switched text-to-speech synthesis capability in large language models
  with only monolingual corpora}.
\newblock In \emph{2025 IEEE Automatic Speech Recognition and Understanding
  Workshop (ASRU)}, pages 1--8.

\bibitem[{Xue et~al.(2019)Xue, Song, Xu, Xie, and Wu}]{xue19_interspeech}
Liumeng Xue, Wei Song, Guanghui Xu, Lei Xie, and Zhizheng Wu. 2019.
\newblock \href {https://doi.org/10.21437/Interspeech.2019-3191} {{Building a
  Mixed-Lingual Neural TTS System with Only Monolingual Data}}.
\newblock In \emph{{Interspeech 2019}}, pages 2060--2064.

\bibitem[{Yadav and Chakraborty(2020)}]{yadav2020unsupervised}
Siddharth Yadav and Tanmoy Chakraborty. 2020.
\newblock Unsupervised sentiment analysis for code-mixed data.
\newblock \emph{arXiv preprint arXiv:2001.11384}.

\bibitem[{Yang et~al.(2024{\natexlab{a}})Yang, Huang, Lu, Kuan, Hsiao, and
  Lee}]{10626762}
Chih-Kai Yang, Kuan-Po Huang, Ke-Han Lu, Chun-Yi Kuan, Chi-Yuan Hsiao, and
  Hung-Yi Lee. 2024{\natexlab{a}}.
\newblock \href {https://doi.org/10.1109/ICASSPW62465.2024.10626762}
  {Investigating zero-shot generalizability on {Mandarin}-{English}
  code-switched {ASR} and speech-to-text translation of recent foundation
  models with self-supervision and weak supervision}.
\newblock In \emph{2024 IEEE International Conference on Acoustics, Speech, and
  Signal Processing Workshops (ICASSPW)}, pages 540--544.

\bibitem[{Yang et~al.(2025)Yang, Peng, Huang, and Li}]{yang25p_interspeech}
Hongli Yang, Yizhou Peng, Hao Huang, and Sheng Li. 2025.
\newblock \href {https://doi.org/10.21437/Interspeech.2025-2549} {{Adapting
  Whisper for Parameter-efficient Code-Switching Speech Recognition via Soft
  Prompt Tuning}}.
\newblock In \emph{{Interspeech 2025}}, pages 5203--5207.

\bibitem[{Yang et~al.(2020{\natexlab{a}})Yang, Ma, Zhang, Wu, Li, and
  Zhou}]{yang2020alternating}
Jian Yang, Shuming Ma, Dongdong Zhang, Shuangzhi Wu, Zhoujun Li, and Ming Zhou.
  2020{\natexlab{a}}.
\newblock Alternating language modeling for cross-lingual pre-training.
\newblock In \emph{Proceedings of the AAAI Conference on Artificial
  Intelligence}, volume~34, pages 9386--9393.

\bibitem[{Yang and Chai(2025)}]{yang-chai-2025-codemixbench}
Yilun Yang and Yekun Chai. 2025.
\newblock \href {https://doi.org/10.18653/v1/2025.emnlp-main.109}
  {{C}ode{M}ix{B}ench: Evaluating code-mixing capabilities of {LLM}s across 18
  languages}.
\newblock In \emph{Proceedings of the 2025 Conference on Empirical Methods in
  Natural Language Processing}, pages 2139--2169, Suzhou, China. Association
  for Computational Linguistics.

\bibitem[{Yang et~al.(2024{\natexlab{b}})Yang, Peng, Huang, Chng, and
  Zhong}]{10849308}
Yuhang Yang, Yizhou Peng, Hao Huang, Eng~Siong Chng, and Xionghu Zhong.
  2024{\natexlab{b}}.
\newblock \href {https://doi.org/10.1109/APSIPAASC63619.2025.10849308}
  {Adapting {OpenAI’s} {Whisper} for speech recognition on code-switch
  {Mandarin}-{English} {SEAME} and {ASRU2019} datasets}.
\newblock In \emph{2024 Asia Pacific Signal and Information Processing
  Association Annual Summit and Conference (APSIPA ASC)}, pages 1--6.

\bibitem[{Yang et~al.(2020{\natexlab{b}})Yang, Hu, Han, Huang, and
  Ju}]{yang-etal-2020-csp}
Zhen Yang, Bojie Hu, Ambyera Han, Shen Huang, and Qi~Ju. 2020{\natexlab{b}}.
\newblock \href {https://doi.org/10.18653/v1/2020.emnlp-main.208}
  {{CSP}:code-switching pre-training for neural machine translation}.
\newblock In \emph{Proceedings of the 2020 Conference on Empirical Methods in
  Natural Language Processing (EMNLP)}, pages 2624--2636, Online. Association
  for Computational Linguistics.

\bibitem[{Ye et~al.(2024)Ye, Chen, Hu, and Xu}]{ye2024sc}
Shuaishuai Ye, Shunfei Chen, Xinhui Hu, and Xinkang Xu. 2024.
\newblock {SC}-{MoE}: Switch conformer mixture of experts for unified streaming
  and non-streaming code-switching {ASR}.
\newblock \emph{arXiv preprint arXiv:2406.18021}.

\bibitem[{Yeh and Lee(2014)}]{6853590}
Ching-Feng Yeh and Lin-Shan Lee. 2014.
\newblock \href {https://doi.org/10.1109/ICASSP.2014.6853590} {Transcribing
  code-switched bilingual lectures using deep neural networks with unit merging
  in acoustic modeling}.
\newblock In \emph{2014 IEEE International Conference on Acoustics, Speech and
  Signal Processing (ICASSP)}, pages 220--224.

\bibitem[{Yeh et~al.(2012)Yeh, Lin, and Lee}]{6423531}
Ching-Feng Yeh, Yiu-Chang Lin, and Lin-Shan Lee. 2012.
\newblock \href {https://doi.org/10.1109/ISCSLP.2012.6423531} {Minimum phone
  error model training on merged acoustic units for transcribing bilingual
  code-switched speech}.
\newblock In \emph{2012 8th International Symposium on Chinese Spoken Language
  Processing}, pages 320--324.

\bibitem[{Yeo et~al.(2025)Yeo, Hu, Gopal, Peng, Liu, and Chng}]{11249050}
Yue~Heng Yeo, Yuchen Hu, Shreyas Gopal, Yizhou Peng, Hexin Liu, and Eng~Siong
  Chng. 2025.
\newblock \href {https://doi.org/10.1109/APSIPAASC65261.2025.11249050}
  {Improving code-switching speech recognition with {TTS} data augmentation}.
\newblock In \emph{2025 Asia Pacific Signal and Information Processing
  Association Annual Summit and Conference (APSIPA ASC)}, pages 992--997.

\bibitem[{Yeo et~al.(2026)Yeo, Li, Peng, Gopal, Liu, Garcia-Perera, Sailor,
  Wong, and Chng}]{yeo2026improving}
Yue~Heng Yeo, Haoyang Li, Yizhou Peng, Shreyas Gopal, Hexin Liu, Leibny~Paola
  Garcia-Perera, Hardik~B Sailor, Jeremy~HM Wong, and Eng~Siong Chng. 2026.
\newblock Improving code-switching {ASR} with code-mixing guided synthetic
  speech.
\newblock \emph{arXiv preprint arXiv:2606.19381}.

\bibitem[{Yong et~al.(2023)Yong, Zhang, Forde, Wang, Subramonian, Lovenia,
  Cahyawijaya, Winata, Sutawika, Cruz, Tan, Phan, Garcia, Solorio, and
  Aji}]{yong-etal-2023-prompting}
Zheng-Xin Yong, Ruochen Zhang, Jessica~Zosa Forde, Skyler Wang, Arjun
  Subramonian, Holy Lovenia, Samuel Cahyawijaya, Genta~Indra Winata, Lintang
  Sutawika, Jan Christian~Blaise Cruz, Yin~Lin Tan, Long Phan, Rowena Garcia,
  Thamar Solorio, and Alham~Fikri Aji. 2023.
\newblock \href {https://aclanthology.org/2023.calcs-1.5/} {Prompting
  multilingual large language models to generate code-mixed texts: The case of
  {South} {E}ast {A}sian languages}.
\newblock In \emph{Proceedings of the 6th Workshop on Computational Approaches
  to Linguistic Code-Switching}, pages 43--63, Singapore. Association for
  Computational Linguistics.

\bibitem[{Yoo et~al.(2026)Yoo, Jin, Cho, and Oh}]{yoo2026gradual}
Haneul Yoo, Jiho Jin, Kyunghyun Cho, and Alice Oh. 2026.
\newblock Gradual code-switching as inference-time cross-lingual
  representational alignment for {LLMs}.
\newblock In \emph{Third Conference on Language Modeling}.

\bibitem[{Yoo et~al.(2025{\natexlab{a}})Yoo, Park, Yun, Oh, and
  Lee}]{yoo-etal-2025-code-switching}
Haneul Yoo, Cheonbok Park, Sangdoo Yun, Alice Oh, and Hwaran Lee.
  2025{\natexlab{a}}.
\newblock \href {https://doi.org/10.18653/v1/2025.findings-acl.407}
  {Code-switching curriculum learning for multilingual transfer in {LLM}s}.
\newblock In \emph{Findings of the Association for Computational Linguistics:
  ACL 2025}, pages 7816--7836, Vienna, Austria. Association for Computational
  Linguistics.

\bibitem[{Yoo et~al.(2025{\natexlab{b}})Yoo, Yang, and
  Lee}]{yoo-etal-2025-code}
Haneul Yoo, Yongjin Yang, and Hwaran Lee. 2025{\natexlab{b}}.
\newblock \href {https://doi.org/10.18653/v1/2025.acl-long.657} {Code-switching
  red-teaming: {LLM} evaluation for safety and multilingual understanding}.
\newblock In \emph{Proceedings of the 63rd Annual Meeting of the Association
  for Computational Linguistics (Volume 1: Long Papers)}, pages 13392--13413,
  Vienna, Austria. Association for Computational Linguistics.

\bibitem[{Yu et~al.(2023)Yu, Hu, Qian, Jin, Liu, Liu, Shi, Qian, Lin, and
  Zeng}]{10096317}
Haibin Yu, Yuxuan Hu, Yao Qian, Ma~Jin, Linquan Liu, Shujie Liu, Yu~Shi, Yanmin
  Qian, Edward Lin, and Michael Zeng. 2023.
\newblock \href {https://doi.org/10.1109/ICASSP49357.2023.10096317}
  {Code-switching text generation and injection in {Mandarin}-{English} {ASR}}.
\newblock In \emph{ICASSP 2023 - 2023 IEEE International Conference on
  Acoustics, Speech and Signal Processing (ICASSP)}, pages 1--5.

\bibitem[{Yue et~al.(2018)Yue, Chen, Zhang, Lin, and
  Yang}]{10.1007/978-3-319-99501-4_37}
Tianchi Yue, Chen Chen, Shaowu Zhang, Hongfei Lin, and Liang Yang. 2018.
\newblock Ensemble of neural networks with sentiment words translation for
  code-switching emotion detection.
\newblock In \emph{Natural Language Processing and Chinese Computing}, pages
  411--419, Cham. Springer International Publishing.

\bibitem[{Yılmaz et~al.(2018)Yılmaz, {van den Heuvel}, and {van
  Leeuwen}}]{ylmaz18_interspeech}
Emre Yılmaz, Henk {van den Heuvel}, and David {van Leeuwen}. 2018.
\newblock \href {https://doi.org/10.21437/Interspeech.2018-52} {{Acoustic and
  Textual Data Augmentation for Improved ASR of Code-Switching Speech}}.
\newblock In \emph{{Interspeech 2018}}, pages 1933--1937.

\bibitem[{Zan et~al.(2022)Zan, Ding, Shen, Cao, Liu, and Tao}]{zan2022bridging}
Changtong Zan, Liang Ding, Li~Shen, Yu~Cao, Weifeng Liu, and Dacheng Tao. 2022.
\newblock Bridging cross-lingual gaps during leveraging the multilingual
  sequence-to-sequence pretraining for text generation and understanding.
\newblock \emph{arXiv preprint arXiv:2204.07834}.

\bibitem[{Zeng et~al.(2025)Zeng, Liao, Zhou, Huang, Lin, and Yang}]{11033355}
Meiyu Zeng, Xingming Liao, Yongmei Zhou, Wentao Huang, Nankai Lin, and Aimin
  Yang. 2025.
\newblock \href {https://doi.org/10.1109/CSCWD64889.2025.11033355} {Enhancing
  cross-lingual aspect-based sentiment analysis with code-mixed in-context
  demonstrations and language-specific tags}.
\newblock In \emph{2025 28th International Conference on Computer Supported
  Cooperative Work in Design (CSCWD)}, pages 716--721.

\bibitem[{Zeng et~al.(2026)Zeng, Lu, Zhou, Qi, Yu, Zhao, Yanaka, Xuan, and
  Yokoya}]{zeng-etal-2026-code}
Qingcheng Zeng, Yuheng Lu, Zeqi Zhou, Heli Qi, Puxuan Yu, Fuheng Zhao, Hitomi
  Yanaka, Weihao Xuan, and Naoto Yokoya. 2026.
\newblock \href {https://doi.org/10.18653/v1/2026.findings-acl.636}
  {Code-switching information retrieval: Benchmarks, analysis, and the limits
  of current retrievers}.
\newblock In \emph{Findings of the {A}ssociation for {C}omputational
  {L}inguistics: {ACL} 2026}, pages 13055--13071, San Diego, California, United
  States. Association for Computational Linguistics.

\bibitem[{Zeng et~al.(2019)Zeng, Khassanov, Pham, Xu, Chng, and
  Li}]{zeng19_interspeech}
Zhiping Zeng, Yerbolat Khassanov, Van~Tung Pham, Haihua Xu, Eng~Siong Chng, and
  Haizhou Li. 2019.
\newblock \href {https://doi.org/10.21437/Interspeech.2019-1429} {{On the
  End-to-End Solution to Mandarin-English Code-Switching Speech Recognition}}.
\newblock In \emph{{Interspeech 2019}}, pages 2165--2169.

\bibitem[{Zhang et~al.(2025)Zhang, Geng, Huang, Shan, Yi, and Qu}]{10890030}
Fengrun Zhang, Wang Geng, Hukai Huang, Yahui Shan, Cheng Yi, and He~Qu. 2025.
\newblock \href {https://doi.org/10.1109/ICASSP49660.2025.10890030} {Boosting
  code-switching {ASR} with mixture of experts enhanced speech-conditioned
  {LLM}}.
\newblock In \emph{ICASSP 2025 - 2025 IEEE International Conference on
  Acoustics, Speech and Signal Processing (ICASSP)}, pages 1--5.

\bibitem[{Zhang and Lin(2021)}]{zhang2021improve}
Haitong Zhang and Yue Lin. 2021.
\newblock Improve cross-lingual voice cloning using low-quality code-switched
  data.
\newblock \emph{arXiv preprint arXiv:2110.07210}.

\bibitem[{Zhang et~al.(2020)Zhang, Xu, Pham, Huang, and
  Chng}]{zhang20r_interspeech}
Haobo Zhang, Haihua Xu, Van~Tung Pham, Hao Huang, and Eng~Siong Chng. 2020.
\newblock \href {https://doi.org/10.21437/Interspeech.2020-1582} {{Monolingual
  Data Selection Analysis for English-Mandarin Hybrid Code-Switching Speech
  Recognition}}.
\newblock In \emph{{Interspeech 2020}}, pages 2392--2396.

\bibitem[{Zhang et~al.(2023)Zhang, Cahyawijaya, Cruz, Winata, and
  Aji}]{zhang-etal-2023-multilingual}
Ruochen Zhang, Samuel Cahyawijaya, Jan Christian~Blaise Cruz, Genta Winata, and
  Alham~Fikri Aji. 2023.
\newblock \href {https://doi.org/10.18653/v1/2023.emnlp-main.774} {Multilingual
  large language models are not (yet) code-switchers}.
\newblock In \emph{Proceedings of the 2023 Conference on Empirical Methods in
  Natural Language Processing}, pages 12567--12582, Singapore. Association for
  Computational Linguistics.

\bibitem[{Zhang et~al.(2021{\natexlab{a}})Zhang, Yi, Tian, Bai, Tao, and
  wen}]{9414428}
Shuai Zhang, Jiangyan Yi, Zhengkun Tian, Ye~Bai, Jianhua Tao, and Zhengqi wen.
  2021{\natexlab{a}}.
\newblock \href {https://doi.org/10.1109/ICASSP39728.2021.9414428} {Decoupling
  pronunciation and language for end-to-end code-switching automatic speech
  recognition}.
\newblock In \emph{ICASSP 2021 - 2021 IEEE International Conference on
  Acoustics, Speech and Signal Processing (ICASSP)}, pages 6249--6253.

\bibitem[{Zhang et~al.(2024)Zhang, Majumdar, Ekbal, and
  Yadav}]{Zhang2024CHAIFL}
Wenbo Zhang, Aditya Majumdar, Asif Ekbal, and Amulya Yadav. 2024.
\newblock {CHAI} for {LLM}s: Improving code-mixed translation in large language
  models through reinforcement learning with {AI} feedback.
\newblock \emph{arXiv preprint arXiv:2411.09073}.

\bibitem[{Zhang et~al.(2021{\natexlab{b}})Zhang, He, Peng, Bing, and
  Lam}]{zhang-etal-2021-cross}
Wenxuan Zhang, Ruidan He, Haiyun Peng, Lidong Bing, and Wai Lam.
  2021{\natexlab{b}}.
\newblock \href {https://doi.org/10.18653/v1/2021.emnlp-main.727}
  {Cross-lingual aspect-based sentiment analysis with aspect term
  code-switching}.
\newblock In \emph{Proceedings of the 2021 Conference on Empirical Methods in
  Natural Language Processing}, pages 9220--9230, Online and Punta Cana,
  Dominican Republic. Association for Computational Linguistics.

\bibitem[{Zhang et~al.(2018)Zhang, Riesa, Gillick, Bakalov, Baldridge, and
  Weiss}]{zhang-etal-2018-fast-compact}
Yuan Zhang, Jason Riesa, Daniel Gillick, Anton Bakalov, Jason Baldridge, and
  David Weiss. 2018.
\newblock \href {https://doi.org/10.18653/v1/D18-1030} {A fast, compact,
  accurate model for language identification of codemixed text}.
\newblock In \emph{Proceedings of the 2018 Conference on Empirical Methods in
  Natural Language Processing}, pages 328--337, Brussels, Belgium. Association
  for Computational Linguistics.

\bibitem[{Zhao et~al.(2025)Zhao, Shi, Cui, Wang, Liu, Ni, Ye, and
  Wang}]{10889634}
Jiahui Zhao, Hao Shi, Chenrui Cui, Tianrui Wang, Hexin Liu, Zhaoheng Ni,
  Lingxuan Ye, and Longbiao Wang. 2025.
\newblock \href {https://doi.org/10.1109/ICASSP49660.2025.10889634} {Adapting
  {Whisper} for code-switching through encoding refining and language-aware
  decoding}.
\newblock In \emph{ICASSP 2025 - 2025 IEEE International Conference on
  Acoustics, Speech and Signal Processing (ICASSP)}, pages 1--5.

\bibitem[{Zheng et~al.(2021)Zheng, Dong, Huang, Wang, Chi, Singhal, Che, Liu,
  Song, and Wei}]{zheng-etal-2021-consistency}
Bo~Zheng, Li~Dong, Shaohan Huang, Wenhui Wang, Zewen Chi, Saksham Singhal,
  Wanxiang Che, Ting Liu, Xia Song, and Furu Wei. 2021.
\newblock \href {https://doi.org/10.18653/v1/2021.acl-long.264} {Consistency
  regularization for cross-lingual fine-tuning}.
\newblock In \emph{Proceedings of the 59th Annual Meeting of the Association
  for Computational Linguistics and the 11th International Joint Conference on
  Natural Language Processing (Volume 1: Long Papers)}, pages 3403--3417,
  Online. Association for Computational Linguistics.

\bibitem[{Zheng et~al.(2024)Zheng, Fan, and Li}]{zheng-etal-2024-incorporating}
Jianyu Zheng, Fengfei Fan, and Jianquan Li. 2024.
\newblock \href {https://aclanthology.org/2024.lrec-main.787/} {Incorporating
  lexical and syntactic knowledge for unsupervised cross-lingual transfer}.
\newblock In \emph{Proceedings of the 2024 Joint International Conference on
  Computational Linguistics, Language Resources and Evaluation (LREC-COLING
  2024)}, pages 8986--8997, Torino, Italia. ELRA and ICCL.

\bibitem[{Zhou et~al.(2025)Zhou, Zhao, Wang, Zhang, Sun, Wang, and
  Qin}]{10890024}
Jiaming Zhou, Shiwan Zhao, Hui Wang, Tian-Hao Zhang, Haoqin Sun, Xuechen Wang,
  and Yong Qin. 2025.
\newblock \href {https://doi.org/10.1109/ICASSP49660.2025.10890024} {Improving
  zero-shot {Chinese}-{English} code-switching {ASR} with {kNN}-{CTC} and gated
  monolingual datastores}.
\newblock In \emph{ICASSP 2025 - 2025 IEEE International Conference on
  Acoustics, Speech and Signal Processing (ICASSP)}, pages 1--5.

\bibitem[{Zhou et~al.(2020{\natexlab{a}})Zhou, Yılmaz, Long, Li, and
  Li}]{zhou20b_interspeech}
Xinyuan Zhou, Emre Yılmaz, Yanhua Long, Yijie Li, and Haizhou Li.
  2020{\natexlab{a}}.
\newblock \href {https://doi.org/10.21437/Interspeech.2020-2488}
  {{Multi-Encoder-Decoder Transformer for Code-Switching Speech Recognition}}.
\newblock In \emph{{Interspeech 2020}}, pages 1042--1046.

\bibitem[{Zhou et~al.(2020{\natexlab{b}})Zhou, Tian, Lee, Das, and
  Li}]{9054722}
Xuehao Zhou, Xiaohai Tian, Grandee Lee, Rohan~Kumar Das, and Haizhou Li.
  2020{\natexlab{b}}.
\newblock \href {https://doi.org/10.1109/ICASSP40776.2020.9054722} {End-to-end
  code-switching {TTS} with cross-lingual language model}.
\newblock In \emph{ICASSP 2020 - 2020 IEEE International Conference on
  Acoustics, Speech and Signal Processing (ICASSP)}, pages 7614--7618.

\bibitem[{Zhu et~al.(2024)Zhu, Cheng, Chen, Zhuang, Huang, and
  Zou}]{zhu-etal-2024-code}
Zhihong Zhu, Xuxin Cheng, Zhanpeng Chen, Xianwei Zhuang, Zhiqi Huang, and
  Yuexian Zou. 2024.
\newblock \href {https://doi.org/10.18653/v1/2024.acl-short.15} {Code-switching
  can be better aligners: Advancing cross-lingual {SLU} through
  representation-level and prediction-level alignment}.
\newblock In \emph{Proceedings of the 62nd Annual Meeting of the Association
  for Computational Linguistics (Volume 2: Short Papers)}, pages 153--160,
  Bangkok, Thailand. Association for Computational Linguistics.

\end{thebibliography}

\appendix
\section{Methodology}\label{app:method}
This appendix describes how we collected, screened and classified the papers in this survey.

\subsection{Search Protocol and Funnel}\label{app:search}
\paragraph{Sources and cutoff.} We searched Semantic Scholar with 15 query pools. They cover broad code-mixing text, transfer words, synthetic generation, LLM and pretrained-model work, language identification and transliteration, code-mixed-to-native augmentation without the words ``code-switching'', Arabic dialect mixing, bilingual speech, mixed-script text, code-switched speech (core, recent and transfer), and language-pair names. We also scraped papers from the ACL Anthology, and chased citations one hop from 67 seed papers. Interspeech abstracts were recovered through DOI pages, and other missing abstracts through OpenAlex. The search was completed in early October 2026. The corpus therefore covers papers available up to that point, and the newest is an arXiv preprint from October 2026.

\paragraph{Funnel.} The 15 pools returned 23,186 hits, or 9,803 unique papers. Scraping the ACL Anthology gave 1,132 hits, and citation chasing found 2,833 citing papers. Together they added 1,284 papers not found by the search. Pooling all sources thus gave 11,087 unique papers. A rule-based filter kept 4,758 candidates, of which only 48 came from citation chasing alone. Table~\ref{tab:funnel} shows each step.

\begin{table}[ht]\centering\small\begin{adjustbox}{max width=\columnwidth}\begin{tabular}{lr}\toprule Step & Papers \\\midrule
Search hits (15 pools, with duplicates) & 23,186 \\
Unique papers from search & 9,803 \\
\quad added by ACL Anthology and citations & 1,284 \\
Pooled unique papers & 11,087 \\
Candidates after rule-based filter & 4,758 \\
Labelled (title + abstract) by an LLM & 4,758 \\
In scope (yes) & 1,476 \\
\quad removed by keyword audit & 25 \\
\quad included in the survey & 1,451 \\
Maybe (no stated transfer) & 1,681 \\
Out of scope & 1,601 \\
\bottomrule\end{tabular}\end{adjustbox}\caption{Funnel from search to included papers.}\label{tab:funnel}\end{table}

\paragraph{Inclusion.} A paper is in scope if it concerns the CM setting or its native constituents and states one of the five kinds of transfer defined in Section~\ref{sec:background} in its title or abstract. A CM paper stating no transfer is marked ``maybe''; the rest are out of scope.

\subsection{Screening and Classification}\label{app:classify}
\paragraph{LLM as a judge.} We used an LLM, Claude Haiku 5.5 (\texttt{claude-haiku-5-5}), as a judge. It read each candidate's title and abstract, or the title alone when no abstract was available, without full-text reading. It judged 1,476 papers in scope, 1,681 ``maybe'' and 1,601 out of scope. For each in-scope paper, it also assigned direction, mechanism, task and modality (Appendix~\ref{app:schema}). Era and language-pair group come from rules on year and language codes. Table~\ref{tab:direction-modality} gives the resulting classification by direction and modality. Tables~\ref{tab:mechanism-modality} and~\ref{tab:era-direction} give it by mechanism and by era. Appendix~\ref{app:humaneval} reports a human evaluation of these labels.

\paragraph{Keyword audit.} We then audited the in-scope set with rules. A paper was kept only if its title or abstract contained a code-mixing keyword, such as \emph{code-mixing} or \emph{code-switching}. This removed 25 papers that are not about code-mixing, for example multilingual ASR on monolingual utterances, or romanised text without mixing. The survey thus includes 1,451 papers (860 text, 570 speech, 21 multimodal).

\begin{table}[ht]\centering\small\setlength{\tabcolsep}{4pt}\begin{adjustbox}{max width=\columnwidth}\begin{tabular}{lrrrr}
\toprule
 & TEXT & SPEECH & MULTIMODAL & Total \\
\midrule
N2CM & 559 & 473 & 13 & 1045 \\
CM2N & 78 & 2 & 2 & 82 \\
BIDIR & 0 & 5 & 0 & 5 \\
ANALYSIS & 191 & 86 & 5 & 282 \\
NONE & 32 & 4 & 1 & 37 \\
\bottomrule
\end{tabular}
\end{adjustbox}\caption{Included papers per transfer direction and modality.}\label{tab:direction-modality}\end{table}

\begin{table}[ht]\centering\small\setlength{\tabcolsep}{4pt}\begin{adjustbox}{max width=\columnwidth}\begin{tabular}{lrrrr}
\toprule
 & TEXT & SPEECH & MULTIMODAL & Total \\
\midrule
FINETUNE & 176 & 82 & 4 & 262 \\
EVAL\_ANALYSIS & 160 & 76 & 4 & 240 \\
PIPELINE\_MAP & 128 & 71 & 2 & 201 \\
AUG\_SYN & 109 & 71 & 0 & 180 \\
RESOURCE & 69 & 38 & 2 & 109 \\
MONO\_DATA & 23 & 78 & 0 & 101 \\
PRETRAIN & 44 & 29 & 3 & 76 \\
EMBED\_ALIGN & 38 & 29 & 1 & 68 \\
ADAPT\_MERGE & 12 & 53 & 0 & 65 \\
MULTITASK & 21 & 28 & 2 & 51 \\
AUG\_SUB & 37 & 1 & 0 & 38 \\
PROMPT\_ICL & 20 & 4 & 0 & 24 \\
DISTILL & 0 & 3 & 2 & 5 \\
OTHER & 23 & 7 & 1 & 31 \\
\bottomrule
\end{tabular}
\end{adjustbox}\caption{Included papers per mechanism and modality.}\label{tab:mechanism-modality}\end{table}

\begin{table}[ht]\centering\small\setlength{\tabcolsep}{4pt}\begin{adjustbox}{max width=\columnwidth}\begin{tabular}{lrrrrrr}
\toprule
 & N2CM & CM2N & BIDIR & ANALYSIS & NONE & Total \\
\midrule
PRE & 73 & 1 & 0 & 8 & 3 & 85 \\
NEURAL & 104 & 2 & 0 & 11 & 4 & 121 \\
PLM & 365 & 27 & 3 & 40 & 6 & 441 \\
LLM & 503 & 52 & 2 & 223 & 24 & 804 \\
\bottomrule
\end{tabular}
\end{adjustbox}\caption{Included papers per era and transfer direction.}\label{tab:era-direction}\end{table}

\subsection{Human Evaluation}\label{app:humaneval}
To check the final labels, we drew 100 papers at random from the 1,451 included papers. For each paper, an author compared its scope, direction, mechanism, task and modality labels with the paper itself. Two papers were labelled incorrectly, and three were partially incorrect, where part of the labelling did not match the paper. The other 95 papers were labelled correctly.

\subsection{Citation Selection and Analysis}\label{app:citesel}
\paragraph{Citation selection.} We cite only part of the included papers, together with earlier surveys that fall outside our corpus. Some surveys in Table~\ref{tab:surveys} are themselves part of the corpus. Appendix~\ref{app:papers} lists the most-cited papers per direction and modality. The others were chosen while writing, as examples of each mechanism and as sources for specific results. We applied no fixed rule to them.

\paragraph{Citation analysis.} The cross-citation figures in \S\ref{sec:intro} and \S\ref{sec:findings} are computed over the 653 papers whose reference lists Semantic Scholar provides, counting only links inside our corpus. For the text--speech figures, we compare the observed number of cross-modality links with the number expected if each paper cited text and speech papers in proportion to their share of the corpus. The per-mechanism figures in \S\ref{sec:findings} count only links between papers with the same mechanism label.

\section{Taxonomy Tree and Earlier Surveys}\label{app:taxonomy}
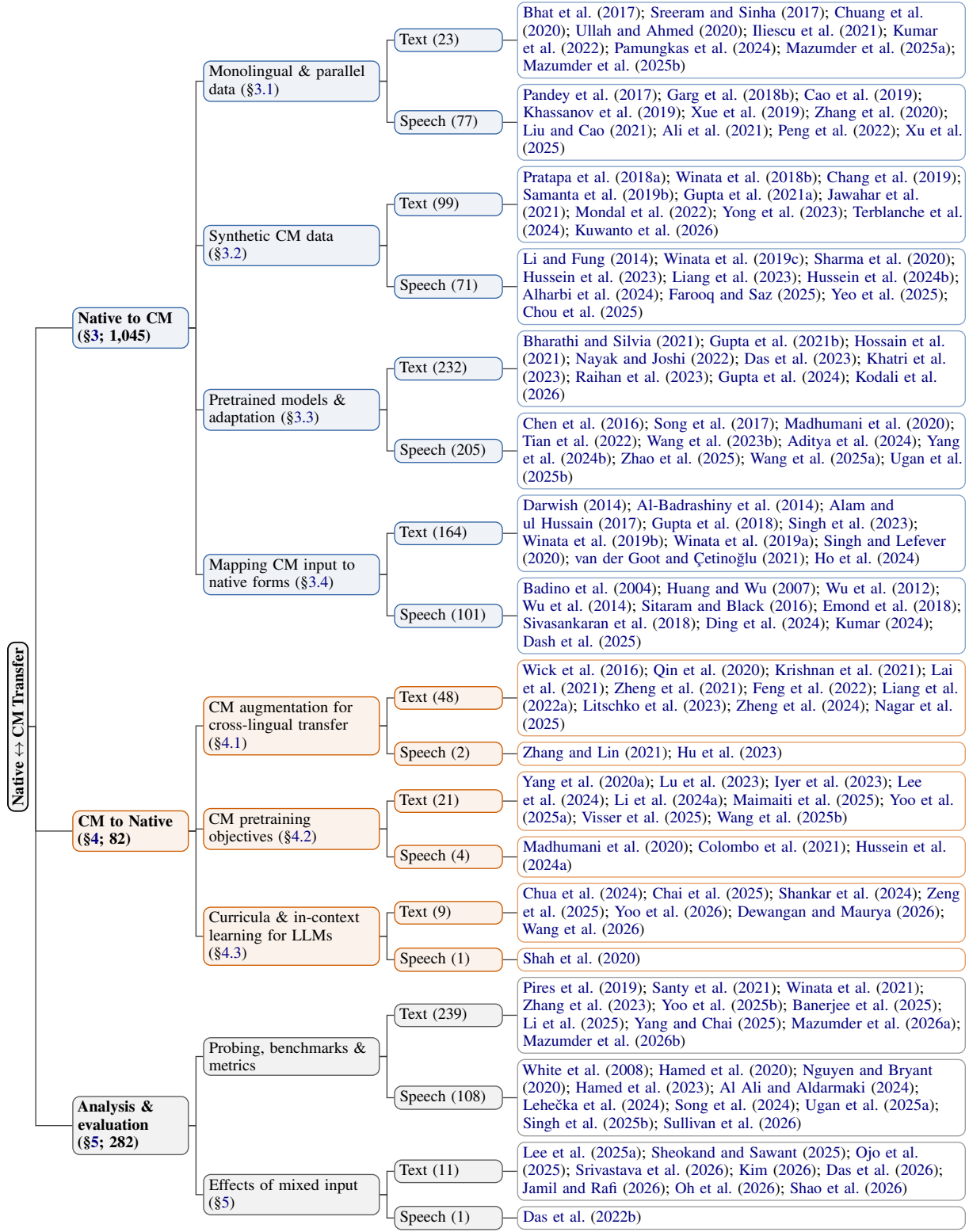
\begin{figure*}[p]
\centering
\hyphenpenalty=10000\exhyphenpenalty=10000
\begin{adjustbox}{max width=\textwidth, max totalheight=0.9\textheight}
\begin{forest}
  forked edges,
  for tree={
    grow=east, reversed=true, anchor=west, parent anchor=east,
    child anchor=west, font=\footnotesize, rectangle,
    rounded corners, node options={align=left}, minimum width=2.5em,
    edge+={darkgray, line width=0.6pt}, s sep=3pt, l sep=8pt, fork sep=4pt,
    inner xsep=3pt, inner ysep=2.5pt, line width=0.6pt,
    ver/.style={rotate=90, child anchor=north, parent anchor=south, anchor=center},
  },
  where level=1{text width=5.4em, font=\footnotesize\bfseries}{},
  where level=2{text width=8.5em}{},
  where level=3{text width=5.0em}{},
  where level=4{text width=22.5em}{},
  [{Native $\leftrightarrow$ CM Transfer}, ver, root, font=\small\bfseries
    % ---------------- Native -> CM ----------------
    [{Native to CM\\(\S\ref{sec:n2cm-data}; 1,045)}, nc
      [{Monolingual \& parallel data (\S\ref{sec:n2cm-mono})}, nc
        [{Text (23)}, nc
          [{\citet{bhat-etal-2017-joining}; \citet{sreeram2017language}; \citet{9053775}; \citet{ullah2020code}; \citet{iliescu-etal-2021-much}; \citet{kumar-etal-2022-utilizing}; \citet{10.1145/3639233.3639247}; \citet{10.1145/3726866}; \citet{mazumder-etal-2025-revealing}}, leafnc]
        ]
        [{Speech (77)}, nc
          [{\citet{8300583}; \citet{garg18_interspeech}; \citet{8682927}; \citet{khassanov19_interspeech}; \citet{xue19_interspeech}; \citet{zhang20r_interspeech}; \citet{9414158}; \citet{ali21b_interspeech}; \citet{peng2022internal}; \citet{11434663}}, leafnc]
        ]
      ]
      [{Synthetic CM data (\S\ref{sec:n2cm-synth})}, nc
        [{Text (99)}, nc
          [{\citet{pratapa-etal-2018-language}; \citet{winata2018learn}; \citet{chang19_interspeech}; \citet{ijcai2019p719}; \citet{gupta-etal-2021-training}; \citet{jawahar-etal-2021-exploring}; \citet{mondal-etal-2022-cocoa}; \citet{yong-etal-2023-prompting}; \citet{terblanche-etal-2024-prompting}; \citet{kuwanto-etal-2026-linguistics}}, leafnc]
        ]
        [{Speech (71)}, nc
          [{\citet{li-fung-2014-language}; \citet{winata-etal-2019-code}; \citet{sharma20c_interspeech}; \citet{10023313}; \citet{liang2023improving}; \citet{10446857}; \citet{alharbi24_syndata4genai}; \citet{11204332}; \citet{11249050}; \citet{11434800}}, leafnc]
        ]
      ]
      [{Pretrained models \& adaptation (\S\ref{sec:n2cm-models})}, nc
        [{Text (232)}, nc
          [{\citet{b-a-2021-ssncse}; \citet{gupta-etal-2021-task}; \citet{hossain-etal-2021-nlp-cuet}; \citet{nayak-joshi-2022-l3cube}; \citet{das-etal-2023-improving}; \citet{khatri-etal-2023-translate}; \citet{raihan2023mixed}; \citet{gupta2024code}; \citet{10.1145/3799830.3799852}}, leafnc]
        ]
        [{Speech (205)}, nc
          [{\citet{e99-d_10_2554}; \citet{8300538}; \citet{madhumani2020learning}; \citet{tian2022lae}; \citet{wang2023languageroutingmixtureexpertsmultilingual}; \citet{10446258}; \citet{10849308}; \citet{10889634}; \citet{10890869}; \citet{ugan2025adapting}}, leafnc]
        ]
      ]
      [{Mapping CM input to native forms (\S\ref{sec:n2cm-map})}, nc
        [{Text (164)}, nc
          [{\citet{darwish-2014-arabizi}; \citet{al-badrashiny-etal-2014-automatic}; \citet{8289449}; \citet{gupta-etal-2018-transliteration}; \citet{10.1007/978-3-031-23793-5_30}; \citet{winata-etal-2019-hierarchical}; \citet{winata-etal-2019-learning}; \citet{singh-lefever-2020-sentiment}; \citet{van-der-goot-cetinoglu-2021-lexical}; \citet{10.1145/3651781.3651819}}, leafnc]
        ]
        [{Speech (101)}, nc
          [{\citet{badino04b_interspeech}; \citet{4288089}; \citet{6289009}; \citet{6763085}; \citet{sitaram-black-2016-speech}; \citet{8639699}; \citet{sivasankaran-etal-2018-phone}; \citet{10832233}; \citet{kumar2024two}; \citet{dash2025generative}}, leafnc]
        ]
      ]
    ]
    % ---------------- CM -> Native ----------------
    [{CM to Native\\(\S\ref{sec:cm2n}; 82)}, cn
      [{CM augmentation for cross-lingual transfer (\S\ref{sec:cm2n_aug})}, cn
        [{Text (48)}, cn
          [{\citet{3016100.3016300}; \citet{ijcai2020p533}; \citet{krishnan-etal-2021-multilingual}; \citet{lai-etal-2021-saliency-based}; \citet{zheng-etal-2021-consistency}; \citet{feng-etal-2022-toward}; \citet{liang-etal-2022-label}; \citet{litschko-etal-2023-boosting}; \citet{zheng-etal-2024-incorporating}; \citet{nagar2025breaking}}, leafcn]
        ]
        [{Speech (2)}, cn
          [{\citet{zhang2021improve}; \citet{10389644}}, leafcn]
        ]
      ]
      [{CM pretraining objectives (\S\ref{sec:cm2n_pretrain})}, cn
        [{Text (21)}, cn
          [{\citet{yang2020alternating}; \citet{lu-etal-2023-take}; \citet{iyer-etal-2023-code}; \citet{lee-etal-2024-commit}; \citet{li-etal-2024-prealign}; \citet{MAIMAITI2025113919}; \citet{yoo-etal-2025-code-switching}; \citet{visser2025code}; \citet{wang-etal-2025-investigating-scaling}}, leafcn]
        ]
        [{Speech (4)}, cn
          [{\citet{madhumani2020learning}; \citet{colombo-etal-2021-code}; \citet{hussein24_interspeech}}, leafcn]
        ]
      ]
      [{Curricula \& in-context learning for LLMs (\S\ref{sec:cm2n_llm})}, cn
        [{Text (9)}, cn
          [{\citet{chua2024crosslingual}; \citet{chai2025xcot}; \citet{shankar-etal-2024-context}; \citet{11033355}; \citet{yoo2026gradual}; \citet{dewangan-maurya-2026-claocs}; \citet{wang2026efficient}}, leafcn]
        ]
        [{Speech (1)}, cn
          [{\citet{shah2020learning}}, leafcn]
        ]
      ]
    ]
    % ---------------- Analysis ----------------
    [{Analysis \&\\evaluation\\(\S\ref{sec:analysis}; 282)}, an
      [{Probing, benchmarks \& metrics}, an
        [{Text (239)}, an
          [{\citet{pires-etal-2019-multilingual}; \citet{santy-etal-2021-bertologicomix}; \citet{winata-etal-2021-multilingual}; \citet{zhang-etal-2023-multilingual}; \citet{yoo-etal-2025-code}; \citet{banerjee2025attributional}; \citet{li-etal-2025-impact}; \citet{yang-chai-2025-codemixbench}; \citet{mazumder2026evaluating}; \citet{mazumder2026neither}}, leafan]
        ]
        [{Speech (108)}, an
          [{\citet{white08_interspeech}; \citet{hamed-etal-2020-arzen}; \citet{nguyen-bryant-2020-canvec}; \citet{10023181}; \citet{al-ali-aldarmaki-2024-mixat}; \citet{lehecka24_interspeech}; \citet{song2024comparative}; \citet{10889660}; \citet{SINGH2025111886}; \citet{sullivan2026nadi}}, leafan]
        ]
      ]
      [{Effects of mixed input (\S\ref{sec:cm2n_effects})}, an
        [{Text (11)}, an
          [{\citet{lee-etal-2025-controlling}; \citet{sheokand2025codemixbench}; \citet{ojo2025divers}; \citet{11679710}; \citet{article}; \citet{das2026indi}; \citet{jamil-rafi-2026-code}; \citet{oh-etal-2026-ola}; \citet{shao-etal-2026-role}}, leafan]
        ]
        [{Speech (1)}, an
          [{\citet{das2022analysis}}, leafan]
        ]
      ]
    ]
  ]\end{forest}
\end{adjustbox}
\caption{Taxonomy of transfer between the code-mixed (CM) setting and its native languages. Branches follow transfer direction, then route, then modality; leaves list up to ten representative papers cited in this survey, ordered by year. Direction counts give papers with that direction label alone, without the 5 bidirectional papers. Leaf counts place bidirectional papers under both directions, and resource and evaluation papers under Analysis whatever their direction. Blue: native-to-CM; orange: CM-to-native; grey: analysis.}
\label{fig:taxonomy}
\end{figure*}

Figure~\ref{fig:taxonomy} shows the taxonomy tree, with direction, route and modality as its levels. Table~\ref{tab:route-mech} lists the mechanism labels inside each route. Table~\ref{tab:surveys} compares earlier surveys of code-switching with ours.

% Taxonomy tree: direction -> mechanism -> modality -> representative papers.
\definecolor{nccol}{RGB}{52,101,164}
\definecolor{cncol}{RGB}{206,92,0}
\definecolor{ancol}{RGB}{90,90,90}
\tikzset{
  root/.style={draw=black, fill=gray!10, line width=0.8pt},
  nc/.style={draw=nccol, fill=nccol!8},
  cn/.style={draw=cncol, fill=cncol!8},
  an/.style={draw=ancol, fill=ancol!8},
  leafnc/.style={draw=nccol!60, fill=white},
  leafcn/.style={draw=cncol!60, fill=white},
  leafan/.style={draw=ancol!60, fill=white},
}

\begin{table}[t]
\centering\footnotesize
\setlength{\tabcolsep}{2pt}
\begin{tabular}{@{}lcrcccc@{}}
\toprule
Survey & Year & No. & Text & Sp. & Dir. & C2N \\
\midrule
\citet{sitaram2019survey} & 2019 & -- & Y & Y & N & Y \\
\citet{9074205} & 2020 & -- & Y & P & N & N \\
\citet{dogruoz-etal-2021-survey} & 2021 & -- & Y & P & N & N \\
\citet{winata-etal-2023-decades} & 2023 & 400+ & Y & Y & N & P \\
\citet{hamed-etal-2025-survey} & 2025 & -- & Y & Y & N & N \\
\citet{11255834} & 2025 & -- & N & Y & N & N \\
\citet{sheth-etal-2026-beyond} & 2026 & 327 & Y & Y & N & P \\
\citet{agro-etal-2026-code} & 2026 & 127 & N & Y & N & P \\
\citet{gupta2026code} & 2026 & -- & Y & N & P & Y \\
\midrule
\textbf{Ours} & 2026 & 1,451 & Y & Y & Y & Y \\
\bottomrule
\end{tabular}
\caption{Surveys of code-switching compared. No.: number of reviewed papers, if stated. Sp.: speech. Dir.: organised by transfer direction. C2N: covers CM-to-native transfer. Y yes, P partly, N no.}
\label{tab:surveys}
\end{table}

\begin{table}[t]
\centering\footnotesize
\setlength{\tabcolsep}{3pt}
\begin{tabular}{@{}>{\raggedright\arraybackslash}p{0.47\columnwidth}>{\raggedright\arraybackslash}p{0.49\columnwidth}@{}}
\toprule
Route (section) & Mechanism labels \\
\midrule
\multicolumn{2}{@{}l}{\textit{Native to CM}} \\
Monolingual \& parallel data (\S\ref{sec:n2cm-mono}) & Monolingual data \\
Synthetic CM data (\S\ref{sec:n2cm-synth}) & Synthetic CM, substitution \\
Pretrained models \& adaptation (\S\ref{sec:n2cm-models}) & Fine-tuning, adapters, pretraining, multitask, prompting, distillation \\
Mapping to native forms (\S\ref{sec:n2cm-map}) & Pipeline mapping, embedding alignment \\
\midrule
\multicolumn{2}{@{}l}{\textit{CM to native}} \\
CM augmentation (\S\ref{sec:cm2n_aug}) & Substitution, synthetic CM \\
CM pretraining objectives (\S\ref{sec:cm2n_pretrain}) & Pretraining, embedding alignment, multitask \\
Curricula \& in-context learning (\S\ref{sec:cm2n_llm}) & Prompting, fine-tuning \\
\midrule
\multicolumn{2}{@{}l}{\textit{Analysis}} \\
Probing, benchmarks \& metrics (\S\ref{sec:analysis}) & Evaluation/analysis, resource \\
Effects of mixed input (\S\ref{sec:cm2n_effects}) & Evaluation/analysis, resource \\
\bottomrule
\end{tabular}
\caption{Mechanism labels grouped under each route of the taxonomy tree (Figure~\ref{fig:taxonomy}), most frequent first. Papers whose main contribution is a resource or an evaluation fall under Analysis, whatever their direction. Papers in the residual class are not in the tree.}
\label{tab:route-mech}
\end{table}

\section{Extended Discussion by Mechanism}\label{app:extended}
This appendix extends each route of \S\ref{sec:n2cm-data}--\S\ref{sec:analysis} with further work. Each paragraph follows its route roughly in time order and ends with what the extra work adds.

\paragraph{Monolingual and parallel data (\S\ref{sec:n2cm-mono}).}
The simplest use of native data is to train on it directly. Early language models did so with Hindi translations of Hinglish. Adding Hindi POS tags as input factors cut perplexity from 318 to 58 \citep{sreeram2017language}. An English--Spanish model later alternated monolingual batches and tied the two embedding spaces with a squared-error loss. Perplexity fell from 300 with Spanish data alone to 80 \citep{ullah2020code}. Classification and parsing took up the same idea. Native treebanks parse Hinglish at least ten LAS points better than a monolingual parser chosen by the matrix language \citep{bhat-etal-2017-joining}. Meta-learning over labelled English and Hindi tasks improves results on GLUECoS \citep{kumar-etal-2022-utilizing}. For humour and sarcasm, native samples raise F1 by up to 6.76\% and 8.64\% relative. Multi-task training with hate detection raises these gains to 10.67\% and 12.35\% \citep{mazumder-etal-2025-revealing}. The Hindi samples there are machine-translated, and humour is partly lost in translation. For hate speech, native samples alone reach 0.60 F1, close to the 0.63 of native and CM samples together. CM samples alone reach at most 0.54 \citep{10.1145/3726866}. Native data is therefore more than a fallback when it matches the task.
In speech, early pipelines segmented an utterance, identified each language and ran monolingual recognisers, so errors passed from one stage to the next. Later work trained one model on pooled data. Decoupling pronunciation from language reduces the need for paired CM data \citep{9414428}. Mixing monolingual data in proportion to each language's share of the CM data also helps \citep{hussein2021balanced}. The best model depends on the type of switching. WFST systems suit intersentential CM, and transformers suit intrasentential CM \citep{ali21b_interspeech}. For the language model, order matters. Pretraining on monolingual text and then fine-tuning on CM text helps, but only in that order \citep{gonen-goldberg-2019-language}. Overlap with the test text matters even more. WER is 10.63\% when the test sentences are in the language-model training text, and 41.66\% when they are not \citep{8300583}. Synthesis followed a similar path. Early work mapped frames across speakers \citep{he12_interspeech}, and word-to-phone mapping beat transliteration in the front end \citep{rallabandi17_interspeech}. End-to-end models then added speaker embeddings \citep{xue19_interspeech} and language embeddings \citep{8682927}. One model trained on monolingual speech in many languages can also synthesise code-switched sentences \citep{nekvinda20_interspeech}. A polyglot Telugu--Hindi system reaches MOS 3.59 \citep{8554693}. In both modalities, native data helps most when it matches the target data.

\paragraph{Synthetic CM data (\S\ref{sec:n2cm-synth}).}
When native data is not enough, it can be turned into CM data. Linguistic theory came first. A human evaluation found grammar-based generators at least as good as heuristic ones \citep{pratapa-choudhury-2021-comparing}. Neural generators followed, including a deep generative model \citep{ijcai2019p719} and an encoder--decoder that uses pretrained features \citep{gupta-etal-2020-semi}. Many pipelines align parallel sentences and replace the aligned words \citep{appicharla-etal-2021-iitp,chatterjee-etal-2022-pacman,kartik-etal-2024-synthetic}. The same route can carry sentiment labels from monolingual to synthetic CM text \citep{samanta-etal-2019-improved}. Speech transcripts offer another source. Swapping intonation units between a transcript and its translation yields CM text for English and 13 other languages \citep{kang-2024-covoswitch}. On this data, MT models copy English tokens far more reliably than non-English ones. LLMs are now a common generator \citep{heredia-etal-2026-conditioning,11306433}. Their output varies by pair, and GPT writes far better Afrikaans--English than Yoruba--English \citep{terblanche-etal-2024-prompting}.
Speech uses generated text in the same way. LLM-generated text gave 8\% and 5.5\% reductions for Arabic--English \citep{alharbi24_syndata4genai}. Injecting generated text into a Mandarin--English streaming transducer gave a 16\% relative reduction in token error rate \citep{10096317}. Other text generators also serve speech \citep{kronis-etal-2024-code}. Audio itself can be synthesised, through TTS and splicing \citep{11204332,lee-etal-2025-unicom}. Speech Collage reduces WER by 16.2\% zero-shot \citep{10446857}. Fine-tuning multilingual TTS on CM corpora lowers MER from 12.1\% to 10.1\% and from 17.8\% to 16.0\% \citep{11249050}. Other work steers synthesis toward correct language boundaries \citep{yeo2026improving}. Synthetic audio can also expose weaknesses, as Whisper hallucinates on synthetic CM speech \citep{mitra-etal-2026-llm}. Synthetic data thus helps in both modalities, but its quality still has to be checked.

\paragraph{Pretrained models (\S\ref{sec:n2cm-models}).}
Most text systems now start from a multilingual encoder. Further work fine-tunes such encoders on Kannada--English and Tamil--English text \citep{hande2021hope,krasitskii-etal-2025-advancing}. A model built for the right languages can still win, as IndoBERTweet does on Indonesian--Javanese--English \citep{10730138}. Other work continues pretraining on CM corpora \citep{xie2025enhancing}, and small DistilBERT-based models compete with mBERT and XLM-R \citep{raihan2023mixed}.
Speech made the same move to large pretrained models. Further work adapts web-scale models to CM speech \citep{peng2023prompting,10929894,biswas25_interspeech}, and some tunes soft prompts to avoid forgetting \citep{yang25p_interspeech}. Language-aware encoders discard context across languages, so later work adds a speech-translation task \citep{10389662}. A streaming system pairs a multilingual first pass with a monolingual second pass and lowers WER by 4.4\% relative \citep{10023346}. A contextualised CTC loss improves ASR on both CM and monolingual speech \citep{9413806}. Mixture-of-experts connectors on speech LLMs use language-specialised experts to raise accuracy on CM speech \citep{10890030}. Combining frozen MMS language adapters serves the same goal \citep{kulkarni2023adapting}. Other adapter designs aim to cut cost \citep{10317410,10627333,suhasini2026parameter}. In both modalities, the pretrained model now does most of the work, and adaptation decides how well it handles switches.

\paragraph{Mapping to native forms (\S\ref{sec:n2cm-map}).}
Mapping avoids CM modelling by converting the input into native forms. Much of this work concerns romanised text. Arabizi is converted to Arabic script \citep{10.1145/3364319,IJECE22767,HAJBI2024e02073}, and Roman Urdu to Urdu script \citep{8289449}. Back-transliteration after word-level language identification appeared early \citep{10.1145/2824864.2824871}. Later pipelines combine transliteration and translation. For Maltese, mixed pipelines beat applying either one to every word \citep{micallef-etal-2024-cross}. Bengali--English becomes Bengali by translating the English segments and back-transliterating the Bengali ones \citep{10.1145/3368567.3368579}. Neural models translate Kannada--English into monolingual text \citep{11167161}, while LLMs still translate Singlish messages poorly \citep{ng2024talking}. A bilingual dictionary can also translate only the sentiment words of Chinese--English text \citep{10.1007/978-3-319-99501-4_37}. Such pipelines need to know which languages are present. MaskLID extends language identification to several languages in one sentence \citep{kargaran2024masklid}. Embedding methods instead map mixed and monolingual sentences into a shared space for sentiment analysis \citep{10.1007/978-3-031-23804-8_9,yadav2020unsupervised}.
In speech, mapping works on sound units. Phones of the two languages can be merged using IPA and acoustic distance \citep{6289015}, and HMM states or Gaussians can be shared \citep{6423531,6853590}. Homophones can also be merged \citep{srivastava18_interspeech}. On the output side, one system projects transcripts into Devanagari \citep{kumar2024two}. Another merges Latin and Indic spellings of the same word into one lexicon and language model \citep{wiesner21_interspeech}. Language diarization labels the language of each frame in Mandarin--English speech \citep{6625316}. TTS often maps mixed text into one phonetic space \citep{badino04b_interspeech,sitaram-black-2016-speech,sitaram16_ssw,manghat20_interspeech}. Other TTS work keeps separate phone sets and learns a mixlingual grapheme-to-phoneme model \citep{bansal20_interspeech}. Embedding alignment in speech covers cross-lingual word embeddings \citep{9054722}. Mapping thus trades CM modelling for the errors of each conversion step.

\paragraph{CM augmentation (\S\ref{sec:cm2n_aug}).}
In the reverse direction, augmentation refines how words are switched. Further work changes the substitution itself \citep{wang-etal-2023-self-augmentation,li2024improving}. One version switches only named entities \citep{whitehouse-etal-2022-entitycs}. Soft code-switching mixes the embeddings of original and replacement words rather than the words \citep{10.1145/3459637.3482176}. For dialogue, ChatZero builds a pseudo-target version of source dialogues from dictionary words and placeholders. It then aligns the three versions with contrastive learning \citep{doi:10.3233/FAIA240950}. The recipe now serves dialogue state tracking and language understanding \citep{10.1145/3575803,10.1145/3490725.3490737,yu-etal-2023-improving,haisa2024phrase}, aspect-based sentiment \citep{vsmid2026generative,fehle2026zero} and low-resource MT \citep{kuwanto2023low,robinson-etal-2022-data,jasim-etal-2020-phraseout}. For cross-lingual SLU, code-switched utterances are blended with manifold mixup \citep{10393998} or aligned with original sentences at the representation and prediction levels \citep{zhu-etal-2024-code}. Others add contrastive objectives \citep{10534816} or an auxiliary translation task \citep{cheng23b_interspeech}. Retrieval shows why such alignment matters. Naive code-mixing can hurt it, but a contrastive term that aligns English and mixed queries restores zero-shot gains \citep{do-etal-2024-contrastivemix}. Code-switching also improves cross-lingual sentence embeddings \citep{10692736}. Augmentation thus helps most when the switched input is tied back to the original.

\paragraph{CM pretraining (\S\ref{sec:cm2n_pretrain}).}
Pretraining objectives build switching into the model from the start. Further work mixes languages during pretraining or continued pretraining for MT and cross-lingual transfer \citep{yang-etal-2020-csp,zan2022bridging,iyer-etal-2023-code,visser2025code}. Code-switched continual pretraining improves semantic retrieval across 20+ languages \citep{MAIMAITI2025113919}. Code-switching data also transfers the embeddings of domain-expert models to new languages \citep{hong-etal-2025-migrate}. Input code-switching during pretraining can preserve an alignment set \citep{li-etal-2024-prealign}. In small models trained from scratch, CM curricula mainly align representations across languages \citep{rooryck2026feeding}. As with augmentation, alignment is the common thread.

\paragraph{LLM curricula and prompting (\S\ref{sec:cm2n_llm}).}
With LLMs, switching moves into prompts and curricula. Beyond the work in the main text, code-switched prompts also help cross-lingual triplet extraction \citep{dewangan-maurya-2026-claocs}. Speech has no counterpart yet, so this route remains a text-only lever.

\paragraph{Analysis (\S\ref{sec:analysis}).}
Analysis work began with probing. Further studies ask how encoders represent CM input \citep{11104253}, and a benchmark compares LoRA-tuned models on language identification, POS tagging and NER \citep{11542676}. Language identification is itself a recurring benchmark. Most work tags single words \citep{zhang-etal-2018-fast-compact,10590279,10.1145/3734947.3735663,11565004}. One such model reports a 19.5\% average absolute gain on three code-mixed datasets \citep{zhang-etal-2018-fast-compact}. Other work tags whole sentences with several language labels \citep{burchell-etal-2024-code}. With LLMs, benchmarks and safety studies took over. GPT-4 fails to beat fine-tuned XLM-R baselines on new code-switched tasks \citep{laureano-de-leon-etal-2025-evaluating}, and automatic scores can miss errors that change meaning \citep{suresh-etal-2025-cs}. Safety studies find the same weakness \citep{song-etal-2025-multilingual,Vijesh2026TanglishGuardBA}. Script matters as well, since hate speech is detected more reliably in the native script than in Roman script \citep{10.1145/3576913}.
Speech analysis centres on corpora and metrics. Corpora include Mixat for Emirati Arabic--English and TEDxTN for Tunisian Arabic \citep{al-ali-aldarmaki-2024-mixat,bougares-etal-2025-tedxtn}. Others cover Turkish--German and Mandarin--English \citep{cetinoglu-2017-code,chua23_interspeech}. A Hindi--English corpus of adult and child speech reports baseline WERs of 16 to 40\% \citep{SINGH2025111886}. Further studies compare models \citep{10446737,vanvuren23_interspeech}. WER is the main point of debate. PolyWER questions it directly \citep{kadaoui-etal-2024-polywer}, and WER inflates quality gaps about threefold because it penalises valid transliterations \citep{abdoli2026benchmarking}. PIER scores only the words of interest, and shows that WER can hide worse recognition of switched words \citep{10889660}. HiKE adopts such metrics and labels loanwords \citep{paik-etal-2026-hike}. TTS benchmarks are few, with SawtArabi as one example \citep{lodagala25_interspeech}. Analysis thus shows where models fail, but the metrics for measuring it are still being settled.

\section{Matched Comparison on SEAME}\label{app:seame}
SEAME is the only CM benchmark that many papers in our corpus share. Table~\ref{tab:seame} lists 11 papers that report absolute error rates on its two standard evaluation sets, dev\_man and dev\_sge. We took every number from the paper's full text. Each row compares a baseline and the best system \emph{within one paper}, so the two share data, model size and decoding. Rows are not directly comparable with each other. Text benchmarks such as GLUECoS and LinCE are used by too few papers in our corpus for a similar comparison.

\begin{table*}[t]
\centering\footnotesize
\setlength{\tabcolsep}{3.2pt}
\begin{tabularx}{\textwidth}{@{}>{\raggedright\arraybackslash}p{2.65cm}>{\raggedright\arraybackslash}X>{\raggedright\arraybackslash}p{3.3cm}rrrrrr@{}}
\toprule
& & & \multicolumn{3}{c}{dev\_man} & \multicolumn{3}{c}{dev\_sge} \\
\cmidrule(lr){4-6}\cmidrule(l){7-9}
Paper & Change over the baseline & Extra data & Base & Best & $\Delta$ & Base & Best & $\Delta$ \\
\midrule
\multicolumn{9}{@{}l}{\textit{Trained from scratch on SEAME}} \\
\citet{zeng19_interspeech}$^\dagger$ & LID multitask, LM rescoring & None & 34.5 & 25.0 & $-$9.5 & 46.4 & 34.5 & $-$11.9 \\
\citet{fan2023language} & Language boundaries & None & 16.96 & 16.29 & $-$0.7 & 23.73 & 22.81 & $-$0.9 \\
\citet{peng2022internal} & Fusion with a CS text LM & CS text for the LM & 16.4 & 15.4 & $-$1.0 & 23.2 & 21.7 & $-$1.5 \\
\midrule
\multicolumn{9}{@{}l}{\textit{Adding native data}} \\
\citet{zhou20b_interspeech}$^\dagger$ & Language-specific encoders & AISHELL-2, LibriSpeech & 18.6 & 16.7 & $-$1.9 & 25.9 & 23.1 & $-$2.8 \\
\citet{9413562} & Monolingual data in training & AISHELL-1, TED-LIUM~2 & 20.2 & 18.5 & $-$1.7 & 27.7 & 26.3 & $-$1.4 \\
\citet{10832233} & Native data, same model & AISHELL-2, LibriSpeech & 22.26 & 17.64 & $-$4.6 & 32.30 & 25.35 & $-$7.0 \\
& \quad + romanised output & AISHELL-2, LibriSpeech & 17.64 & 15.22 & $-$2.4 & 25.35 & 21.54 & $-$3.8 \\
\midrule
\multicolumn{9}{@{}l}{\textit{Starting from a pretrained Whisper model}} \\
\citet{10849308} & Whisper vs.\ scratch Conformer & Whisper-small & 16.6 & 14.3 & $-$2.3 & 23.3 & 20.4 & $-$2.9 \\
\citet{10889634} & Encoder refiner, language-aware decoding & Whisper-small & 14.6 & 14.0 & $-$0.6 & 22.2 & 20.6 & $-$1.6 \\
\citet{yang25p_interspeech} & Soft-prompt tuning & Whisper-medium & 11.44 & 11.17 & $-$0.3 & 17.55 & 16.61 & $-$0.9 \\
\citet{11249050} & +100\,h TTS speech & Whisper-large-v3 & 12.1 & 10.1 & $-$2.0 & 17.8 & 16.0 & $-$1.8 \\
\citet{yeo2026improving} & +100\,h CM-guided synthetic speech & Whisper-large-v3 & 12.1 & 8.9 & $-$3.2 & 17.8 & 14.2 & $-$3.6 \\
\bottomrule
\end{tabularx}
\caption{Matched comparisons on SEAME, a Mandarin--English CM speech corpus. Each row gives one paper's baseline (Base) and best system (Best) in mixed error rate (MER, \%). $\Delta$ is the change in points. dev\_man is dominated by Mandarin, dev\_sge by Singapore English. \emph{Extra data} lists added corpora or the pretrained Whisper model. AISHELL is Mandarin speech, and LibriSpeech and TED-LIUM are English. LID: language identification. $^\dagger$Token error rate, computed like MER. Rows are not comparable with each other.}
\label{tab:seame}
\end{table*}

\section{Labelling Schema}\label{app:schema}
Each candidate was labelled from its title and abstract (title only when no abstract was available) with: scope (in scope / maybe / out), transfer direction (native$\to$code-mixed, code-mixed$\to$native, both, analysis-only), a main and optional secondary mechanism from the 14 codes listed below, task family, speech or text, language pair, datasets and a short result. Era and language-pair group were assigned by rules from year and method words. A human evaluation of 100 random included papers found 2 incorrect and 3 partially incorrect labellings (Appendix~\ref{app:humaneval}). The maybe papers were not adjudicated.

\paragraph{Direction.} \emph{Native$\to$CM} (N2CM, 1,045 papers) means monolingual data, models or knowledge are used to help code-mixed (CM) processing. \emph{CM$\to$native} (CM2N, 82) means CM material is used to help monolingual or cross-lingual performance. \emph{Both} (BIDIR, 5) covers two-way transfer. \emph{Analysis-only} (ANALYSIS, 282) covers papers that compare or probe without proposing a transfer method. 37 papers have no direction.

\paragraph{Mechanism.} We use 13 mechanisms plus a residual class. \emph{Monolingual data} (MONO\_DATA) adds plain monolingual or parallel text or speech to training or language-model interpolation \citep{garg18_interspeech,ylmaz18_interspeech,gonen-goldberg-2019-language,bhat-etal-2017-joining}. \emph{Synthetic CM} (AUG\_SYN) builds new CM data from monolingual or parallel data, by linguistic theory, alignment, generative models or LLMs \citep{pratapa-etal-2018-language,chang19_interspeech,winata-etal-2019-code,yong-etal-2023-prompting}. \emph{Substitution} (AUG\_SUB) swaps words or phrases using dictionaries or machine translation \citep{ijcai2020p533,song-etal-2019-code,liu2019attentioninformedmixedlanguagetrainingzeroshot,zhang-etal-2021-cross}. \emph{Pretraining} continues self-supervised training aimed at CM \citep{nayak-joshi-2022-l3cube,liu-etal-2021-continual,garg-etal-2018-code}. \emph{Fine-tuning} transfers through the weights of an existing pretrained model \citep{tsai-etal-2019-small,mathur-etal-2018-detecting,winata-etal-2020-meta,chowdhury21_interspeech}. \emph{Multitask} learning trains jointly with monolingual tasks or auxiliary losses \citep{madhumani2020learning,zeng19_interspeech,winata-etal-2018-code}. \emph{Adapters} (ADAPT\_MERGE) covers adapters, LoRA, mixtures of experts, routing and model merging \citep{lu20f_interspeech,wang2023languageroutingmixtureexpertsmultilingual,tian2022lae}. \emph{Pipeline mapping} (PIPELINE\_MAP) turns CM input into monolingual form or tools through normalisation, transliteration, translation or language-ID routing \citep{darwish-2014-arabizi,al-badrashiny-etal-2014-automatic,barik-etal-2019-normalization,8683223}. \emph{Embedding alignment} (EMBED\_ALIGN) builds shared spaces: cross-lingual or meta-embeddings and a common sentiment space in text \citep{duong-etal-2017-multilingual-semantic,winata-etal-2019-learning,10.1007/978-3-031-23804-8_9}, and, in speech, cross-lingual word embeddings that place both languages in one space for code-switching TTS \citep{9054722}. \emph{Distillation} uses teacher--student training and is rare (5 papers) \citep{raj-khan-etal-2021-towards-developing,10832290}. \emph{Prompting} (PROMPT\_ICL) covers prompts, in-context learning and instruction tuning with CM content \citep{khatri-etal-2023-translate,peng2023prompting,chai2025xcot,shankar-etal-2024-context}. \emph{Evaluation/analysis} (EVAL\_ANALYSIS) covers CM-versus-monolingual comparison, probing and benchmark studies \citep{winata-etal-2021-multilingual,pires-etal-2019-multilingual,zhang-etal-2023-multilingual,vilares-etal-2015-sentiment}. \emph{Resource} papers release datasets or lexicons with an explicit transfer angle \citep{hamed-etal-2020-arzen,srivastava-singh-2020-phinc}. \emph{Other} is the residual class \citep{peng-etal-2014-learning}.

\paragraph{Remaining axes.} \emph{Task family} uses short codes. They are LM (language modelling), ASR (speech recognition), ST (speech translation), SLU (spoken language understanding), TTS (speech synthesis), LID (language identification), SEQ (tagging), CLS (classification), MT (translation), NLU, GEN (generation) and BENCH (benchmarks). \emph{Modality} is text (860 papers), speech (570) or multimodal (21). \emph{Language pair} is grouped into Indic--English, Chinese--English, Arabic--X, South-East Asian--English, Spanish--English, European, African, East Asian--English, multi-pair and other. \emph{Resource setting} records whether CM data is scarce, one constituent is low-resource, large natural CM data exists, or only synthetic CM data is used. \emph{Era} is set by year and methods: before 2014, 2014--2018, 2019--2022, and 2023 onward.

\section{Representative Papers per Cell}\label{app:papers}
Tables~\ref{tab:papers-1}--\ref{tab:papers-6} list the most-cited papers per direction and modality, grouped under headers. Codes are defined in Appendix~\ref{app:schema}. \emph{multi} marks papers covering several unnamed pairs, and -- marks papers naming no pair.

\begin{table}[tp]\centering\footnotesize\setlength{\tabcolsep}{2pt}
\begin{tabular}{@{}>{\raggedright\arraybackslash}p{2.0cm}rll>{\raggedright\arraybackslash}p{0.95cm}@{}}
\toprule Paper & Year & Mechanism & Task & Pair \\\midrule
\multicolumn{5}{@{}l}{\textit{Analysis --- speech}} \\
\citet{sitaram2019survey} & 2019 & EVAL\_ANALYSIS & OTHER & multi \\
\citet{cetinoglu-etal-2016-challenges} & 2016 & EVAL\_ANALYSIS & OTHER & multi \\
\citet{6639083} & 2013 & EVAL\_ANALYSIS & LID & zh-en \\
\citet{weller-etal-2022-end} & 2022 & RESOURCE & ST & es-en \\
\citet{liu21d_interspeech} & 2021 & EVAL\_ANALYSIS & LID & multi \\
\citet{white08_interspeech} & 2008 & EVAL\_ANALYSIS & ASR & -- \\
\citet{chowdhury20c_interspeech} & 2020 & EVAL\_ANALYSIS & ASR & ar-arz \\
\citet{10626762} & 2024 & EVAL\_ANALYSIS & ASR & zh-en \\
\citet{al-ali-aldarmaki-2024-mixat} & 2024 & RESOURCE & ASR & ar-en \\
\citet{10446737} & 2023 & EVAL\_ANALYSIS & BENCH & multi \\
\citet{10889660} & 2025 & EVAL\_ANALYSIS & ASR & multi \\
\citet{10023181} & 2022 & EVAL\_ANALYSIS & ASR & ar-en \\
\citet{11447403} & 2025 & EVAL\_ANALYSIS & ASR & zh-en \\
\citet{baghel23_interspeech} & 2023 & EVAL\_ANALYSIS & SLU & multi \\
\citet{ICLR2026_3122aaa2} & 2025 & EVAL\_ANALYSIS & OTHER & ar-en \\
\citet{bhattacharya24_interspeech} & 2024 & EVAL\_ANALYSIS & OTHER & multi \\
\citet{SINGH2025111886} & 2025 & RESOURCE & ASR & hi-en \\
\citet{li23z_interspeech} & 2023 & RESOURCE & SEQ & zh-en \\
\citet{sullivan2026nadi} & 2026 & RESOURCE & BENCH & ar \\
\citet{bhogale2026voice} & 2026 & RESOURCE & ASR & multi \\
\multicolumn{5}{@{}l}{\textit{Analysis --- text}} \\
\citet{pires-etal-2019-multilingual} & 2019 & EVAL\_ANALYSIS & BENCH & multi \\
\citet{dogruoz-etal-2021-survey} & 2021 & EVAL\_ANALYSIS & OTHER & multi \\
\citet{zhang-etal-2023-multilingual} & 2023 & EVAL\_ANALYSIS & BENCH & multi \\
\citet{winata-etal-2021-multilingual} & 2021 & EVAL\_ANALYSIS & SEQ & multi \\
\citet{vilares-etal-2015-sentiment} & 2015 & EVAL\_ANALYSIS & CLS & es-en \\
\citet{BALY2017266} & 2017 & EVAL\_ANALYSIS & CLS & ar-arz \\
\bottomrule\end{tabular}
\caption{Most-cited papers per direction and modality, part 1 of 6.}\label{tab:papers-1}
\end{table}

\begin{table}[tp]\centering\footnotesize\setlength{\tabcolsep}{2pt}
\begin{tabular}{@{}>{\raggedright\arraybackslash}p{2.0cm}rll>{\raggedright\arraybackslash}p{0.95cm}@{}}
\toprule Paper & Year & Mechanism & Task & Pair \\\midrule
\multicolumn{5}{@{}l}{\textit{Analysis --- text (cont.)}} \\
\citet{das-etal-2022-hate-speech} & 2022 & EVAL\_ANALYSIS & CLS & bn-en \\
\citet{lee-wang-2015-emotion} & 2015 & EVAL\_ANALYSIS & CLS & zh-en \\
\citet{10.1145/3503162.3503179} & 2021 & EVAL\_ANALYSIS & CLS & ta-en; ml-en \\
\citet{santy-etal-2021-bertologicomix} & 2021 & EVAL\_ANALYSIS & CLS & multi \\
\citet{yoo-etal-2025-code} & 2024 & EVAL\_ANALYSIS & CLS & multi \\
\citet{10225623} & 2024 & EVAL\_ANALYSIS & CLS & -- \\
\citet{zhang-etal-2018-fast-compact} & 2018 & EVAL\_ANALYSIS & LID & multi \\
\citet{kumar-etal-2024-semeval} & 2024 & RESOURCE & CLS & hi-en \\
\citet{priyadharshini-etal-2023-overview} & 2023 & RESOURCE & CLS & ta-en; te-en \\
\citet{chandu-etal-2018-language} & 2018 & EVAL\_ANALYSIS & LM & hi-en \\
\citet{7918035} & 2016 & EVAL\_ANALYSIS & LM & -- \\
\citet{song-etal-2025-multilingual} & 2024 & EVAL\_ANALYSIS & OTHER & multi \\
\citet{bhat2016grammatical} & 2016 & EVAL\_ANALYSIS & LM & hi-en \\
\citet{srivastava-singh-2021-hinge} & 2021 & RESOURCE & GEN & hi-en \\
\citet{srivastava-singh-2022-overview} & 2022 & RESOURCE & MT & hi-en \\
\citet{srivastava-singh-2020-phinc} & 2020 & RESOURCE & MT & hi-en \\
\multicolumn{5}{@{}l}{\textit{Both directions --- speech}} \\
\citet{shah2020learning} & 2020 & FINETUNE & ASR & multi \\
\citet{madhumani2020learning} & 2020 & MULTITASK & ASR & multi \\
\citet{hussein24_interspeech} & 2024 & MULTITASK & ASR & multi \\
\citet{9413806} & 2020 & OTHER & ASR & th-en \\
\citet{10389644} & 2023 & AUG\_SYN & ASR & zh-en \\
\multicolumn{5}{@{}l}{\textit{CM to native --- speech}} \\
\citet{colombo-etal-2021-code} & 2021 & PRETRAIN & SLU & multi \\
\citet{zhang2021improve} & 2021 & MONO\_DATA & TTS & zh-en \\
\bottomrule\end{tabular}
\caption{Most-cited papers per direction and modality, part 2 of 6.}\label{tab:papers-2}
\end{table}

\begin{table}[tp]\centering\footnotesize\setlength{\tabcolsep}{2pt}
\begin{tabular}{@{}>{\raggedright\arraybackslash}p{2.0cm}rll>{\raggedright\arraybackslash}p{0.95cm}@{}}
\toprule Paper & Year & Mechanism & Task & Pair \\\midrule
\multicolumn{5}{@{}l}{\textit{CM to native --- text}} \\
\citet{ijcai2020p533} & 2020 & AUG\_SUB & NLU & multi \\
\citet{song-etal-2019-code} & 2019 & AUG\_SUB & MT & ru-en; zh-en \\
\citet{liu2019attentioninformedmixedlanguagetrainingzeroshot} & 2019 & AUG\_SUB & NLU & multi \\
\citet{yang2020alternating} & 2020 & PRETRAIN & LM & multi \\
\citet{zhang-etal-2021-cross} & 2021 & AUG\_SUB & CLS & multi \\
\citet{yang-etal-2020-csp} & 2020 & PRETRAIN & MT & multi \\
\citet{chai2025xcot} & 2024 & PROMPT\_ICL & NLU & multi \\
\citet{zheng-etal-2021-consistency} & 2021 & AUG\_SUB & BENCH & multi \\
\citet{pan-etal-2021-multilingual} & 2020 & AUG\_SUB & BENCH & multi \\
\citet{krishnan-etal-2021-multilingual} & 2021 & AUG\_SUB & NLU & multi \\
\citet{3016100.3016300} & 2016 & AUG\_SUB & OTHER & multi \\
\citet{yoo-etal-2025-code-switching} & 2024 & PRETRAIN & LM & multi \\
\citet{li-etal-2024-prealign} & 2024 & PRETRAIN & LM & multi \\
\citet{wang-etal-2025-investigating-scaling} & 2025 & PRETRAIN & OTHER & multi \\
\citet{liu-etal-2021-continual} & 2021 & PRETRAIN & MT & multi \\
\multicolumn{5}{@{}l}{\textit{Native to CM --- speech}} \\
\citet{6289015} & 2012 & PIPELINE\_MAP & ASR & zh-en \\
\citet{8682674} & 2018 & EMBED\_ALIGN & ASR & ja-en \\
\citet{winata-etal-2019-code} & 2019 & AUG\_SYN & ASR & zh-en \\
\citet{zeng19_interspeech} & 2018 & MULTITASK & ASR & zh-en \\
\citet{8683223} & 2019 & PIPELINE\_MAP & ASR & zh-en \\
\citet{6639306} & 2013 & MULTITASK & LM & zh-en \\
\citet{8682850} & 2019 & MULTITASK & ASR & zh-en \\
\citet{peng2023prompting} & 2023 & PROMPT\_ICL & ASR & multi \\
\citet{lu20f_interspeech} & 2020 & ADAPT\_MERGE & ASR & zh-en \\
\citet{ylmaz18_interspeech} & 2018 & MONO\_DATA & ASR & fy-nl \\
\citet{nekvinda20_interspeech} & 2020 & MONO\_DATA & TTS & multi \\
\bottomrule\end{tabular}
\caption{Most-cited papers per direction and modality, part 3 of 6.}\label{tab:papers-3}
\end{table}

\begin{table}[tp]\centering\footnotesize\setlength{\tabcolsep}{2pt}
\begin{tabular}{@{}>{\raggedright\arraybackslash}p{2.0cm}rll>{\raggedright\arraybackslash}p{0.95cm}@{}}
\toprule Paper & Year & Mechanism & Task & Pair \\\midrule
\multicolumn{5}{@{}l}{\textit{Native to CM --- speech (cont.)}} \\
\citet{winata-etal-2020-meta} & 2020 & FINETUNE & ASR & zh-en \\
\citet{zhou20b_interspeech} & 2020 & MONO\_DATA & ASR & zh-en \\
\citet{9413562} & 2020 & MULTITASK & ASR & zh-en \\
\citet{6639094} & 2013 & AUG\_SYN & ASR & -- \\
\citet{luo2018towards} & 2018 & MULTITASK & ASR & zh-en \\
\citet{6473715} & 2012 & ADAPT\_MERGE & ASR & ms-en \\
\citet{li-fung-2014-language} & 2014 & AUG\_SYN & LM & zh-en \\
\citet{wang2023languageroutingmixtureexpertsmultilingual} & 2023 & ADAPT\_MERGE & ASR & multi \\
\citet{hamed-etal-2020-arzen} & 2020 & RESOURCE & ASR & ar-en \\
\citet{HAMED2022101278} & 2021 & ADAPT\_MERGE & ASR & ar-en \\
\citet{guo18b_interspeech} & 2018 & OTHER & ASR & zh-en \\
\citet{chowdhury21_interspeech} & 2021 & FINETUNE & ASR & multi \\
\citet{slottje2021using} & 2021 & MONO\_DATA & ASR & zh-en \\
\citet{10023346} & 2023 & FINETUNE & ASR & multi \\
\citet{10037997} & 2022 & ADAPT\_MERGE & ASR & zh-en \\
\citet{10.1145/3746027.3764194} & 2025 & DISTILL & BENCH & multi \\
\citet{9060847} & 2019 & PIPELINE\_MAP & ST & ja-en \\
\citet{alastruey-etal-2023-towards} & 2023 & FINETUNE & ST & es-en \\
\citet{dash2025generative} & 2025 & PIPELINE\_MAP & ASR & ta-en \\
\citet{10022517} & 2023 & AUG\_SYN & LM & multi \\
\citet{ugan2025adapting} & 2025 & FINETUNE & ASR & multi \\
\citet{singh25d_interspeech} & 2025 & PRETRAIN & OTHER & multi \\
\citet{shastri2022adversarial} & 2022 & AUG\_SYN & LID & hi-en \\
\citet{fan2023language} & 2023 & FINETUNE & ASR & zh-en \\
\citet{blevins-2026-systematic} & 2026 & PIPELINE\_MAP & ASR & de-en \\
\citet{10450050} & 2023 & PIPELINE\_MAP & ST & hi-en \\
\bottomrule\end{tabular}
\caption{Most-cited papers per direction and modality, part 4 of 6.}\label{tab:papers-4}
\end{table}

\begin{table}[tp]\centering\footnotesize\setlength{\tabcolsep}{2pt}
\begin{tabular}{@{}>{\raggedright\arraybackslash}p{2.0cm}rll>{\raggedright\arraybackslash}p{0.95cm}@{}}
\toprule Paper & Year & Mechanism & Task & Pair \\\midrule
\multicolumn{5}{@{}l}{\textit{Native to CM --- speech (cont.)}} \\
\citet{paik2026towards} & 2026 & ADAPT\_MERGE & ASR & multi \\
\citet{10390242} & 2023 & FINETUNE & SEQ & id-en; ms-en \\
\citet{liang2022npu} & 2022 & ADAPT\_MERGE & ASR & zh-en \\
\citet{lee2025sage} & 2025 & PRETRAIN & LID & multi \\
\citet{app11199106} & 2021 & MULTITASK & ASR & zh-en \\
\citet{sailor21_interspeech} & 2021 & MULTITASK & ASR & multi \\
\citet{ye2024sc} & 2024 & ADAPT\_MERGE & ASR & zh-en \\
\citet{hsu2025breezyvoice} & 2025 & FINETUNE & TTS & zh \\
\multicolumn{5}{@{}l}{\textit{Native to CM --- text}} \\
\citet{khanuja-etal-2020-gluecos} & 2020 & AUG\_SYN & BENCH & hi-en; es-en \\
\citet{pratapa-etal-2018-language} & 2018 & AUG\_SYN & LM & es-en \\
\citet{solorio-liu-2008-learning} & 2008 & AUG\_SYN & GEN & es-en \\
\citet{mathur-etal-2018-detecting} & 2018 & FINETUNE & CLS & hi-en \\
\citet{mathur-etal-2018-offend} & 2018 & FINETUNE & CLS & hi-en \\
\citet{darwish-2014-arabizi} & 2013 & PIPELINE\_MAP & OTHER & ar \\
\citet{9220904} & 2020 & FINETUNE & CLS & -- \\
\citet{al-badrashiny-etal-2014-automatic} & 2014 & PIPELINE\_MAP & OTHER & ar-arz \\
\citet{yong-etal-2023-prompting} & 2023 & AUG\_SYN & GEN & multi \\
\citet{chang19_interspeech} & 2018 & AUG\_SYN & LM & zh-en \\
\citet{10.1145/3404835.3463102} & 2021 & RESOURCE & SEQ & multi \\
\citet{SHARMA2022102760} & 2021 & PIPELINE\_MAP & CLS & hi-en \\
\citet{nayak-joshi-2022-l3cube} & 2022 & PRETRAIN & LM & hi-en \\
\citet{gupta-etal-2020-semi} & 2020 & AUG\_SYN & GEN & multi \\
\citet{bhat-etal-2018-universal} & 2018 & RESOURCE & SEQ & hi-en \\
\citet{CHAKRAVARTHI2022100119} & 2022 & AUG\_SYN & CLS & multi \\
\citet{barik-etal-2019-normalization} & 2019 & PIPELINE\_MAP & OTHER & id-en \\
\citet{pratapa-etal-2018-word} & 2018 & AUG\_SYN & LM & es-en \\
\bottomrule\end{tabular}
\caption{Most-cited papers per direction and modality, part 5 of 6.}\label{tab:papers-5}
\end{table}

\begin{table}[tp]\centering\footnotesize\setlength{\tabcolsep}{2pt}
\begin{tabular}{@{}>{\raggedright\arraybackslash}p{2.0cm}rll>{\raggedright\arraybackslash}p{0.95cm}@{}}
\toprule Paper & Year & Mechanism & Task & Pair \\\midrule
\multicolumn{5}{@{}l}{\textit{Native to CM --- text (cont.)}} \\
\citet{10.1007/978-3-031-23804-8_9} & 2018 & EMBED\_ALIGN & CLS & hi-en \\
\citet{10.1007/978-3-030-00563-4_54} & 2018 & RESOURCE & CLS & ar-arq \\
\citet{Zhang2024CHAIFL} & 2024 & FINETUNE & MT & multi \\
\citet{sai-sharma-2021-towards} & 2021 & PIPELINE\_MAP & CLS & multi \\
\citet{muller2020can} & 2020 & FINETUNE & SEQ & ar-fr \\
\citet{khatri-etal-2023-translate} & 2023 & PROMPT\_ICL & MT & multi \\
\citet{ismail2026multi} & 2026 & ADAPT\_MERGE & OTHER & ar-fr \\
\citet{perez-etal-2022-robertuito} & 2021 & PRETRAIN & LM & es-en \\
\citet{tsai-etal-2019-small} & 2019 & FINETUNE & SEQ & multi \\
\citet{8944868} & 2019 & OTHER & CLS & hi-en \\
\citet{gahoi-etal-2022-gui} & 2022 & PIPELINE\_MAP & MT & hi-en \\
\citet{khan-etal-2022-sit} & 2022 & FINETUNE & MT & hi-en \\
\citet{adouane-bernardy-2020-multi} & 2020 & MULTITASK & SEQ & ar-ar\_dz \\
\citet{dabre-2022-nict} & 2022 & AUG\_SYN & MT & hi-en \\
\citet{11361448} & 2025 & FINETUNE & CLS & si-en \\
\citet{raheem-etal-2022-domain} & 2022 & MULTITASK & MT & hi-en \\
\citet{emran2024masontigers} & 2024 & FINETUNE & CLS & hi-en \\
\citet{10651379} & 2024 & ADAPT\_MERGE & CLS & multi \\
\citet{11525909} & 2025 & MULTITASK & CLS & bn-en \\
\citet{dhar-etal-2018-enabling} & 2018 & PIPELINE\_MAP & MT & hi-en \\
\citet{10129187} & 2023 & AUG\_SUB & CLS & bn-en \\
\citet{tan-joty-2021-code} & 2021 & AUG\_SUB & NLU & multi \\
\citet{peng-etal-2014-learning} & 2014 & OTHER & OTHER & es-en; zh-en \\
\citet{bansal-etal-2020-code} & 2020 & OTHER & CLS & hi-en \\
\citet{mohanapriya-etal-2025-comparative} & 2025 & MULTITASK & CLS & multi \\
\bottomrule\end{tabular}
\caption{Most-cited papers per direction and modality, part 6 of 6.}\label{tab:papers-6}
\end{table}

\end{document}